\pdfoutput=1
\documentclass[11pt]{article}

\usepackage[final]{acl}

\usepackage{times}
\usepackage{latexsym}

\usepackage[T1]{fontenc}
\usepackage[utf8]{inputenc}

\usepackage{microtype}

\usepackage{inconsolata}

\usepackage{graphicx}

\usepackage{multirow}
\usepackage{makecell}
\usepackage{booktabs}

\usepackage{amsmath}
\usepackage{amssymb}
\usepackage{algorithm}
\usepackage{algorithmic}
\usepackage{bm}
\usepackage{bm}
\usepackage{bbold}

\usepackage{color, colortbl}  % color

\usepackage{enumitem}

\usepackage{adjustbox}

\usepackage{gradient-text}

\usepackage{wasysym}
\usepackage{pifont} % For checkmarks and crosses
\definecolor{kellygreen}{rgb}{0.3, 0.73, 0.09}
\definecolor{alizarin}{rgb}{0.82, 0.1, 0.26}
\definecolor{royalblue}{rgb}{0.25,0.41,1}
\newcommand{\cmark}{{\color{kellygreen} \ding{51}}}
\newcommand{\xmark}{{\color{alizarin} \ding{55}}}

\newcommand{\major}[1]{\textbf{#1}}
\newcommand{\minor}[1]{\textcolor{gray}{#1}}

\definecolor{row-blue}{HTML}{F5FFFA}
\definecolor{row-green}{HTML}{F1F6EC}
\definecolor{row-yellow}{HTML}{FFFFF0}
\definecolor{row-pink}{HTML}{FFF5F5}
\definecolor{superlightgrey}{gray}{0.7}
\newcommand{\gray}[1]{\textcolor{gray}{#1}}

\newcommand{\best}[1]{{\textbf{#1}}}
\newcommand{\second}[1]{{\underline{#1}}}

\definecolor{best}{HTML}{FFE8D9}
\definecolor{second}{HTML}{DAE3F5}
\newcommand{\bestcell}[1]{\colorbox{best}{#1}}
\newcommand{\secondcell}[1]{\colorbox{second}{#1}}
\newcommand{\up}[1]{\textcolor{red}{\scriptsize (+#1)}}
\newcommand{\down}[1]{\textcolor[RGB]{46,139,87}{\scriptsize (-#1)}}
\newcommand{\tie}[1]{\textcolor{gray}{\scriptsize (-)}}

\definecolor{narrative}{HTML}{E9C5C4}
\definecolor{event}{HTML}{FAF5BF}
\definecolor{atomic}{HTML}{CFF1B6}

\usepackage{xspace}
\newcommand\dataset{\textsc{Tuna}-1K\xspace}
\newcommand\bench{\textsc{Tuna}\xspace}

\newcommand\benchcap{\textsc{Tuna}-\textsc{cap}\xspace}
\newcommand\benchmcq{\textsc{Tuna}-\textsc{mcq}\xspace}

\title{
\raisebox{-0.25\height}{\includegraphics[width=0.05\textwidth]{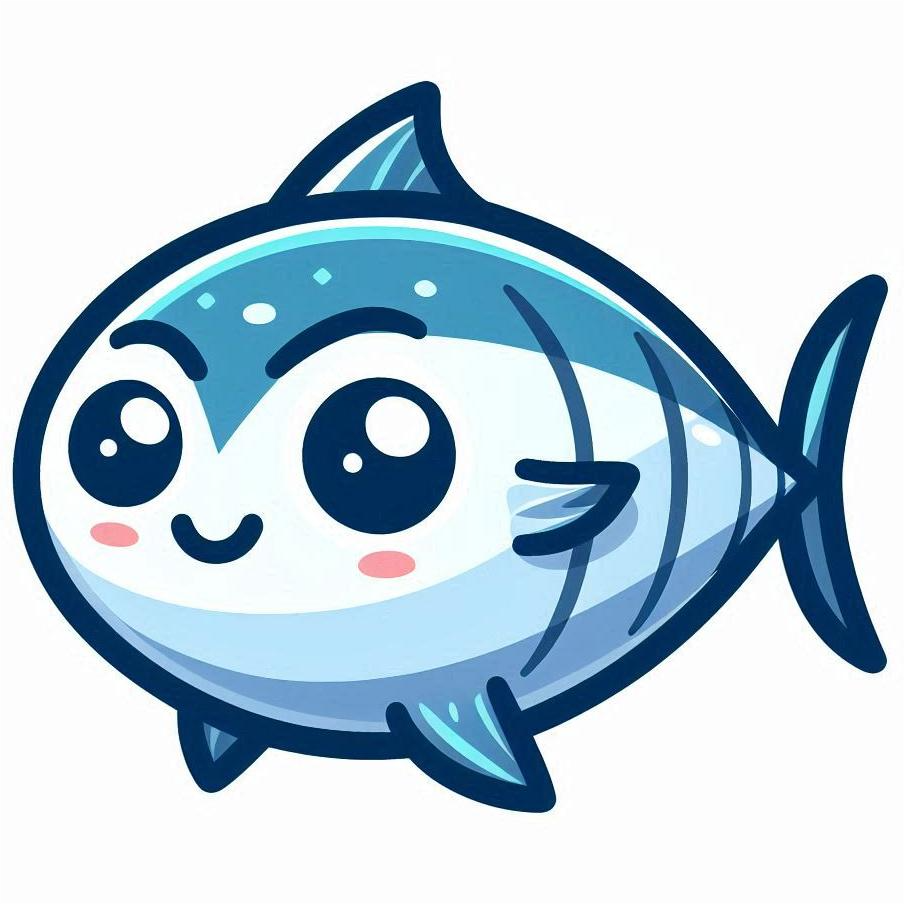}}\ \bench: Comprehensive Fine-grained Temporal \\ \qquad Understanding Evaluation on Dense Dynamic Videos}

\author{
Fanheng Kong\textsuperscript{1\thanks{Work done during an internship at Kuaishou Technology.}}, Jingyuan Zhang\textsuperscript{2}, Hongzhi Zhang\textsuperscript{2}, Shi Feng\textsuperscript{1\thanks{Corresponding Author.}}, \textbf{Daling Wang}\textsuperscript{1}, \\
\textbf{Linhao Yu}\textsuperscript{2}, \textbf{Xingguang Ji}\textsuperscript{2}, \textbf{Yu Tian}\textsuperscript{2}, \textbf{Victoria W.}, \textbf{Fuzheng Zhang}\textsuperscript{2} \\
\textsuperscript{1}Northeastern University \quad
\textsuperscript{2}Kuaishou Technology \\
\tt\small kongfanheng426@gmail.com, \tt\small fengshi@cse.neu.edu.cn
}

\begin{document}
\maketitle
\begin{abstract}

Videos are unique in their integration of temporal elements, including camera, scene, action, and attribute, along with their dynamic relationships over time. However, existing benchmarks for video understanding often treat these properties separately or narrowly focus on specific aspects, overlooking the holistic nature of video content. To address this, we introduce \bench, a temporal-oriented benchmark for fine-grained understanding on dense dynamic videos, with two complementary tasks: captioning and QA. Our \bench features diverse video scenarios and dynamics, assisted by interpretable and robust evaluation criteria.
We evaluate several leading models on our benchmark, providing fine-grained performance assessments across various dimensions. This evaluation reveals key challenges in video temporal understanding, such as limited action description, inadequate multi-subject understanding, and insensitivity to camera motion, offering valuable insights for improving video understanding models. 
The data and code are available at \url{https://friedrichor.github.io/projects/TUNA}.

\end{abstract}

\section{Introduction}

Vision enables us to perceive the world, and video, as a key form of visual media, offers rich spatial and temporal information \citep{tang2023video, madan2024foundation}. With the rapid growth of video content, video understanding has become a crucial area of research, enabling applications that address the increasing volume of video data \citep{nguyen2024video} and facilitate video generation as general-purpose simulators of the physical world \citep{videoworldsimulators2024}. Despite these advancements, the lack of robust evaluation methods remains a pressing challenge for the community. Accurate and comprehensive benchmarks are essential to assess the performance of video understanding models and improve their ability to interpret and analyze diverse video data effectively.

\begin{figure}[t]
    \centering
    \includegraphics[width=\linewidth]{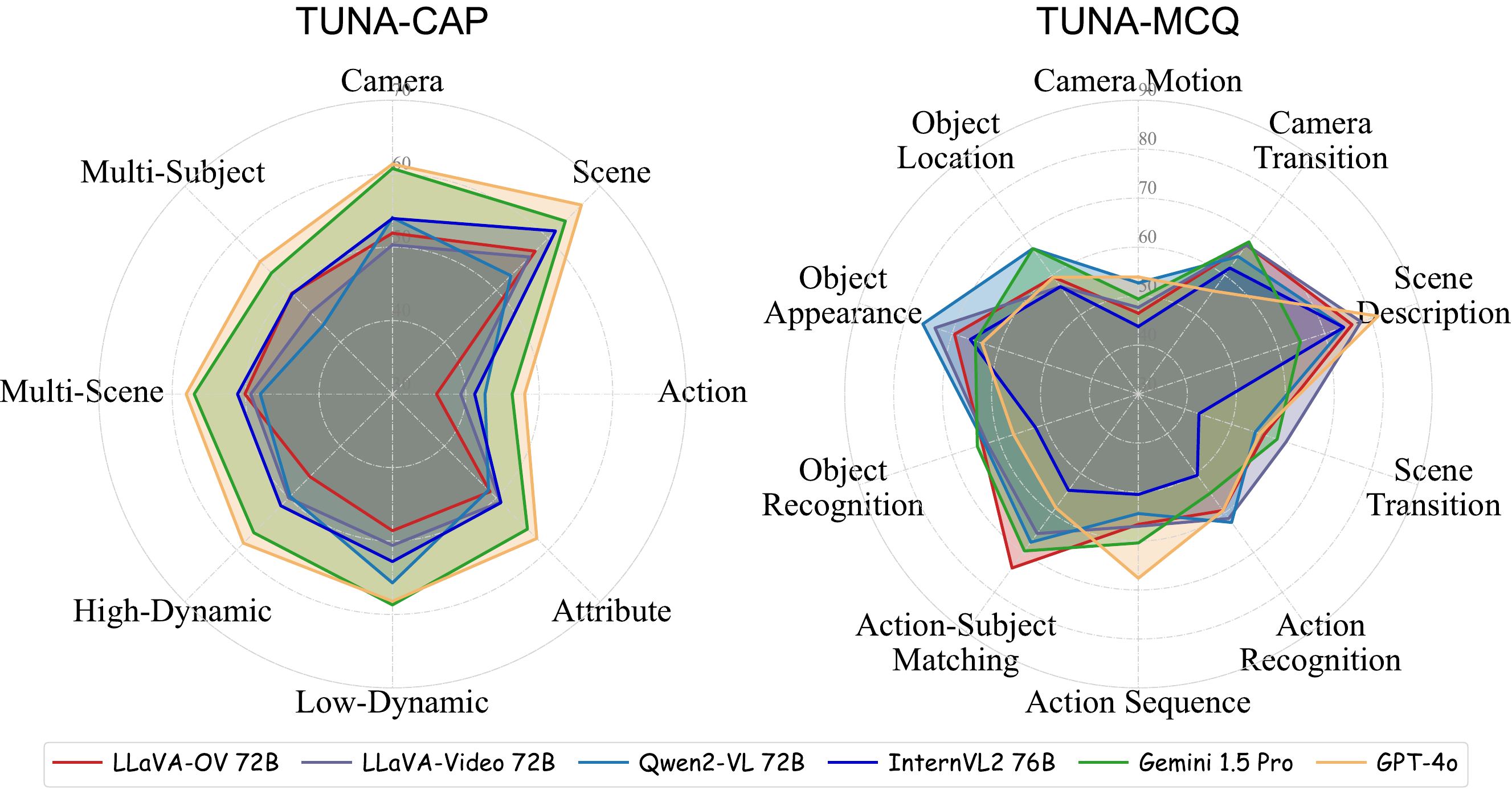}
    \caption{Performance of several advanced models on our \bench.  \bench offers robust and interpretable evaluations on video captioning and QA tasks, providing clear guidance for advancements in video understanding.}
    \label{fig:radar_map}
    \vspace{-1.5em}
\end{figure}

\begin{table*}[!htbp]
  \centering
  \resizebox{\textwidth}{!}{
    \begin{tabular}{lcccccccccccccc}
    \toprule
    \multirow{2}[2]{*}{Benchmark} & \multirow{2}[2]{*}{\makecell{\#Videos}} & \multirow{2}[2]{*}{\#Samp.} & \multirow{2}[2]{*}{Anno.} & \multirow{2}[2]{*}{Domain} & \multirow{2}[2]{*}{\makecell{Temporal \\ Oriented}} & \multirow{2}[2]{*}{\makecell{Scene \\ Trans.}} & \multicolumn{5}{c}{Captioning} & \multicolumn{2}{c}{VQA} \\
    \cmidrule(lr){8-12} \cmidrule(lr){13-14}
    & & & & & & & Camera & Scene & Key. & Sem. & M.D. & Global & Fine. \\
    \midrule
    \rowcolor{row-green}\multicolumn{14}{l}{\gray{\textit{\textbf{VQA Benchmark}}}}\\
    \rowcolor{row-green}
    NExT-QA \citep{xiao2021nextqa} & 1,000 & 8,564 & M & daily life & \xmark & \xmark & - & - & - & - & - & \xmark & \cmark \\
    \rowcolor{row-green}
    EgoSchema \citep{mangalam2023egoschema} & 5,063 & 5,063 & M\&A & egocentric & \cmark & \xmark & - & - & - & - & - & \cmark & \xmark \\
    \rowcolor{row-green}
    PerceptionTest \citep{patraucean2024perceptiontest} & 11,620 & 44,000 & M & indoor & \cmark & \xmark & - & - & - & - & - & \cmark & \cmark \\
    \rowcolor{row-green}
    MVBench \citep{li2024mvbench} & 3,641 & 4,000 & A & open & \cmark & \cmark & - & - & - & - & - & \cmark & \cmark \\
    \rowcolor{row-green}
    Video-MME \citep{fu2024videomme} & 900 & 2,700 & M & open & \xmark & \cmark & - & - & - & - & - & \xmark & \cmark \\
    \rowcolor{row-green}
    MMBench-Video \citep{fang2024mmbenchvideo} & 609 & 1,998 & M & open & \xmark & \cmark & - & - & - & - & - & \cmark & \cmark \\
    \rowcolor{row-green}
    VideoVista \citep{li2024videovista} & 894 & 24,906 & A & open & \xmark & \cmark & - & - & - & - & - & \xmark & \cmark \\
    \rowcolor{row-green}
    TOMATO \citep{shangguan2024tomato} & 1,417 & 1,484 & M & open & \cmark & \cmark & - & - & - & - & - & \cmark & \xmark \\
    \midrule
    \rowcolor{row-yellow}\multicolumn{14}{l}{\gray{\textit{\textbf{Captioning Benchmark}}}}\\
    \rowcolor{row-yellow}
    DREAM-1K \citep{wang2024tarsier} & 1,000 & 1,000 & M & open & \cmark & \cmark & \xmark & \xmark & \xmark & \cmark & \xmark & - & - \\
    \rowcolor{row-yellow}
    VDC \citep{chai2024auroracap} & 1,027 & 1,027 & A & open & \cmark & \cmark & \cmark & \cmark & \xmark & \cmark & \cmark & - & - \\
    \midrule
    \rowcolor{row-pink}\multicolumn{14}{l}{\gray{\textit{\textbf{Multi-task Benchmark}}}} \\
    \rowcolor{row-pink}
    MLVU \citep{zhou2024mlvu} & 1,334 & 2,593 & M & open & \xmark & \cmark & \xmark & \cmark & \xmark & \cmark & \xmark & \xmark & \cmark \\
    \rowcolor{row-pink}
    TempCompass \citep{liu2024tempcompass} & 410 & 7,540 & M\&A & open & \cmark & \cmark & \xmark & \xmark & \xmark & \cmark & \cmark & \cmark & \cmark \\
    \rowcolor{row-pink}
    E.T.Bench \citep{liu2024etbench} & 7,002 & 7,289 & M & open & \cmark & \cmark & \xmark & \xmark & \xmark & \cmark & \xmark & \cmark & \cmark \\
    \rowcolor{row-pink}
    TemporalBench \citep{cai2024temporalbench} & 2,179 & 2,179 & M & open & \cmark & \xmark & \xmark & \xmark & \xmark & \xmark & \xmark & \cmark & \cmark \\
    \midrule
    \rowcolor{row-blue}
    \bench & 1,000 & 2,432 & M\&A & open & \cmark & \cmark & \cmark & \cmark & \cmark & \cmark & \cmark & \cmark & \cmark \\
    \bottomrule
    \end{tabular}%
}
  \caption{Comparison with various video understanding benchmarks across several aspects: number of videos (\textbf{\#Videos}); number of samples (\textbf{\#Samp.}); annotation method (\textbf{Anno.}, with M/A denoting manual/automatic); domain (\textbf{Domain}); temporal orientation (\textbf{Temporal Orientated}); presence of scene transitions (\textbf{Scene Trans.}); consideration of camera (\textbf{Camera}) and scene (\textbf{Scene}); use of keypoints (\textbf{Key.}) for controllability and interpretability; Judgement of semantically identical yet diverse representations (\textbf{Sem.}); availability of multi-dimensional scores (\textbf{M.D.}); if global (\textbf{Global}) and fine-grained (\textbf{Fine.}) understanding are concerned.
  }
  \label{tab:benchmark_comparison}%
  \vspace{-1.5em}
\end{table*}

Recent works \citep{fu2024videomme, zhou2024mlvu} have evaluated video understanding across various tasks such as temporal perception and reasoning, video captioning, and long-video comprehension, providing metrics to guide the development of video LMMs. However, these evaluations often focus on specific aspects, such as subject actions, while neglecting other crucial video elements like camera states and background scenes along with the relationships between these elements \citep{chai2024auroracap, xiong2024lvd2m, polyak2024moviegen}. Additionally, the bias toward long-form videos \citep{fu2024videomme, li2024videovista, mangalam2023egoschema} entangles video understanding with long-context modeling, making it difficult to attribute performance to specific capabilities. Furthermore, existing benchmarks lack an analysis of the model's sensitivity towards key factors affecting video understanding, such as diversity of video dynamics and visual characteristics. These limitations hinder comprehensive evaluation and effective error analysis to advance video understanding models.

To address the need for comprehensive video understanding, we introduce \bench, a challenging multimodal benchmark for \textbf{T}emporal \textbf{U}nderstanding of dense dy\textbf{NA}mic videos. Unlike previous evaluations that focus on isolated video elements, \bench emphasizes holistic video comprehension. We carefully curated 1,000 representative videos from diverse sources, spanning 12 domains such as Film and Driving, categorized across four visual characteristics: High-Dynamic, Low-Dynamic, Multi-Scene, and Multi-Subject. Each video in our dataset, \dataset, is meticulously segmented into fine-grained events and annotated with detailed temporal captions, capturing camera states, background scenes, subject actions, object attributes. Table \ref{tab:benchmark_comparison} shows the comparison with various vidoe understanding benchmarks.

Building on \dataset, we propose \bench, a multi-task benchmark towards temporal dynamics through two complementary tasks: \benchcap for captioning and \benchmcq for VQA. \benchcap features an automated evaluation pipeline that performs event splitting, matching, and relationship classification, closely aligning with human judgment to assess dense captioning capabilities. \benchmcq comprises 1,432 carefully crafted multiple-choice questions that specifically require full video context for accurate answers, ensuring that answers cannot be derived from a single frame or limited frames, providing a rigorous test of temporal understanding. Together, these tasks provide comprehensive evaluation metrics and valuable insights for advancing video understanding research.

\begin{figure*}[t]
    \centering
    \includegraphics[width=\linewidth]{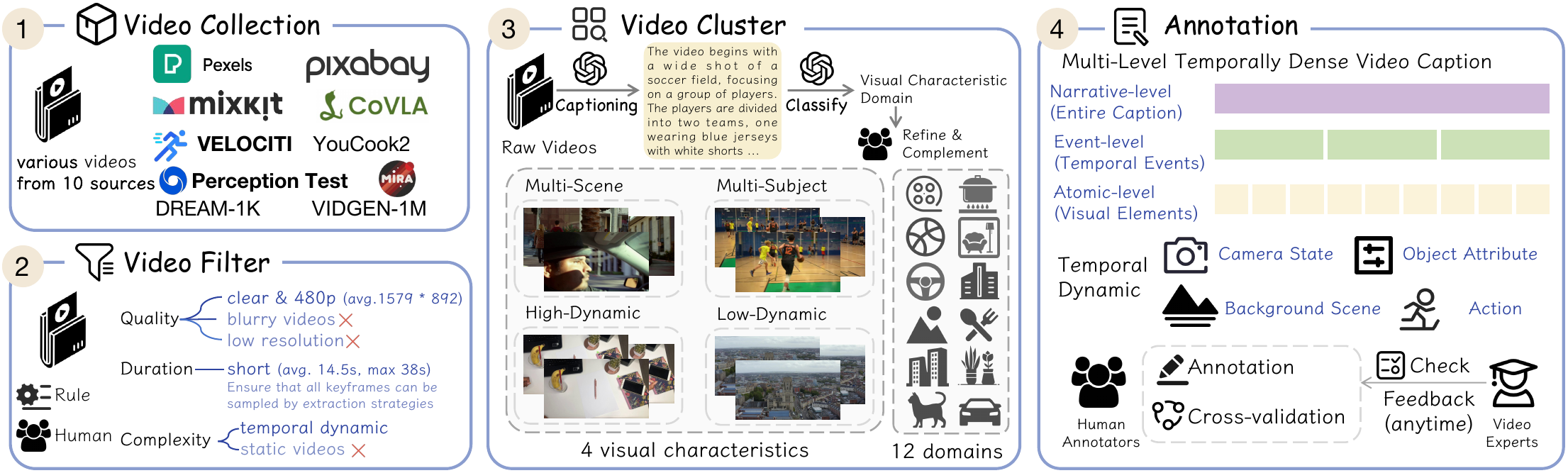}
    \caption{Overview of \dataset construction. We collect and filter high-quality, short videos featuring dynamic temporal content from various sources. Each video is then categorized based on its visual characteristics and domain. Trained annotators provide temporally dense descriptions, followed by cross-validation. Video experts continuously review annotations, guiding annotators to refine their works, thus ensuring quality of the annotations.}
    \label{fig:dataset_construction}
    \vspace{-1.2em}
\end{figure*}

We benchmark 21 popular LMMs on \bench, revealing key challenges in video understanding. Figure \ref{fig:radar_map} shows the performance of selected models. Dense video captioning remains a difficult task, with GPT-4o \citep{openai2024gpt4o} achieving the best performance but only reaching an F1 score of 58.5\%, yet open-source models lag notably behind commercial models. Additionally, LMMs struggle with complex scenarios involving multi-scene, multi-subject, and high-dynamic video content. Interestingly, in the VQA task, open-source models demonstrate competitive performance. However, all models show consistent weaknesses in comprehending camera motion and action sequence. The notable performance disparity between captioning and VQA tasks underscores the current limitations in holistic video understanding capabilities. These findings provide crucial insights for advancing video LMMs, particularly in temporal and visual comprehension capabilities.

In summary, our contributions are:
\begin{itemize}[leftmargin=*, nolistsep, noitemsep]
\item We introduce \dataset, a meticulously annotated video-caption dataset that captures fine-grained temporal dynamics across camera, scene, action, and attribute on dense dynamic videos.
\item We develop \bench, a novel benchmark for comprehensive temporal video understanding, measuring the performance across various novel dimensions, such as different visual characteristics, temporal elements and video complexities.
\item We conduct a comprehensive evaluation of several popular models, uncovering their strengths and weaknesses across various dimensions. Hopefully, this provides solid guidance for advancing video understanding.
\end{itemize}

\section{Related Work}

% \vspace{-0.5em}
\noindent\textbf{Video Captioning.} Recent works \citep{zhang2024llavahound, zhang2024llavavideo, chen2024sharegpt4video, liu-etal-2024-devan} have revealed the importance of detailed captions for video understanding. Compared to image captioning, video captioning presents a greater challenge as it requires advanced techniques to handle the diversity of human and object appearances in various scenes, as well as their evolving relationships over time \citep{INACIO2023100488}. While video captioning data is served as training data for video LMMs, it is challenging to robustly and interpretably evaluate video captioning. Traditional n-gram overlaps based metrics \citep{papineni-etal-2002-bleu, lin-2004-rouge, vedantam2015cider} fail to measure genuine semantic similarity, with weakly consistency with human judgement. LLM-based scoring methods \citep{chan-etal-2023-clair, maaz2023videochatgpt} can deal with captions with the same semantics yet distinct expressions, but directly asking LLM to generate digital scores is not dependable due to their ambiguous meaning of each rating. 
Recently, Dream-1K \citep{wang2024tarsier} evaluates captions from events, providing a robust results. However, these efforts don't centre on temporal dynamics, overlooking the essential features of video and pay minimal attention to changes in camera and scene.

\noindent\textbf{Video QA.} Recent works has provided benchmarks for comprehensively evaluating video LMM's ability to understand video, e.g., Video-MME \citep{fu2024videomme}, MLVU \citep{zhou2024mlvu}. Temporal dynamics are crucial as a unique feature of video. Existing temporal understanding benchmarks \citep{patraucean2024perceptiontest, li2024mvbench} focus on restricted scenes (e.g., indoor, egocentric), or just on subject's actions and attributes, without attention to changes in camera and scene, which are incomplete for temporal understanding evaluation. Our \bench  aims to comprehensively evaluate temporal perception skills towards open-domain videos.

\begin{figure*}[t]
    \centering
    \vspace{-1.2em}
    \includegraphics[width=\linewidth]{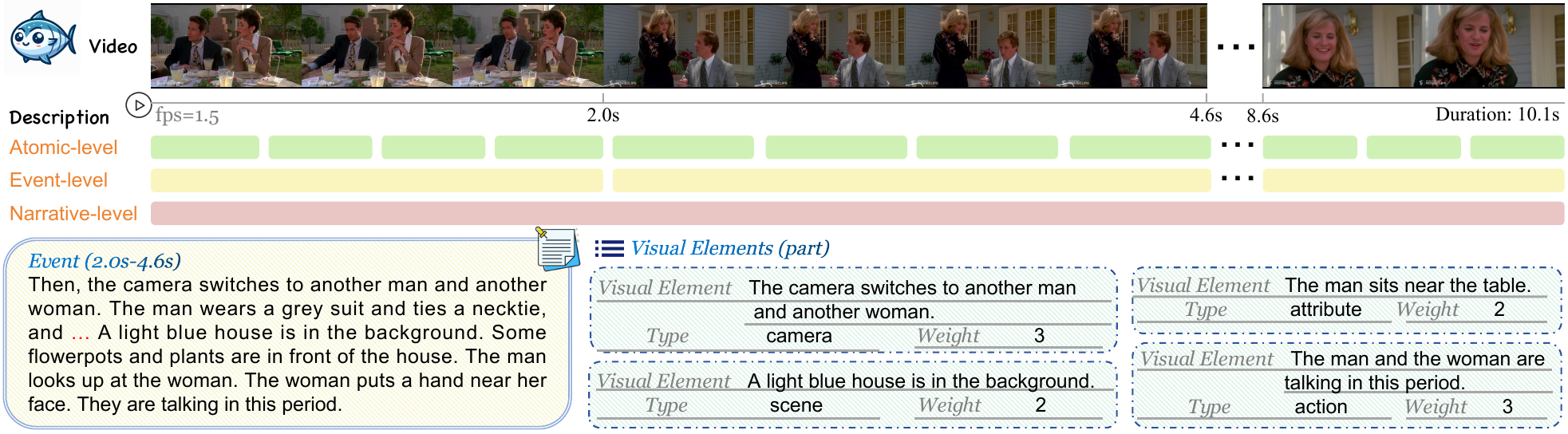}
    \caption{An instance in \dataset consists of three levels of description: (a) an overall caption (\colorbox{narrative}{\makebox(54,5){Narrative-level}}), (b) a chronological sequence of events (\colorbox{event}{\makebox(42,5){Event-level}}), and (c) fine-grained visual elements (\colorbox{atomic}{\makebox(48,5){Atomic-level}}) along with their types and weights. A complete sample can be found in Figure \ref{fig:dataset_example_detail}.}
    \label{fig:dataset_example}
    \vspace{-1.5em}
\end{figure*}

\vspace{-0.5em}
\section{\bench}

\vspace{-0.5em}
In this section, we present \dataset, a temporally dense video-caption dataset, and \bench, a multi-task temporal understanding benchmark.

\vspace{-0.5em}
\subsection{\dataset}

The construction workflow of \dataset is shown in Figure \ref{fig:dataset_construction}, consisting of four major phases: video collection, filter, cluster, and annotation.

\noindent\textbf{Video Collection.} Temporally dense videos should have diverse contents and include changes in the camera states and scenes besides subject actions and object attributes \citep{polyak2024moviegen, xiong2024lvd2m}. To capture these complexities, we carefully collect 1,000 open-domain videos from 10 sources: 
% Pexels\footnote{\url{https://www.pexels.com/videos/}}, Pixabay\footnote{\url{https://pixabay.com/videos/}}, MixKit\footnote{\url{https://mixkit.co/free-stock-video/}}; 
(1) \textit{Academic Video Understanding Data}: DREAM-1K \citep{wang2024tarsier}, Perception Test \citep{patraucean2024perceptiontest}, VELOCITI \citep{saravanan2024velociti}, YouCook2 \citep{zhou2018youcook2}; (2) \textit{Academic Video Generation Data}: MiraData \citep{ju2024miradata}, VIDGEN-1M \citep{tan2024vidgen1m}); (3) \textit{Other Academic Video Data}: CoVLA \citep{arai2024covla}; and (4) \textit{Web Data}: 
Pexels \citep{pexels}, Pixabay \citep{pixabay}, MixKit \citep{mixkit}. Unlike concurrent works \citep{cai2024temporalbench}, we maintain the original videos containing multiple scenes or complex actions without segmenting them into clips, as these are essential for our tasks.

\noindent\textbf{Video Filter.} We remove blurry, low-resolution and long duration videos to ensure that our videos are high-quality and short, with an average resolution of $1579*892$, and an average duration of $14.5s$. We select short videos to ensure that sampling frame strategies can extract all keyframes from videos, to purely examine the video understanding ability. To ensure the videos are temporal-dynamic, one criterion is rich in either camera motion, scene transitions, or subject activities. Coarse filtering (e.g., resolution) is achieved by rules, and humans make complex filters (e.g., dynamic degree).

\noindent\textbf{Video Cluster.} We use GPT-4o \citep{openai2024gpt4o} to generate description for each video, and cluster videos based on their descriptions, including four visual characteristics and 12 domains. Annotators then correct and complete the classification results.

\begin{figure*}[t]
    \centering
    \includegraphics[width=\linewidth]{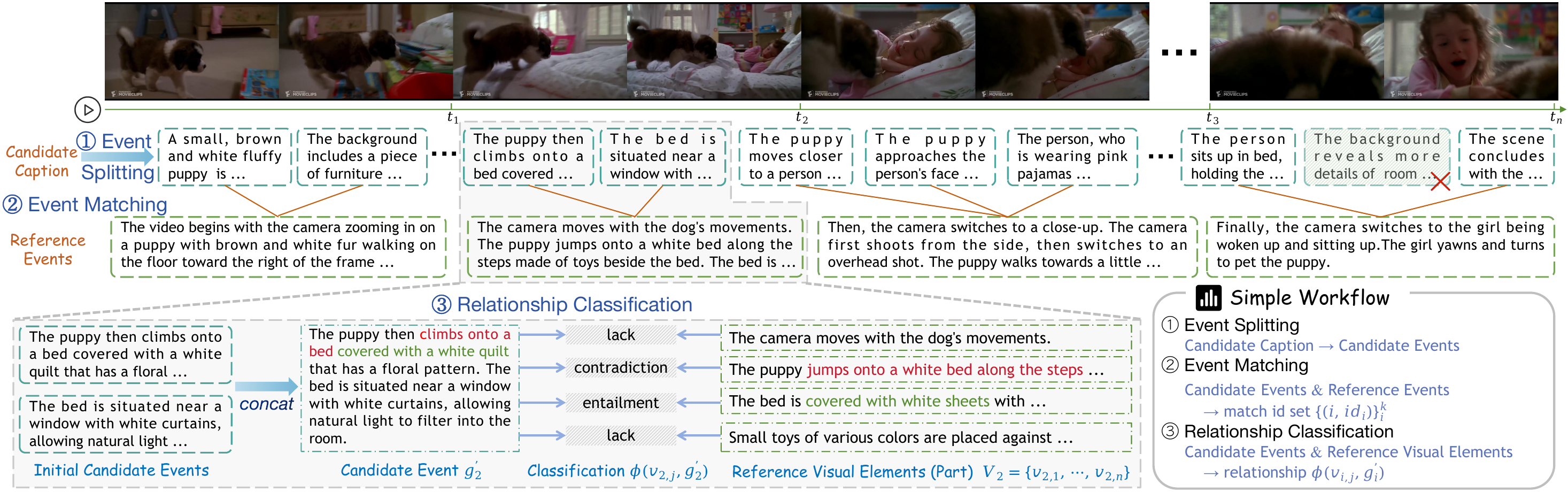}
    \caption{Overview of the evaluation workflow for \benchcap. We first split candidate caption into multiple events and match them to reference events in \dataset. Then we discard the mismatched events (useless content or inconsistent chronology), and connect the matched candidate events with the same reference event, considering the temporal sequence of the captions. Finally, we classify the relationship of visual elements to the candidate event.}
    \label{fig:captioning_evaluation}
    \vspace{-1.5em}
\end{figure*}

\noindent\textbf{Annotation.} Existing models often miss critical events \citep{wang2024tarsier}, and lack sensitivity to camera states, struggling to accurately describe camera changes \citep{chai2024auroracap}. Consequently, generating temporally dense video captions through automatic methods is challenging. Instead, our data is manually annotated. 
Trained human annotators are tasked to provide detailed video descriptions, focusing on camera states, background scenes, subject actions, and object attributes. The target captions features several chronologically evolving events, without summaries and subjective feelings. Additionally, annotators split each event into multiple visual elements, assigning types and weights to these elements. The types include \texttt{camera}, \texttt{scene}, \texttt{action}, and \texttt{attribute}, while weights indicates the element's importance for the video on a scale of 1-3. 

Formally, a typical instance in \dataset involves a collection of temporally evolving events $E_{ref}=[r_1, r_2, \ldots, r_T]$ forming an overall caption $C_{ref}$, where $T$ denotes the count of events in the sequence. Each event $r_i$ further contains various visual elements $V_i = \{v_{i1}, \ldots, v_{i,n_i}\}$, where $n_i$ represents the number of visual elements in event $r_i$. Moreover, each visual element $v_{ij}$ is labelled with a type $t \in \{\texttt{camera}, \texttt{scene}, \texttt{action}, \texttt{attribute}\}$ and their weight $w_{ij} \in \{1,2,3\}$. An example of TUNA-1K is shown in Figure \ref{fig:dataset_example}.

\noindent\textbf{Quality Review.} All annotated video-caption pairs undergo cross-inspection by annotators. In parallel, video experts (non-authors) review the annotations, providing feedback and prompting annotators to refine results to ensure high-quality annotation.

% \vspace{-0.5em}
\subsection{\bench}
\label{sec:method_bench}

% \vspace{-0.3em}
\subsubsection{Task Definition}

Temporal dynamics distinguish videos from static images. While several benchmarks consider temporal sequences, they sole focus on actions and attributes, neglecting changes in camera state and scene. Additionally, some evaluation tasks fail to capture the perception ability of relationships and evolution of various elements in the video. For example, many questions are just about a single frame cue in the video. To fill these gaps, we emphasize the in-context understanding throughout the entire video, and measure temporal understanding across 4 key dynamic elements: camera state, background scene, subject action, and object attribute. Specifically, we introduce two complementary tasks: \benchcap for captioning and \benchmcq for VQA.

% \vspace{-0.5em}
\subsubsection{\benchcap}
\label{sec:method_captioning}

An effective way to evaluate the temporal understanding ability of LMMs is reflected by their captioning skills \citep{INACIO2023100488, chen2024sharegpt4video}. However, it remains a challenge to reliably and interpretably assess the correctness and completeness of video captions. Event-level methods \citep{wang2024tarsier, wang2022geb+} have proven effective but solely focus on subject actions, overlooking camera states and scenes. To this end, we propose a strategy to assess the temporally dense captions that incorporate dynamic elements evolutions over time.

As shown in Figure \ref{fig:captioning_evaluation}, our evaluation proceeds in three stages: (1) Event Splitting, (2) Event Matching, and (3) Relationship Classification.

\noindent\textbf{Event Splitting \& Matching.} To examine temporal perception skills through model-generated captions, we consider that an effective solution is to verify whether the models accurately describe several events in the correct temporal sequence. To achieve this, the candidate caption $C_{\text{gen}}$ is first split into an event sequence $G=[g_1, g_2, \ldots, g_k]$. Then, each candidate event $g_i$ is matched to a reference event $r_j$. Formally, the target is to obtain $\{(i, id_i)\}_{i=1}^k$ pairs, where $id_i \in \{1, \ldots, T, \texttt{None}\}$ denotes the index of the reference event $r_{id_i}$ matched with the candidate event $g_i$ and $id_1 \leq id_2 \leq \cdots \leq id_k$. These ensures that events which are effective and described in a correct temporal order are extracted.

\noindent\textbf{Relationship Classification.} For the captioning task, the classification-based approach is more interpretable and robust than the direct scoring methods \citep{wang2024tarsier}. Each reference event $r_j$ corresponds to a set of visual elements $V_j$. Thus, we can transition from a tuple of concatenated candidate events with reference events $(g_i',r_j)$ to a tuple of candidate events with visual elements $(g_i',V_j)$. Subsequently, the relationship $\phi(v_{ij}, g_i') \in \{\texttt{entailment}, \texttt{lack}, \texttt{contradiction}\}$ between visual element $v_{ij}$ and candidate event $g_i'$ is classified. This element-based approach improves the interpretability of the evaluation. The workflow is implemented by GPT-4o \citep{openai2024gpt4o}, an LLM with powerful instruction-following capabilities.

\begin{table*}[!htbp]
  \centering
  \resizebox{\textwidth}{!}{
    \begin{tabular}{l|cccc|cccc|c}
    \toprule
    \multirow{2}[2]{*}{Model} & \multicolumn{4}{c|}{Dynamic Element Type} & \multicolumn{4}{c|}{Visual Characteristic} & \multirow{2}[2]{*}{Overall} \\
    \cmidrule(lr){2-5} \cmidrule(lr){6-9}
    & Camera & Scene & Action & Attribute & Low-Dynamic & High-Dynamic & Multi-Scene & Multi-Subject & \\
    \midrule
    \rowcolor{gray!10}\multicolumn{10}{c}{\textit{\textbf{Open-Source LMMs}}}\\
    PLLaVA-7B 
    & 49.4/22.6/28.9 & 52.2/30.9/36.6 & 30.5/12.6/16.5 & 44.5/19.5/25.3 & 66.5/23.0/32.7 & 56.6/17.1/24.7 & 55.7/15.5/22.8 & 56.2/15.3/22.5 & 60.0/19.1/27.4 \\
    LongVA-7B
    & 52.3/26.0/32.5 & 56.5/34.4/40.6 & 38.9/17.2/22.0 & 50.6/22.0/28.4 & 75.9/26.5/37.3 & 69.4/20.1/29.0 & 68.3/19.0/27.6 & 67.3/15.7/23.7 & 71.6/22.3/31.8 \\
    Tarsier-7B 
    & 56.9/27.3/34.8 & 45.3/28.2/33.1 & 56.7/28.9/36.2 & 56.4/26.0/33.3 & \best{81.2}/34.3/46.5 & 68.7/24.5/34.5 & 71.7/25.3/35.8 & 67.8/23.2/33.2 & 73.0/27.9/38.6 \\
    Kangaroo 
    & 65.2/36.5/44.1 & 67.8/45.4/51.9 & 49.3/26.0/31.9 & 59.8/32.2/39.5 & 73.2/34.7/45.6 & 67.6/31.3/41.1 & 66.2/29.7/39.3 & 63.5/26.3/35.7 & 69.5/32.5/42.7 \\
    LLaVA-OV-7B 
    & 75.2/42.0/51.0 & 71.8/51.2/57.6 & 54.1/30.4/36.8 & 66.2/42.0/49.3 & 78.6/38.4/50.0 & 71.0/38.8/48.9 & 71.7/38.3/48.4 & 67.1/33.8/43.8 & 73.6/38.6/49.3 \\
    LLaVA-Video-7B 
    & 74.0/41.5/50.4 & \second{73.6}/52.3/58.9 & 57.0/30.8/37.8 & \best{72.1}/\best{44.8}/\best{53.1} & \second{80.7}/40.0/52.2 & 75.1/39.5/50.3 & \second{77.1}/38.6/50.0 & 73.5/34.6/45.8 & 77.0/39.7/51.0 \\
    Qwen2-VL-7B 
    & 72.3/40.7/49.0 & 71.9/50.0/56.7 & 55.9/30.1/37.0 & 68.2/38.4/46.7 & \best{81.2}/42.0/\second{53.8} & \best{76.0}/35.3/46.4 & 76.8/33.2/44.4 & \second{73.6}/28.9/39.9 & \best{77.8}/37.6/48.9 \\
    InternVL2-8B 
    & 64.8/33.7/41.7 & 59.4/38.7/44.7 & 45.2/24.7/30.0 & 59.8/35.5/42.3 & 71.6/34.0/44.5 & 64.9/29.7/38.9 & 65.6/29.1/38.4 & 61.5/26.6/35.2 & 67.2/31.1/40.8 \\
    MiniCPM-V-2.6 
    & \best{76.5}/\best{47.8}/\best{56.0} & \best{75.0}/\second{54.1}/\second{60.6} & 57.2/31.8/38.8 & \second{68.7}/42.3/50.2 & 79.3/41.4/53.0 & 74.3/\second{40.4}/\second{51.0} & 76.5/\second{40.8}/\best{51.7} & 73.5/38.3/49.0 & 76.0/40.7/\second{51.7} \\
    \midrule
    PLLaVA-34B 
    & 60.8/29.6/37.4 & 56.2/33.7/39.9 & 38.7/17.3/22.3 & 55.1/26.1/33.2 & 74.5/28.1/38.9 & 64.3/22.6/31.8 & 63.9/21.3/30.2 & 60.7/19.2/27.6 & 67.8/24.5/34.2 \\
    Tarsier-34B 
    & 63.6/34.3/42.3 & 59.0/38.4/44.4 & \best{65.6}/\best{39.9}/\best{47.6} & 63.6/34.3/42.2 & 79.6/37.2/49.1 & \second{75.8}/36.5/47.8 & \best{77.6}/38.1/49.6 & \best{74.4}/36.0/47.3 & \second{77.1}/36.7/48.2 \\
    \midrule
    LLaVA-OV-72B 
    & 73.5/43.7/51.9 & 71.5/51.1/57.5 & 51.2/30.2/36.0 & 65.7/41.4/48.8 & 75.4/37.3/48.6 & 71.3/36.7/45.9 & 71.4/40.1/50.1 & 72.3/\second{39.1}/\best{49.4} & 72.7/39.2/49.6 \\
    LLaVA-Video-72B 
    & 72.7/41.7/50.3 & 71.1/49.9/56.4 & 55.7/32.7/39.3 & 68.1/43.2/50.8 & 77.3/39.2/50.6 & 71.9/39.8/50.0 & 73.9/38.6/49.3 & 70.5/35.1/45.7 & 73.7/39.6/50.2 \\
    Qwen2-VL-72B 
    & 73.6/45.9/54.0 & 67.6/46.3/52.8 & \second{59.1}/\second{35.7}/\second{42.6} & 66.6/40.7/48.5 & 79.2/\best{44.6}/\best{55.7} & 72.4/39.3/49.7 & 73.6/37.2/48.0 & 69.1/32.8/43.3 & 74.7/\second{41.1}/\second{51.7} \\
    InternVL2-76B 
    & \second{75.1}/\second{45.4}/\second{53.9} & 73.3/\best{55.8}/\best{61.4} & 55.7/34.9/41.2 & 64.3/\second{44.5}/\second{50.9} & 72.0/\second{43.1}/52.8 & 70.1/\best{41.9}/\best{51.5} & 71.4/\best{41.1}/\second{51.1} & 68.6/\best{39.7}/\second{49.3} & 70.7/\best{42.3}/\best{51.9} \\
    \midrule
    \rowcolor{gray!10}\multicolumn{10}{c}{\textit{\textbf{Closed-Source LMMs}}}\\
    Gemini 1.5 Flash & 74.6/52.8/59.6 & \second{77.2}/\second{59.3}/\second{65.1} & 58.7/36.4/42.9 & \second{69.0}/48.4/55.2 & 74.0/46.5/56.0 & 72.0/46.4/55.5 & 73.4/46.2/55.9 & \second{73.4}/\best{46.2}/\best{55.9} & 72.7/46.4/55.7 \\
    Gemini 1.5 Pro & \second{78.7}/\second{53.0}/\second{60.7} & 75.7/57.4/63.3 & \second{59.0}/\second{40.3}/\second{46.3} & \second{69.0}/\second{49.4}/\second{56.0} & \second{76.7}/\best{48.7}/\best{58.7} & \second{72.1}/\second{47.8}/\second{56.7} & \second{73.4}/\best{47.7}/\second{57.0} & 69.9/44.1/53.3 & \second{73.7}/\second{48.1}/\second{57.4} \\
    GPT-4o & \best{80.1}/\best{53.3}/\best{61.3} & \best{79.5}/\best{60.2}/\best{66.4} & \best{64.0}/\best{41.1}/\best{48.0} & \best{73.8}/\best{50.1}/\best{57.8} & \best{79.1}/\second{47.3}/\second{58.2} & \best{77.0}/\best{48.6}/\best{58.7} & \best{78.7}/\second{47.2}/\best{58.1} & \best{76.8}/\second{44.4}/\second{55.5} & \best{77.7}/\best{48.2}/\best{58.5} \\
    \bottomrule
    \end{tabular}%
}
  \caption{\benchcap performance of representative video LMMs. We provide detailed scores for selected tested models in various perception skills and visual characteristic categories. Each cell contains "\textbf{Precision} / \textbf{Recall} / \textbf{F1 Score}". The best and second-best results are marked with \best{bold} and \second{underline}, respectively.}
  \label{tab:results_cap_full}%
  \vspace{-1.2em}
\end{table*}

\noindent\textbf{Metrics.} We employ precision (P) and recall (R) to measure the correctness and completeness of the captions, introducing a novel metric calculation:

{
\vspace{-0.7em}
\small
\begin{flalign}
    \text{P} &=\frac{\sum_{i=1}^{T} \sum_{j=1}^{n_i} \mathbb{1}\left(\phi(v_{ij}, g_i') = \operatorname{ent.}\right) \cdot w_{ij}}{\sum_{i=1}^{T} \sum_{j=1}^{n_i} \mathbb{1}\left(\phi(v_{ij}, g_i') \in \{\operatorname{ent.}, \operatorname{con.}\}\right) \cdot w_{ij}} \\
    \text{R} &=\frac{\sum_{i=1}^{T} \sum_{j=1}^{n_i} \mathbb{1}(\phi(v_{ij}, g_i') = \operatorname{ent.}) \cdot w_{ij}}{\sum_{i=1}^{T} \sum_{j=1}^{n_i} w_{ij}} \\
    \text{F1} &=\frac{2\times \text{P}\times \text{R}}{\text{P}+\text{R}}
    \label{equ:cap_eval}
\end{flalign}
\vspace{-1em}
}

\vspace{-0.5em}
\noindent where $\mathbb{1}(\cdot)$ denotes the indicator function. Recognizing that each visual element $v_{ij}$ has a distinct importance within the video, each element is weighted by its corresponding factor $w_{ij}$.

\subsubsection{\benchmcq}
\label{sec:method_mcq}

Based on fine-grained \dataset, we design a pipeline, integrating automatic construction and manual refinement, to create instructions for multi-choice questions. The pipeline involves two main flows: error-prone points extraction and multi-choice QA generation. We consider 10 task types: 1) \textit{camera motion}, e.g, zooming, panning, and rotating. 2) \textit{camera transition}. 3) \textit{scene description}. 4) \textit{scene transition}. 5) \textit{action recognition}. 6) \textit{action sequence}. 7) \textit{action-subject matching}. 8) \textit{object recognition}. 9) \textit{object appearance}, e.g., age, dress, color, shape, number. 10) \textit{object location}. Beyond previous works \citep{li2024mvbench, liu2024tempcompass} that focus on subject actions and object attributes, we additionally emphasize camera states and scene transitions, to provide a more comprehensive assessment of temporal understanding.

\noindent\textbf{Error-prone Points Extraction.} 
To generate challenging questions, we develop an automatic approach to identify error-prone points in videos. The process involves feeding video frames and their ground-truth descriptions to video LMMs, which then identify visual elements that appear inconsistent with the textual descriptions. Leveraging LMMs' inherent limitations in visual interpretation, we utilize their misidentified elements as naturally occurring error-prone points for question generation.

\noindent\textbf{Multi-Choice QA Generation.} 
Based on predefined task types, error-prone points and textual descriptions, LLM generates several multi-choice questions for each video. To ensure these questions effectively capture temporal dynamics, we employ a temporal-indispensability filtering mechanism similar to MMBench-Video~\cite{fang2024mmbenchvideo}. Specifically, a question is considered temporal-indispensable only if it cannot be correctly answered using a single frame but requires $n$ frames (default $n=16$) for accurate comprehension. This rigorous filtering process helps maintain a high temporal-indispensability ratio in \benchmcq.

\noindent\textbf{Quality Review.} To ensure that data is high-quality and time-sensitive, we employ crowdsourcing to further filter and refine the automatically constructed data. In addition, human annotators perform cross-inspections to ensure annotation quality.

% \vspace{-0.5em}
\section{Experiments}
% \vspace{-0.3em}

\subsection{Settings}

We evaluate 21 closed-source models and open-source models with various sizes, including: Gemini 1.5 Pro \citep{reid2024gemini1.5}, Gemini 1.5 Flash \citep{reid2024gemini1.5}, GPT-4o \citep{openai2024gpt4o}, PLLaVA \citep{xu2024pllava}, LongVA \citep{zhang2024longva}, Tarsier \citep{wang2024tarsier}, InternVL2 \citep{chen2024internvl1.5}, Kangaroo \citep{liu2024kangaroo}, LLaVA-OneVision \citep{li2024llavaov}, MiniCPM-V-2.6 \citep{yao2024minicpmv}, LLaVA-Video \citep{zhang2024llavavideo}, and Qwen2-VL \citep{wang2024qwen2vl}.

By default, we uniformly sample 32 frames from each video, which is sufficient to capture the entire content of videos in our \bench. Some models have varying constraints on input length or specific recommended settings. To accommodate these variations, we employ tailored sampling strategies for these models. More details are available in Appendix \ref{appendix:captioning_settings} and Appendix \ref{appendix:mcq_settings}.

\subsection{Video Captioning}
\label{sec:experiments_captioning}

\begin{figure*}[t]
    \centering
    \includegraphics[width=\linewidth]{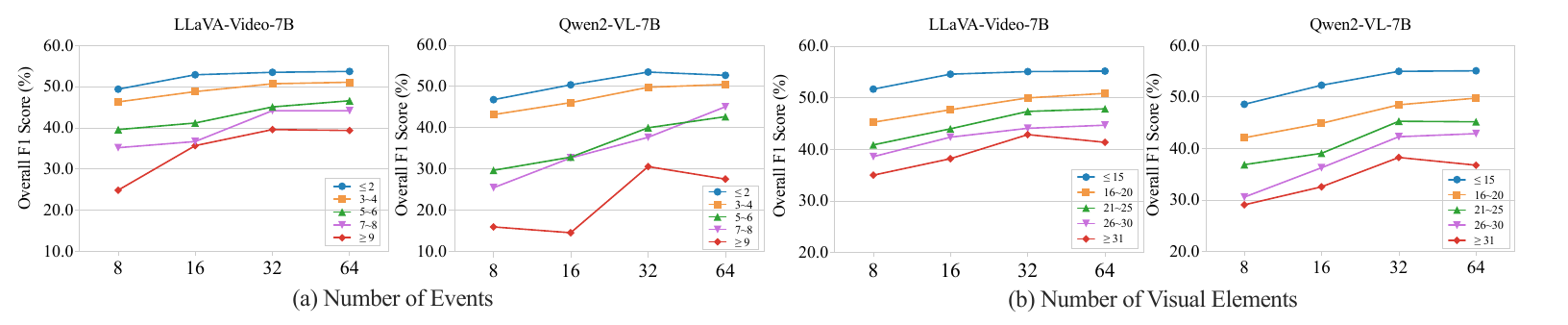}
    \setlength{\abovecaptionskip}{-0.5em}
    \caption{Performance comparison of different input frames with different video complexity for models trained in long contexts (over 8K tokens). The horizontal coordinate is the number of input frames.}
    \label{fig:comparison_nframes_complexity}
    \vspace{-1.2em}
\end{figure*}

We evaluate the temporal understanding skills of the models and their abilities to perceive videos towards different dynamic elements and visual characteristics. Precision reflects the correctness of the content mentioned in the descriptions, while recall reflects the completeness of the descriptions. As shown in Table \ref{tab:results_cap_full}, majority of video LMMs achieve a precision over 70\%, but recall is below 50\%, indicating that many visual elements in videos are often overlooked or misdescribed. The state-of-the-art model GPT-4o only achieve an F1 score of 58.5\%, with a recall of 48.2\%, highlighting that LMMs still have a great potential for improvement in the task of temporally dense captioning.

\noindent\textbf{Temporal Dynamic Elements.} Recent researches in video understanding and video generation have increasingly emphasized the dynamics of camera states and scenes \citep{chai2024auroracap, xiong2024lvd2m, polyak2024moviegen}. In this work, we comprehensively thoroughly analyze four key dynamic element types: \textit{camera}, \textit{scene}, \textit{action}, and \textit{attribute}, aiming to explore the challenges that existing models face in captioning dynamic videos. As shown in Table \ref{tab:results_cap_full}, LMMs demonstrate superior performance in scene perception compared to the other dimensions. Existing LMMs often extract multiple frames from videos and treat them as a series of static images, facilitating a better grasp of static visual scenes. However, \textit{camera} and \textit{attribute} elements, which assess overall dynamics and fine-grained perception respectively, remain challenging, with highest scores reaching only 61.3\% (56.0\% for open-source models) for camera and 57.8\% (53.1\% for open-source models) for attribute. Notably, action perception shows consistently weaker performance across almost all models compared to the other dimensions, indicating substantial shortcomings in accurately describing the dynamic actions. An interesting exception is Tarsier-34B, which performs exceptionally well in the action dimension, falling only 0.4\% behind GPT-4o. This aligns with its strong performance on DREAM-1K \citep{wang2024tarsier}, a video captioning benchmark focused on action events.

\noindent\textbf{Diverse Visual Characteristics.} As shown in Table \ref{tab:results_cap_full}, large performance disparities emerge when models process videos with different visual characteristics. All tested models perform better with low-dynamic content, but they struggle with high-dynamic and multi-scene videos, and show the weakest performance when handling videos containing multiple subjects.

\vspace{-0.5em}
\begin{figure}[!htbp]
    \centering
    \includegraphics[width=\linewidth]{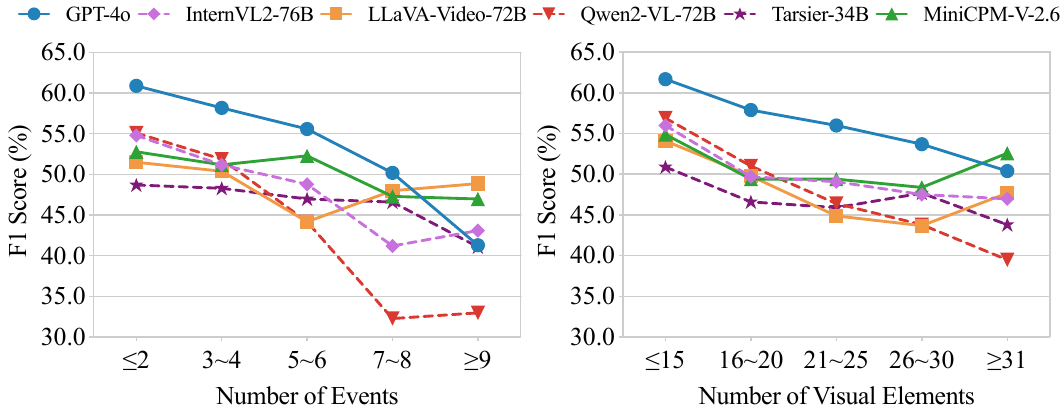}
    \setlength{\abovecaptionskip}{-0.5em}
    \caption{Performance comparison across different video complexities.}
    \label{fig:comparison_complexity}
    \vspace{-1.2em}
\end{figure}

\noindent\textbf{Video Complexity.} We partition \dataset based on the number of events and visual elements to investigate how increasing video complexity affects model performance. As shown in Figure \ref{fig:comparison_complexity}, the F1 scores demonstrate a consistent downward trend as video complexity increases, indicating that the comprehension of complex videos remains a formidable challenge for current models. More details are available in Appendix \ref{appendix:captioning_exp_complexity}.

\noindent\textbf{Enrichment of Visual Inputs.} To explore the challenges posed by complex videos, we further investigate the impact of increasing frame number on videos with varying complexity. As shown in Figure \ref{fig:comparison_nframes_complexity}, we analyze LLaVA-Video and Qwen2-VL, both trained with longer context lengths. Our findings reveal that F1 scores decrease with increasing video complexity at any given number of input frames. Generally, increasing the number of frames results in greater improvements for more complex samples, suggesting that complex videos require more frames for a complete and precise description. Counterintuitively, an unexpected pattern emerges: for the most complex videos, increasing frames from 32 to 64 actually reduces performance, indicating that highly complex videos remain a prominent challenge for LMMs. Further details be found in Appendix \ref{appendix:captioning_nframe}.

\begin{table}[!htbp]
  \centering
  \resizebox{0.48\textwidth}{!}{
    \begin{tabular}{lccc}
    \toprule
    Measure & Kendall's $\tau$ & Spearman's $\rho$ & Pearson $r$ \\
    \midrule
    METEOR \citep{banerjee-lavie-2005-meteor} & 30.8 & 44.8 & 54.7 \\
    BERT-Score \citep{zhang2019bertscore} & 27.4 & 34.8 & 49.2 \\
    CLAIR \citep{chan-etal-2023-clair} & 45.6 & 56.6 & 41.0 \\
    DREAM-1K \citep{wang2024tarsier} & 22.2 & 31.3 & 24.7 \\
    \midrule
    \benchcap & 57.2 & 76.7 & 69.9 \\
    \bottomrule
    \end{tabular}%
}
  \caption{Human judgment correlation scores for our automatic evaluation. All p-values $<0.05$.}
  \label{tab:human_consistency_cap}%
  \vspace{-1em}
\end{table}

\begin{table*}[!htbp]
  \centering
  \resizebox{\textwidth}{!}{
    \begin{tabular}{l|cc|cc|ccc|ccc|c}
    \toprule
    \multirow{2}[2]{*}{Model} & \multicolumn{2}{c|}{Camera State} & \multicolumn{2}{c|}{Background Scene} & \multicolumn{3}{c|}{Subject Action} & \multicolumn{3}{c|}{Object Attribute} & \multirow{2}[2]{*}{Overall} \\
    \cmidrule(lr){2-3} \cmidrule(lr){4-5} \cmidrule(lr){6-8} \cmidrule(lr){9-11}
    & Motion & Transition & Description & Transition & Recognition & Sequence & Matching & Recognition & Appearance & Location &  \\
    \midrule
    \rowcolor{gray!10}\multicolumn{12}{c}{\textit{\textbf{Open-Source LMMs}}}\\
    PLLaVA-7B        & 29.7 & 31.9 & 48.1 & 22.4 & 43.6 & 34.6 & 30.4 & 32.3 & 38.1 & 45.2 & 33.7 \\
    LongVA-7B        & 37.5 & 41.5 & 63.0 & 30.8 & 44.6 & 44.7 & 43.5 & 41.7 & 47.6 & 40.5 & 42.4 \\
    Tarsier-7B       & 23.0 & 24.6 & 40.7 & 20.6 & 38.6 & 26.9 & 45.7 & 20.9 & 25.9 & 23.8 & 26.5 \\
    Kangaroo         & 33.2 & 47.3 & 53.7 & 38.3 & 49.5 & 38.8 & 54.3 & 47.2 & 43.5 & 59.5 & 42.9 \\
    LLaVA-OV-7B      & 42.2 & 54.6 & 57.4 & 48.6 & 42.6 & 41.4 & 60.9 & 47.9 & 50.0 & 59.5 & 47.4 \\
    LLaVA-Video-7B   & 39.1 & 50.7 & 59.3 & 46.7 & 52.5 & 52.4 & 56.5 & 53.6 & 61.9 & 47.6 & 50.6 \\
    Qwen2-VL-7B      & 41.0 & 51.7 & 66.7 & 45.8 & 54.5 & 52.8 & 65.2 & 49.0 & 60.2 & 57.1 & 51.3 \\
    InternVL2-8B     & 41.0 & 53.1 & 66.7 & 40.2 & 45.5 & 50.5 & 50.0 & 45.8 & 56.8 & 45.2 & 48.4 \\
    MiniCPM-V-2.6    & 39.8 & 45.9 & 59.3 & 34.6 & 49.5 & 51.1 & 52.2 & 42.2 & 46.6 & 50.0 & 45.7 \\
    \midrule
    PLLaVA-34B       & 42.6 & 41.5 & 63.0 & 43.9 & 45.5 & 48.5 & 56.5 & 43.2 & 56.8 & 57.1 & 46.9 \\
    Tarsier-34B      & 43.0 & 48.3 & 72.2 & 45.8 & 51.5 & 50.2 & 56.5 & 49.7 & 53.7 & \second{61.9} & 50.1 \\
    \midrule
    LLaVA-OV-72B     & 46.5 & \best{67.6} & \second{75.9} & \second{57.0} & 59.4 & \second{56.6} & \best{73.9} & \best{63.5} & 69.5 & 59.5 & \second{60.0} \\
    LLaVA-Video-72B  & \second{47.7} & \best{67.6} & \best{77.8} & \best{61.7} & \second{61.4} & \best{57.0} & 65.2 & 62.5 & \second{73.7} & 57.1 & \best{60.7} \\
    Qwen2-VL-72B     & \best{52.7} & \second{64.7} & 74.1 & 55.1 & \best{62.4} & 54.4 & \second{67.4} & \second{63.0} & \best{76.3} & \best{66.7} & \best{60.7} \\
    InternVL2-76B    & 43.8 & 61.8 & 74.1 & 43.0 & 50.5 & 50.5 & 54.3 & 52.1 & 66.1 & 57.1 & 53.1 \\
    \midrule
    \rowcolor{gray!10}\multicolumn{12}{c}{\textit{\textbf{Closed-Source LMMs}}}\\
    Gemini 1.5 Flash & 40.8 & \second{58.3} & \second{70.4} & 52.3 & 48.0 & 54.2 & \second{63.0} & 49.0 & 66.7 & \second{64.3} & 53.3 \\
    Gemini 1.5 Pro   & \second{49.4} & \best{68.4} & 64.8 & \best{59.8} & \second{55.0} & \second{60.4} & \best{69.6} & \best{64.6} & \second{65.0} & \best{66.7} & \best{60.8} \\
    GPT-4o           & \best{53.9} & 56.0 & \best{81.5} & \second{56.1} & \best{59.4} & \best{67.6} & 58.7 & \second{56.8} & 63.6 & 59.5 & \second{60.3} \\
    \bottomrule
    \end{tabular}%
}
  \caption{\benchmcq performance of representative video LMMs. We provide detailed scores for selected tested models on 10 temporal tasks. The best and second-best results are marked with \best{bold} and \second{underline}, respectively.}
  \label{tab:main_results_mcq}%
  \vspace{-1.5em}
\end{table*}

\noindent\textbf{Correlation with Human Judgments.} To validate the effectiveness and robustness of our automatic evaluation method, we calculate Kendall's $\tau$, Spearman's $\rho$, and Pearson $r$ correlation scores between several methods and human evaluation. As shown in Table \ref{tab:human_consistency_cap}, these results demonstrate strong correlation, confirming that our method provides a robust and accurate solution for captioning evaluation. More details are available in Appendix \ref{appendix:captioning_correlation}.

% \vspace{-0.5em}
\subsection{Video QA}
\label{sec:experiments_mcq}

\benchmcq specializes in temporal understanding in videos, emphasizing the necessity of the entire video observation rather than single-frame analysis. We assess the temporal understanding skills across 4 dynamic elements and 10 task types.

\noindent\textbf{Overall Performance.} Table \ref{tab:main_results_mcq} showcases the performance of selected models on \benchmcq. All tested models demonstrate limited capabilities, with even the best-performing model barely achieving a passing score. However, a promising trend emerges as open-source models illustrate performance on par with commercial counterparts. Specifically, LLaVA-Video-72B and Qwen2-VL-72B achieve an identical score of 60.7\%, matching the performance of GPT-4o (60.3\%) and Gemini 1.5 Pro (60.8\%). This competitive performance of open-source models aligns with findings from recent studies, such as Video-MME (Short) \citep{fu2024videomme} and TempCompass \citep{liu2024tempcompass}, suggesting a promising direction for open-source development in video understanding.

\noindent\textbf{Camera State.} Recent works \citep{chai2024auroracap, tan2024vidgen1m} emphasize the crucial role of camera state in video understanding and generation. However, open-source video understanding datasets minimally involve this aspect. Our evaluation reveals a considerable weakness in models' camera understanding skill, with average scores notably lower than overall scores. While models show some promise in detecting camera transitions, they struggle particularly with camera motion analysis, achieving a maximum score of only 53.9\%. 

\noindent\textbf{Subject Action.} Action understanding is another challenge, as it requires tracking and interpreting character state evolutions across multiple frames. The action sequence task is notoriously difficult due to its complexity, demanding models to simultaneously recognize individual actions while understanding their temporal order and causal relationships. While GPT-4o leads performance with 67.6\% accuracy, all other models struggle to the passing threshold. Additionally, temporal action recognition remains challenging, with even the best-performing model achieving only 62.4\%.

\noindent\textbf{Background Scene \& Object Attribute.} Advanced video LMMs show promising capabilities in scene and attribute understanding. For background scene tasks, models achieve impressive results with GPT-4o reaching 81.5\% on scene description, while LLaVA-Video-72B attains 61.7\% on scene transition understanding. In object attribute tasks, models also perform well, with top scores of 64.6\% in recognition, 76.3\% in appearance, and 66.7\% in location tasks. These strong performance can be attributed to the transfer of knowledge from well-established image-text understanding techniques, as these tasks share similar characteristics with multi-image analysis scenarios.

These comprehensive results underscore the complex challenges in understanding temporal dynamics in videos, while offering clear directions for future improvements in video LMMs. 

\subsection{Synthesizing Analysis}

Through comprehensive analysis of \benchcap and \benchmcq results, commercial models demonstrate superior performance across both tasks. While open-source models (Qwen2-VL-72B and LLaVA-Video-72B) achieve comparable results on \benchmcq, they notably underperform in \benchcap. This performance gap reveals a critical limitation of open-source LMMs in captioning and even open-ended QA tasks, indicating areas demanding further research efforts.

\vspace{-0.5em}
\section{Conclusion}
\vspace{-0.5em}

In this paper, we present \dataset, a temporally dense video-caption dataset, and its derivative benchmark \bench. Our work focuses on temporal dynamics, the distinctive feature between videos and static images, by examining four critical temporal aspects: camera, scene, action, and attribute. \dataset features comprehensive coverage across diverse visual domains with detailed, fine-grained captions. \bench evaluates LMMs' temporal understanding skills through two complementary tasks: captioning and MCQ. This comprehensive evaluation provides precise insights into models' strengths and weaknesses, offering interpretable metrics for advancing video understanding technology. We envision \bench serving as a catalyst for future research in video understanding. Moreover, the meticulously annotated \dataset, with its high accuracy and completeness, offers versatile applications beyond our current scope. We anticipate its broad utility in diverse research directions and look forward to seeing its impact on future studies in the field.

\section*{Limitations}

Our dataset is highly fine-grained, but the data annotation is extremely labor-intensive. making it costly to apply this construction method to other video datasets. For \benchcap, we conduct a comprehensive evaluation of the video captioning capabilities of video LMMs using an interpretable and robust approach. However, our method has certain limitations. Our scoring system focuses on the alignment with annotated visual elements. If the model outputs visual elements that fail to match the annotated events or elements, our method cannot assess their precision. Specifically, when a generated caption includes excessive irrelevant content, even if this content contains substantial hallucinatory information, our method would be unable to provide a valid assessment in such cases.

\section*{Ethics Policy}

To increase the diversity of our dataset, we collected videos from several sources. These include a number of movies spanning many years and several types. While we made an effort to remove some videos that were poorly observed or NSFW, there may be unintentional data that involve potential social biases and stereotypes, including stereotypical items related to gender, race, ethnicity, age, and socioeconomic status. This requires careful judgment and utilization of the data.

% \section*{Acknowledgments}

% This work is supported by the National Natural Science Foundation of China (No. 62272092, No. 62172086).

\bibliography{custom, anthology}

\begin{thebibliography}{69}
\providecommand{\natexlab}[1]{#1}

\bibitem[{Amirloo et~al.(2024)Amirloo, Fauconnier, Roesmann, Kerl, Boney, Qian, Wang, Dehghan, Yang, Gan et~al.}]{amirloo2024understanding}
Elmira Amirloo, Jean-Philippe Fauconnier, Christoph Roesmann, Christian Kerl, Rinu Boney, Yusu Qian, Zirui Wang, Afshin Dehghan, Yinfei Yang, Zhe Gan, et~al. 2024.
\newblock Understanding alignment in multimodal llms: A comprehensive study.
\newblock \emph{arXiv preprint arXiv:2407.02477}.

\bibitem[{Arai et~al.(2024)Arai, Miwa, Sasaki, Yamaguchi, Watanabe, Aoki, and Yamamoto}]{arai2024covla}
Hidehisa Arai, Keita Miwa, Kento Sasaki, Yu~Yamaguchi, Kohei Watanabe, Shunsuke Aoki, and Issei Yamamoto. 2024.
\newblock Covla: Comprehensive vision-language-action dataset for autonomous driving.
\newblock \emph{arXiv preprint arXiv:2408.10845}.

\bibitem[{Banerjee and Lavie(2005)}]{banerjee-lavie-2005-meteor}
Satanjeev Banerjee and Alon Lavie. 2005.
\newblock \href {W05-0909} {{METEOR}: An automatic metric for {MT} evaluation with improved correlation with human judgments}.
\newblock pages 65--72, Ann Arbor, Michigan.

\bibitem[{Brooks et~al.(2024)Brooks, Peebles, Holmes, DePue, Guo, Jing, Schnurr, Taylor, Luhman, Luhman, Ng, Wang, and Ramesh}]{videoworldsimulators2024}
Tim Brooks, Bill Peebles, Connor Holmes, Will DePue, Yufei Guo, Li~Jing, David Schnurr, Joe Taylor, Troy Luhman, Eric Luhman, Clarence Ng, Ricky Wang, and Aditya Ramesh. 2024.
\newblock \href {https://openai.com/research/video-generation-models-as-world-simulators} {Video generation models as world simulators}.

\bibitem[{Caffagni et~al.(2024)Caffagni, Cocchi, Barsellotti, Moratelli, Sarto, Baraldi, Baraldi, Cornia, and Cucchiara}]{caffagni-etal-2024-revolution}
Davide Caffagni, Federico Cocchi, Luca Barsellotti, Nicholas Moratelli, Sara Sarto, Lorenzo Baraldi, Lorenzo Baraldi, Marcella Cornia, and Rita Cucchiara. 2024.
\newblock \href {https://doi.org/10.18653/v1/2024.findings-acl.807} {The revolution of multimodal large language models: A survey}.
\newblock pages 13590--13618, Bangkok, Thailand and virtual meeting.

\bibitem[{Cai et~al.(2024)Cai, Tan, Zhang, Zou, Zhang, Yao, Zhu, Gu, Zhong, Shang et~al.}]{cai2024temporalbench}
Mu~Cai, Reuben Tan, Jianrui Zhang, Bocheng Zou, Kai Zhang, Feng Yao, Fangrui Zhu, Jing Gu, Yiwu Zhong, Yuzhang Shang, et~al. 2024.
\newblock Temporalbench: Benchmarking fine-grained temporal understanding for multimodal video models.
\newblock \emph{arXiv preprint arXiv:2410.10818}.

\bibitem[{Chai et~al.(2024)Chai, Song, Du, Meng, Madhavan, Bar-Tal, Hwang, Xie, and Manning}]{chai2024auroracap}
Wenhao Chai, Enxin Song, Yilun Du, Chenlin Meng, Vashisht Madhavan, Omer Bar-Tal, Jeng-Neng Hwang, Saining Xie, and Christopher~D Manning. 2024.
\newblock Auroracap: Efficient, performant video detailed captioning and a new benchmark.
\newblock \emph{arXiv preprint arXiv:2410.03051}.

\bibitem[{Chan et~al.(2023)Chan, Petryk, Gonzalez, Darrell, and Canny}]{chan-etal-2023-clair}
David Chan, Suzanne Petryk, Joseph Gonzalez, Trevor Darrell, and John Canny. 2023.
\newblock \href {https://doi.org/10.18653/v1/2023.emnlp-main.841} {{CLAIR}: Evaluating image captions with large language models}.
\newblock pages 13638--13646, Singapore.

\bibitem[{Chen et~al.(2024{\natexlab{a}})Chen, Wei, Li, Dong, Zhang, Zang, Chen, Duan, Lin, Tang et~al.}]{chen2024sharegpt4video}
Lin Chen, Xilin Wei, Jinsong Li, Xiaoyi Dong, Pan Zhang, Yuhang Zang, Zehui Chen, Haodong Duan, Bin Lin, Zhenyu Tang, et~al. 2024{\natexlab{a}}.
\newblock Sharegpt4video: Improving video understanding and generation with better captions.
\newblock \emph{arXiv preprint arXiv:2406.04325}.

\bibitem[{Chen et~al.(2024{\natexlab{b}})Chen, Wang, Tian, Ye, Gao, Cui, Tong, Hu, Luo, Ma et~al.}]{chen2024internvl1.5}
Zhe Chen, Weiyun Wang, Hao Tian, Shenglong Ye, Zhangwei Gao, Erfei Cui, Wenwen Tong, Kongzhi Hu, Jiapeng Luo, Zheng Ma, et~al. 2024{\natexlab{b}}.
\newblock How far are we to gpt-4v? closing the gap to commercial multimodal models with open-source suites.
\newblock \emph{arXiv preprint arXiv:2404.16821}.

\bibitem[{Cheng et~al.(2024)Cheng, Leng, Zhang, Xin, Li, Chen, Zhu, Zhang, Luo, Zhao et~al.}]{cheng2024videollama2}
Zesen Cheng, Sicong Leng, Hang Zhang, Yifei Xin, Xin Li, Guanzheng Chen, Yongxin Zhu, Wenqi Zhang, Ziyang Luo, Deli Zhao, et~al. 2024.
\newblock Videollama 2: Advancing spatial-temporal modeling and audio understanding in video-llms.
\newblock \emph{arXiv preprint arXiv:2406.07476}.

\bibitem[{de~Souza~Inácio and Lopes(2023)}]{INACIO2023100488}
Andrei de~Souza~Inácio and Heitor~Silvério Lopes. 2023.
\newblock \href {https://doi.org/10.1016/j.mlwa.2023.100488} {Evaluation metrics for video captioning: A survey}.
\newblock \emph{Machine Learning with Applications}, 13:100488.

\bibitem[{Fang et~al.(2024)Fang, Mao, Duan, Zhao, Li, Lin, and Chen}]{fang2024mmbenchvideo}
Xinyu Fang, Kangrui Mao, Haodong Duan, Xiangyu Zhao, Yining Li, Dahua Lin, and Kai Chen. 2024.
\newblock Mmbench-video: A long-form multi-shot benchmark for holistic video understanding.
\newblock \emph{Advances in Neural Information Processing Systems}, 37:89098--89124.

\bibitem[{Fu et~al.(2024)Fu, Dai, Luo, Li, Ren, Zhang, Wang, Zhou, Shen, Zhang et~al.}]{fu2024videomme}
Chaoyou Fu, Yuhan Dai, Yondong Luo, Lei Li, Shuhuai Ren, Renrui Zhang, Zihan Wang, Chenyu Zhou, Yunhang Shen, Mengdan Zhang, et~al. 2024.
\newblock Video-mme: The first-ever comprehensive evaluation benchmark of multi-modal llms in video analysis.
\newblock \emph{arXiv preprint arXiv:2405.21075}.

\bibitem[{Ju et~al.(2024)Ju, Gao, Zhang, Yuan, Wang, Zeng, Xiong, Xu, and Shan}]{ju2024miradata}
Xuan Ju, Yiming Gao, Zhaoyang Zhang, Ziyang Yuan, Xintao Wang, Ailing Zeng, Yu~Xiong, Qiang Xu, and Ying Shan. 2024.
\newblock Miradata: A large-scale video dataset with long durations and structured captions.
\newblock \emph{arXiv preprint arXiv:2407.06358}.

\bibitem[{Kong et~al.(2025)Kong, Zhang, Liu, Zhang, Feng, Yang, Wang, Tian, W, Zhang, and Zhou}]{kong2025unite}
Fanheng Kong, Jingyuan Zhang, Yahui Liu, Hongzhi Zhang, Shi Feng, Xiaocui Yang, Daling Wang, Yu~Tian, Victoria W, Fuzheng Zhang, and Guorui Zhou. 2025.
\newblock Modality curation: Building universal embeddings for advanced multimodal information retrieval.
\newblock \emph{arXiv preprint arXiv:2505.19650}.

\bibitem[{Li et~al.(2024{\natexlab{a}})Li, Zhang, Guo, Zhang, Li, Zhang, Zhang, Li, Liu, and Li}]{li2024llavaov}
Bo~Li, Yuanhan Zhang, Dong Guo, Renrui Zhang, Feng Li, Hao Zhang, Kaichen Zhang, Yanwei Li, Ziwei Liu, and Chunyuan Li. 2024{\natexlab{a}}.
\newblock Llava-onevision: Easy visual task transfer.
\newblock \emph{arXiv preprint arXiv:2408.03326}.

\bibitem[{Li et~al.(2024{\natexlab{b}})Li, Gan, Yang, Yang, Li, Wang, Gao et~al.}]{li2024multimodal}
Chunyuan Li, Zhe Gan, Zhengyuan Yang, Jianwei Yang, Linjie Li, Lijuan Wang, Jianfeng Gao, et~al. 2024{\natexlab{b}}.
\newblock Multimodal foundation models: From specialists to general-purpose assistants.
\newblock \emph{Foundations and Trends{\textregistered} in Computer Graphics and Vision}, 16(1-2):1--214.

\bibitem[{Li et~al.(2024{\natexlab{c}})Li, Zhang, Zhang, Zhang, Li, Li, Ma, and Li}]{li2024llavanextinterleave}
Feng Li, Renrui Zhang, Hao Zhang, Yuanhan Zhang, Bo~Li, Wei Li, Zejun Ma, and Chunyuan Li. 2024{\natexlab{c}}.
\newblock Llava-next-interleave: Tackling multi-image, video, and 3d in large multimodal models.
\newblock \emph{arXiv preprint arXiv:2407.07895}.

\bibitem[{Li et~al.(2023)Li, Li, Savarese, and Hoi}]{li2023blip2}
Junnan Li, Dongxu Li, Silvio Savarese, and Steven Hoi. 2023.
\newblock Blip-2: Bootstrapping language-image pre-training with frozen image encoders and large language models.
\newblock In \emph{International conference on machine learning}, pages 19730--19742. PMLR.

\bibitem[{Li et~al.(2024{\natexlab{d}})Li, Wang, He, Li, Wang, Liu, Wang, Xu, Chen, Luo et~al.}]{li2024mvbench}
Kunchang Li, Yali Wang, Yinan He, Yizhuo Li, Yi~Wang, Yi~Liu, Zun Wang, Jilan Xu, Guo Chen, Ping Luo, et~al. 2024{\natexlab{d}}.
\newblock Mvbench: A comprehensive multi-modal video understanding benchmark.
\newblock In \emph{Proceedings of the IEEE/CVF Conference on Computer Vision and Pattern Recognition}, pages 22195--22206.

\bibitem[{Li et~al.(2024{\natexlab{e}})Li, Chen, Hu, Wang, Shi, and Zhang}]{li2024videovista}
Yunxin Li, Xinyu Chen, Baotian Hu, Longyue Wang, Haoyuan Shi, and Min Zhang. 2024{\natexlab{e}}.
\newblock Videovista: A versatile benchmark for video understanding and reasoning.
\newblock \emph{arXiv preprint arXiv:2406.11303}.

\bibitem[{Lin et~al.(2023)Lin, Ye, Zhu, Cui, Ning, Jin, and Yuan}]{lin2023videollava}
Bin Lin, Yang Ye, Bin Zhu, Jiaxi Cui, Munan Ning, Peng Jin, and Li~Yuan. 2023.
\newblock Video-llava: Learning united visual representation by alignment before projection.
\newblock \emph{arXiv preprint arXiv:2311.10122}.

\bibitem[{Lin(2004)}]{lin-2004-rouge}
Chin-Yew Lin. 2004.
\newblock \href {W04-1013} {{ROUGE}: A package for automatic evaluation of summaries}.
\newblock pages 74--81, Barcelona, Spain.

\bibitem[{Lin et~al.(2024)Lin, Yin, Ping, Molchanov, Shoeybi, and Han}]{lin2024vila}
Ji~Lin, Hongxu Yin, Wei Ping, Pavlo Molchanov, Mohammad Shoeybi, and Song Han. 2024.
\newblock Vila: On pre-training for visual language models.
\newblock In \emph{Proceedings of the IEEE/CVF Conference on Computer Vision and Pattern Recognition}, pages 26689--26699.

\bibitem[{Liu et~al.(2024{\natexlab{a}})Liu, Li, Li, and Lee}]{liu2024llava1.5}
Haotian Liu, Chunyuan Li, Yuheng Li, and Yong~Jae Lee. 2024{\natexlab{a}}.
\newblock Improved baselines with visual instruction tuning.
\newblock In \emph{Proceedings of the IEEE/CVF Conference on Computer Vision and Pattern Recognition}, pages 26296--26306.

\bibitem[{Liu et~al.(2024{\natexlab{b}})Liu, Li, Wu, and Lee}]{liu2024llava}
Haotian Liu, Chunyuan Li, Qingyang Wu, and Yong~Jae Lee. 2024{\natexlab{b}}.
\newblock Visual instruction tuning.
\newblock \emph{Advances in neural information processing systems}, 36.

\bibitem[{Liu et~al.(2024{\natexlab{c}})Liu, Wang, Ma, Wu, Ma, Wei, Jiao, Wu, and Hu}]{liu2024kangaroo}
Jiajun Liu, Yibing Wang, Hanghang Ma, Xiaoping Wu, Xiaoqi Ma, Xiaoming Wei, Jianbin Jiao, Enhua Wu, and Jie Hu. 2024{\natexlab{c}}.
\newblock Kangaroo: A powerful video-language model supporting long-context video input.
\newblock \emph{arXiv preprint arXiv:2408.15542}.

\bibitem[{Liu et~al.(2025)Liu, Cheng, Liu, Zhang, Li, Ren, Zou, Yang, Su, Zhu et~al.}]{liu2025llavaplus}
Shilong Liu, Hao Cheng, Haotian Liu, Hao Zhang, Feng Li, Tianhe Ren, Xueyan Zou, Jianwei Yang, Hang Su, Jun Zhu, et~al. 2025.
\newblock Llava-plus: Learning to use tools for creating multimodal agents.
\newblock In \emph{European Conference on Computer Vision}, pages 126--142. Springer.

\bibitem[{Liu et~al.(2024{\natexlab{d}})Liu, Tao, Liu, Fang, Zhou, Huang, He, and Yang}]{liu-etal-2024-devan}
Tingkai Liu, Yunzhe Tao, Haogeng Liu, Qihang Fang, Ding Zhou, Huaibo Huang, Ran He, and Hongxia Yang. 2024{\natexlab{d}}.
\newblock \href {https://doi.org/10.18653/v1/2024.acl-long.772} {{D}e{VA}n: Dense video annotation for video-language models}.
\newblock pages 14305--14321, Bangkok, Thailand.

\bibitem[{Liu et~al.(2024{\natexlab{e}})Liu, Ma, Qi, Wu, Chen, and Shan}]{liu2024etbench}
Ye~Liu, Zongyang Ma, Zhongang Qi, Yang Wu, Chang~Wen Chen, and Ying Shan. 2024{\natexlab{e}}.
\newblock E.t. bench: Towards open-ended event-level video-language understanding.
\newblock In \emph{Neural Information Processing Systems (NeurIPS)}.

\bibitem[{Liu et~al.(2024{\natexlab{f}})Liu, Li, Liu, Wang, Ren, Li, Chen, Sun, and Hou}]{liu2024tempcompass}
Yuanxin Liu, Shicheng Li, Yi~Liu, Yuxiang Wang, Shuhuai Ren, Lei Li, Sishuo Chen, Xu~Sun, and Lu~Hou. 2024{\natexlab{f}}.
\newblock Tempcompass: Do video llms really understand videos?
\newblock \emph{arXiv preprint arXiv:2403.00476}.

\bibitem[{Maaz et~al.(2023)Maaz, Rasheed, Khan, and Khan}]{maaz2023videochatgpt}
Muhammad Maaz, Hanoona Rasheed, Salman Khan, and Fahad~Shahbaz Khan. 2023.
\newblock Video-chatgpt: Towards detailed video understanding via large vision and language models.
\newblock \emph{arXiv preprint arXiv:2306.05424}.

\bibitem[{Madan et~al.(2024)Madan, M{\o}gelmose, Modi, Rawat, and Moeslund}]{madan2024foundation}
Neelu Madan, Andreas M{\o}gelmose, Rajat Modi, Yogesh~S Rawat, and Thomas~B Moeslund. 2024.
\newblock Foundation models for video understanding: A survey.
\newblock \emph{arXiv preprint arXiv:2405.03770}.

\bibitem[{Mangalam et~al.(2023)Mangalam, Akshulakov, and Malik}]{mangalam2023egoschema}
Karttikeya Mangalam, Raiymbek Akshulakov, and Jitendra Malik. 2023.
\newblock Egoschema: A diagnostic benchmark for very long-form video language understanding.
\newblock \emph{Advances in Neural Information Processing Systems}, 36:46212--46244.

\bibitem[{mixkit(2023)}]{mixkit}
mixkit. 2023.
\newblock mixkit.
\newblock \url{https://mixkit.com/videos/}.

\bibitem[{Nguyen et~al.(2024)Nguyen, Bin, Xiao, Qu, Li, Wu, Nguyen, Ng, and Tuan}]{nguyen2024video}
Thong Nguyen, Yi~Bin, Junbin Xiao, Leigang Qu, Yicong Li, Jay~Zhangjie Wu, Cong-Duy Nguyen, See-Kiong Ng, and Luu~Anh Tuan. 2024.
\newblock Video-language understanding: A survey from model architecture, model training, and data perspectives.
\newblock \emph{arXiv preprint arXiv:2406.05615}.

\bibitem[{OpenAI(2024)}]{openai2024gpt4o}
OpenAI. 2024.
\newblock \href {https://openai.com/index/hello-gpt-4o/} {Hello gpt-4o}.

\bibitem[{Pan et~al.(2023)Pan, Dong, Huang, Peng, Chen, and Wei}]{pan2023kosmosg}
Xichen Pan, Li~Dong, Shaohan Huang, Zhiliang Peng, Wenhu Chen, and Furu Wei. 2023.
\newblock Kosmos-g: Generating images in context with multimodal large language models.
\newblock \emph{arXiv preprint arXiv:2310.02992}.

\bibitem[{Papineni et~al.(2002)Papineni, Roukos, Ward, and Zhu}]{papineni-etal-2002-bleu}
Kishore Papineni, Salim Roukos, Todd Ward, and Wei-Jing Zhu. 2002.
\newblock \href {https://doi.org/10.3115/1073083.1073135} {{B}leu: a method for automatic evaluation of machine translation}.
\newblock pages 311--318, Philadelphia, Pennsylvania, USA.

\bibitem[{Patraucean et~al.(2024)Patraucean, Smaira, Gupta, Recasens, Markeeva, Banarse, Koppula, Malinowski, Yang, Doersch et~al.}]{patraucean2024perceptiontest}
Viorica Patraucean, Lucas Smaira, Ankush Gupta, Adria Recasens, Larisa Markeeva, Dylan Banarse, Skanda Koppula, Mateusz Malinowski, Yi~Yang, Carl Doersch, et~al. 2024.
\newblock Perception test: A diagnostic benchmark for multimodal video models.
\newblock \emph{Advances in Neural Information Processing Systems}, 36.

\bibitem[{Pexels(2023)}]{pexels}
Pexels. 2023.
\newblock Pexels.
\newblock \url{https://www.pexels.com/videos/}.

\bibitem[{pixabay(2023)}]{pixabay}
pixabay. 2023.
\newblock pixabay.
\newblock \url{https://pixabay.com/videos/}.

\bibitem[{Polyak et~al.(2024)Polyak, Zohar, Brown, Tjandra, Sinha, Lee, Vyas, Shi, Ma, Chuang et~al.}]{polyak2024moviegen}
Adam Polyak, Amit Zohar, Andrew Brown, Andros Tjandra, Animesh Sinha, Ann Lee, Apoorv Vyas, Bowen Shi, Chih-Yao Ma, Ching-Yao Chuang, et~al. 2024.
\newblock Movie gen: A cast of media foundation models.
\newblock \emph{arXiv preprint arXiv:2410.13720}.

\bibitem[{Reid et~al.(2024)Reid, Savinov, Teplyashin, Lepikhin, Lillicrap, Alayrac, Soricut, Lazaridou, Firat, Schrittwieser et~al.}]{reid2024gemini1.5}
Machel Reid, Nikolay Savinov, Denis Teplyashin, Dmitry Lepikhin, Timothy Lillicrap, Jean-baptiste Alayrac, Radu Soricut, Angeliki Lazaridou, Orhan Firat, Julian Schrittwieser, et~al. 2024.
\newblock Gemini 1.5: Unlocking multimodal understanding across millions of tokens of context.
\newblock \emph{arXiv preprint arXiv:2403.05530}.

\bibitem[{Saravanan et~al.(2024)Saravanan, Singh, Gupta, Khan, Gandhi, and Tapaswi}]{saravanan2024velociti}
Darshana Saravanan, Darshan Singh, Varun Gupta, Zeeshan Khan, Vineet Gandhi, and Makarand Tapaswi. 2024.
\newblock Velociti: Can video-language models bind semantic concepts through time?
\newblock \emph{arXiv preprint arXiv:2406.10889}.

\bibitem[{Shangguan et~al.(2024)Shangguan, Li, Ding, Zheng, Zhao, Fitzgerald, and Cohan}]{shangguan2024tomato}
Ziyao Shangguan, Chuhan Li, Yuxuan Ding, Yanan Zheng, Yilun Zhao, Tesca Fitzgerald, and Arman Cohan. 2024.
\newblock Tomato: Assessing visual temporal reasoning capabilities in multimodal foundation models.
\newblock \emph{arXiv preprint arXiv:2410.23266}.

\bibitem[{Tan et~al.(2024)Tan, Yang, Qin, and Li}]{tan2024vidgen1m}
Zhiyu Tan, Xiaomeng Yang, Luozheng Qin, and Hao Li. 2024.
\newblock Vidgen-1m: A large-scale dataset for text-to-video generation.
\newblock \emph{arXiv preprint arXiv:2408.02629}.

\bibitem[{Tang et~al.(2023)Tang, Bi, Xu, Song, Liang, Wang, Zhang, An, Lin, Zhu et~al.}]{tang2023video}
Yunlong Tang, Jing Bi, Siting Xu, Luchuan Song, Susan Liang, Teng Wang, Daoan Zhang, Jie An, Jingyang Lin, Rongyi Zhu, et~al. 2023.
\newblock Video understanding with large language models: A survey.
\newblock \emph{arXiv preprint arXiv:2312.17432}.

\bibitem[{Vedantam et~al.(2015)Vedantam, Lawrence~Zitnick, and Parikh}]{vedantam2015cider}
Ramakrishna Vedantam, C~Lawrence~Zitnick, and Devi Parikh. 2015.
\newblock Cider: Consensus-based image description evaluation.
\newblock In \emph{Proceedings of the IEEE conference on computer vision and pattern recognition}, pages 4566--4575.

\bibitem[{Wang et~al.(2024{\natexlab{a}})Wang, Yuan, and Zhang}]{wang2024tarsier}
Jiawei Wang, Liping Yuan, and Yuchen Zhang. 2024{\natexlab{a}}.
\newblock Tarsier: Recipes for training and evaluating large video description models.
\newblock \emph{arXiv preprint arXiv:2407.00634}.

\bibitem[{Wang et~al.(2024{\natexlab{b}})Wang, Bai, Tan, Wang, Fan, Bai, Chen, Liu, Wang, Ge et~al.}]{wang2024qwen2vl}
Peng Wang, Shuai Bai, Sinan Tan, Shijie Wang, Zhihao Fan, Jinze Bai, Keqin Chen, Xuejing Liu, Jialin Wang, Wenbin Ge, et~al. 2024{\natexlab{b}}.
\newblock Qwen2-vl: Enhancing vision-language model's perception of the world at any resolution.
\newblock \emph{arXiv preprint arXiv:2409.12191}.

\bibitem[{Wang et~al.(2022)Wang, Gao, Yu, Lei, Feiszli, and Shou}]{wang2022geb+}
Yuxuan Wang, Difei Gao, Licheng Yu, Weixian Lei, Matt Feiszli, and Mike~Zheng Shou. 2022.
\newblock Geb+: A benchmark for generic event boundary captioning, grounding and retrieval.
\newblock In \emph{European Conference on Computer Vision}, pages 709--725. Springer.

\bibitem[{Xiao et~al.(2021)Xiao, Shang, Yao, and Chua}]{xiao2021nextqa}
Junbin Xiao, Xindi Shang, Angela Yao, and Tat-Seng Chua. 2021.
\newblock Next-qa: Next phase of question-answering to explaining temporal actions.
\newblock In \emph{Proceedings of the IEEE/CVF conference on computer vision and pattern recognition}, pages 9777--9786.

\bibitem[{Xiong et~al.(2024)Xiong, Wang, Zhou, Lin, Feng, and Liu}]{xiong2024lvd2m}
Tianwei Xiong, Yuqing Wang, Daquan Zhou, Zhijie Lin, Jiashi Feng, and Xihui Liu. 2024.
\newblock Lvd-2m: A long-take video dataset with temporally dense captions.
\newblock \emph{arXiv preprint arXiv:2410.10816}.

\bibitem[{Xu et~al.(2024)Xu, Zhao, Zhou, Lin, Ng, and Feng}]{xu2024pllava}
Lin Xu, Yilin Zhao, Daquan Zhou, Zhijie Lin, See~Kiong Ng, and Jiashi Feng. 2024.
\newblock Pllava: Parameter-free llava extension from images to videos for video dense captioning.
\newblock \emph{arXiv preprint arXiv:2404.16994}.

\bibitem[{Yao et~al.(2024)Yao, Yu, Zhang, Wang, Cui, Zhu, Cai, Li, Zhao, He et~al.}]{yao2024minicpmv}
Yuan Yao, Tianyu Yu, Ao~Zhang, Chongyi Wang, Junbo Cui, Hongji Zhu, Tianchi Cai, Haoyu Li, Weilin Zhao, Zhihui He, et~al. 2024.
\newblock Minicpm-v: A gpt-4v level mllm on your phone.
\newblock \emph{arXiv preprint arXiv:2408.01800}.

\bibitem[{Zhang et~al.(2024{\natexlab{a}})Zhang, Yu, Li, Dong, Su, Chu, and Yu}]{zhang2024mm}
Duzhen Zhang, Yahan Yu, Chenxing Li, Jiahua Dong, Dan Su, Chenhui Chu, and Dong Yu. 2024{\natexlab{a}}.
\newblock Mm-llms: Recent advances in multimodal large language models.
\newblock \emph{arXiv preprint arXiv:2401.13601}.

\bibitem[{Zhang et~al.(2025)Zhang, Zhang, Haonan, Sun, Wang, Kong, Liu, Wang, Zhang et~al.}]{zhang2025data}
Jingyuan Zhang, Hongzhi Zhang, Zhou Haonan, Chenxi Sun, Jiakang Wang, Fanheng Kong, Yahui Liu, Qi~Wang, Fuzheng Zhang, et~al. 2025.
\newblock Data metabolism: An efficient data design schema for vision language model.
\newblock \emph{arXiv preprint arXiv:2504.12316}.

\bibitem[{Zhang et~al.(2024{\natexlab{b}})Zhang, Dong, Zang, Cao, Qian, Chen, Guo, Duan, Wang, Ouyang et~al.}]{zhang2024internlmxcomposer2.5}
Pan Zhang, Xiaoyi Dong, Yuhang Zang, Yuhang Cao, Rui Qian, Lin Chen, Qipeng Guo, Haodong Duan, Bin Wang, Linke Ouyang, et~al. 2024{\natexlab{b}}.
\newblock Internlm-xcomposer-2.5: A versatile large vision language model supporting long-contextual input and output.
\newblock \emph{arXiv preprint arXiv:2407.03320}.

\bibitem[{Zhang et~al.(2024{\natexlab{c}})Zhang, Zhang, Li, Zeng, Yang, Zhang, Wang, Tan, Li, and Liu}]{zhang2024longva}
Peiyuan Zhang, Kaichen Zhang, Bo~Li, Guangtao Zeng, Jingkang Yang, Yuanhan Zhang, Ziyue Wang, Haoran Tan, Chunyuan Li, and Ziwei Liu. 2024{\natexlab{c}}.
\newblock Long context transfer from language to vision.
\newblock \emph{arXiv preprint arXiv:2406.16852}.

\bibitem[{Zhang et~al.(2024{\natexlab{d}})Zhang, Gui, Sun, Feng, Xu, Zhang, Fu, Li, Hauptmann, Bisk et~al.}]{zhang2024llavahound}
Ruohong Zhang, Liangke Gui, Zhiqing Sun, Yihao Feng, Keyang Xu, Yuanhan Zhang, Di~Fu, Chunyuan Li, Alexander Hauptmann, Yonatan Bisk, et~al. 2024{\natexlab{d}}.
\newblock Direct preference optimization of video large multimodal models from language model reward.
\newblock \emph{arXiv preprint arXiv:2404.01258}.

\bibitem[{Zhang et~al.(2019)Zhang, Kishore, Wu, Weinberger, and Artzi}]{zhang2019bertscore}
Tianyi Zhang, Varsha Kishore, Felix Wu, Kilian~Q Weinberger, and Yoav Artzi. 2019.
\newblock Bertscore: Evaluating text generation with bert.
\newblock \emph{arXiv preprint arXiv:1904.09675}.

\bibitem[{Zhang et~al.(2024{\natexlab{e}})Zhang, Kong, Wang, Sun, SWangLing, Feng, Wang, Zhang, and Song}]{zhang-etal-2024-stickerconv}
Yiqun Zhang, Fanheng Kong, Peidong Wang, Shuang Sun, SWangLing SWangLing, Shi Feng, Daling Wang, Yifei Zhang, and Kaisong Song. 2024{\natexlab{e}}.
\newblock \href {https://doi.org/10.18653/v1/2024.acl-long.417} {{STICKERCONV}: Generating multimodal empathetic responses from scratch}.
\newblock pages 7707--7733, Bangkok, Thailand.

\bibitem[{Zhang et~al.(2024{\natexlab{f}})Zhang, Wu, Li, Li, Ma, Liu, and Li}]{zhang2024llavavideo}
Yuanhan Zhang, Jinming Wu, Wei Li, Bo~Li, Zejun Ma, Ziwei Liu, and Chunyuan Li. 2024{\natexlab{f}}.
\newblock Video instruction tuning with synthetic data.
\newblock \emph{arXiv preprint arXiv:2410.02713}.

\bibitem[{Zhang et~al.(2023)Zhang, Zhang, Li, Zhao, Karypis, and Smola}]{zhang2023multimodal}
Zhuosheng Zhang, Aston Zhang, Mu~Li, Hai Zhao, George Karypis, and Alex Smola. 2023.
\newblock Multimodal chain-of-thought reasoning in language models.
\newblock \emph{arXiv preprint arXiv:2302.00923}.

\bibitem[{Zhou et~al.(2024)Zhou, Shu, Zhao, Wu, Xiao, Yang, Xiong, Zhang, Huang, and Liu}]{zhou2024mlvu}
Junjie Zhou, Yan Shu, Bo~Zhao, Boya Wu, Shitao Xiao, Xi~Yang, Yongping Xiong, Bo~Zhang, Tiejun Huang, and Zheng Liu. 2024.
\newblock Mlvu: A comprehensive benchmark for multi-task long video understanding.
\newblock \emph{arXiv preprint arXiv:2406.04264}.

\bibitem[{Zhou et~al.(2018)Zhou, Xu, and Corso}]{zhou2018youcook2}
Luowei Zhou, Chenliang Xu, and Jason Corso. 2018.
\newblock Towards automatic learning of procedures from web instructional videos.
\newblock In \emph{Proceedings of the AAAI Conference on Artificial Intelligence}, volume~32.

\bibitem[{Zhu et~al.(2023)Zhu, Chen, Shen, Li, and Elhoseiny}]{zhu2023minigpt4}
Deyao Zhu, Jun Chen, Xiaoqian Shen, Xiang Li, and Mohamed Elhoseiny. 2023.
\newblock Minigpt-4: Enhancing vision-language understanding with advanced large language models.
\newblock \emph{arXiv preprint arXiv:2304.10592}.

\end{thebibliography}

\clearpage

\appendix

\section{\dataset}
\label{appendix:dataset}

\subsection{Statistics}
\label{appendix:dataset_statistics}

\begin{figure}[!htbp]
    \centering
    \includegraphics[width=0.7\linewidth]{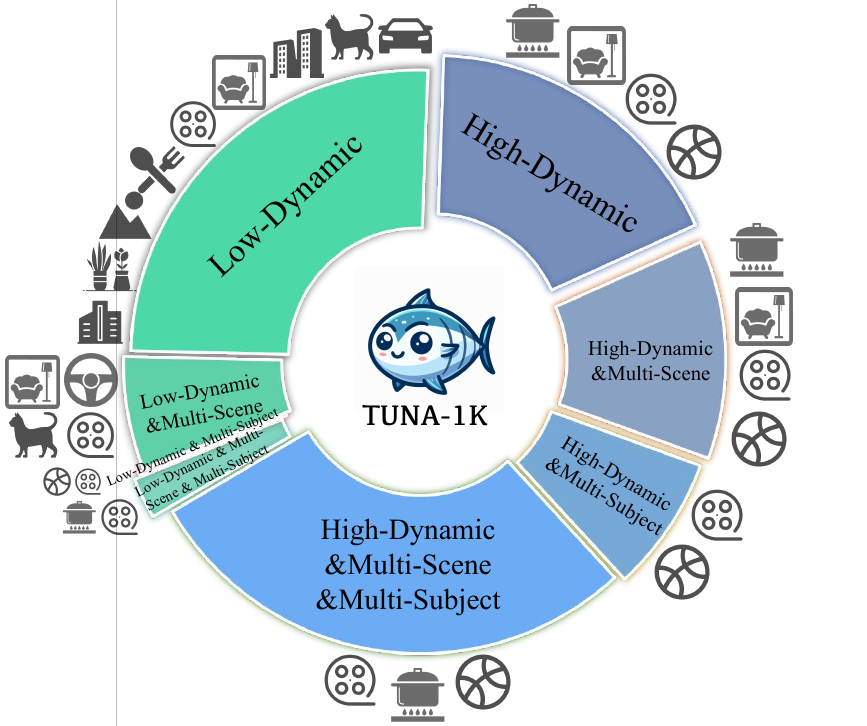}
    \caption{The sample distribution of \dataset, videos covering 4 visual characteristics and 12 domains.}
    \label{fig:category}
\end{figure}

As shown in Table \ref{tab:statistics_dataset}, we illustrate the detailed statistics of \dataset. Each video must belong to one of Low-Dynamic or High-Dynamic categories, while Multi-Scene and Multi-Subject are optional. 

\begin{figure}[!htbp]
    \centering
    \includegraphics[width=\linewidth]{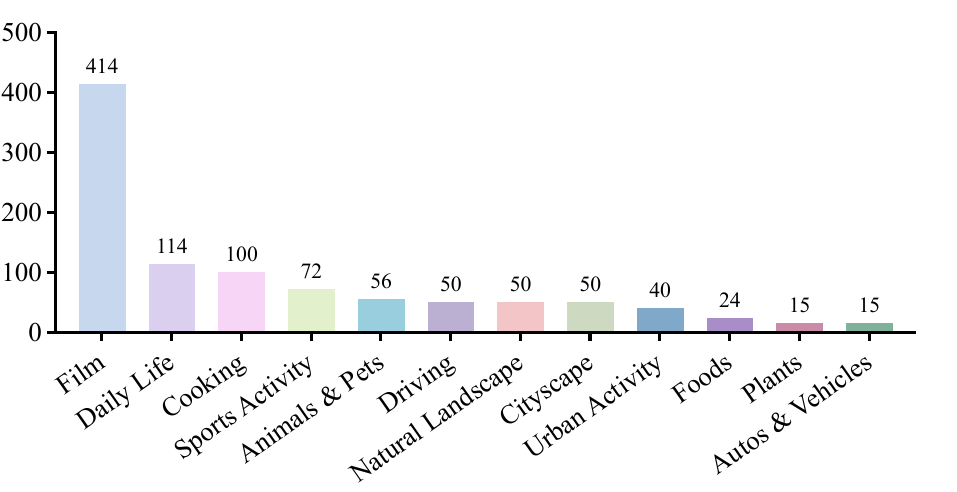}
    \caption{Sample distribution of domains in the \dataset, covering 12 domains.}
    \label{fig:statistics_domain}
\end{figure}

There are 12 domains contained in TUNA-1K, including: (1) Film, (2) Daily Life, (3) Cooking, (4) Sports Activity, (5) Driving, (6) Animals \& Pets, (7) Natural Landscape, (8) Cityscape, (9) Urban Activity, (10) Foods, (11) Plants, and (12) Autos \& Vehicles. As shown in Figure \ref{fig:statistics_domain}, we illustrate the domain statistics of the videos in \dataset.

\begin{figure*}[!htbp]
    \centering
    \includegraphics[width=0.9\linewidth]{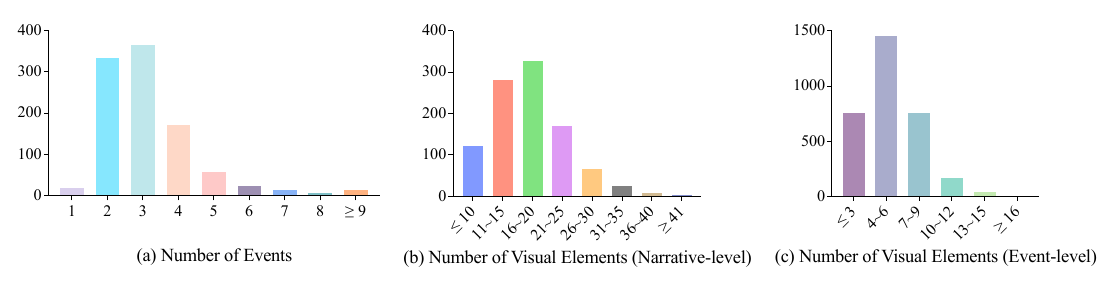}
    \setlength{\abovecaptionskip}{-0.2em}
    \caption{Visual statistics of the number of events and the number of visual elements in \dataset.}
    \label{fig:statistics_events}
    \vspace{-0.5em}
\end{figure*}

We visualize the sample distribution of video complexity in \dataset in Figure \ref{fig:statistics_events}, in terms of (a) the number of events, (b) the number of visual elements in each video, and (c) the number of visual elements in each event.

\begin{table*}[!htbp]
  \centering
  \resizebox{0.8\textwidth}{!}{
    \begin{tabular}{lccccc}
    \toprule
    & Low-Dynamic & High-Dynamic & Multi-Scene & Multi-Subject & Total \\
    \midrule
    \textbf{\#Videos} & 340 & 660 & 493 & 385 & 1,000 \\
    \textbf{Duration} & 18.0s & 12.8s & 12.5s & 9.5s & 14.53s \\
    \textbf{\#Events} & 2.8 & 3.4 & 3.8 & 3.8 & 3.2 \\
    \textbf{\#Elements (Narrative-level)} & 15.8 & 18.3 & 19.9 & 20.2 & 17.5 \\
    \textbf{\#Elements (Event-level)} & 5.7 & 5.4 & 5.3 & 5.3 & 5.48 \\
    \textbf{\#Tokens} & 198.8 & 247.1 & 255.9 & 267.6 & 230.7 \\
    \bottomrule
    \end{tabular}%
}
  \caption{Detailed statistics for \dataset, including: number of videos (\textbf{\#Videos}), video duration (\textbf{Duration}), number of events (\textbf{\#Events}), number of visual elements in captions (\textbf{\#Elements (Narrative-level)}), number of visual elements in events (\textbf{\#Elements (Narrative-level)}), number of tokens of caption (\textbf{\#Tokens}).}
  \label{tab:statistics_dataset}%
\end{table*}

\subsection{More Details of \dataset Construction}

\subsubsection{Video Collection}
\label{appendix:dataset_video_collection}

Table \ref{tab:detailed_video_source} shows the video sources that make up the TUNA-1K, along with their descriptions.

\subsubsection{Annotators}
\label{appendix:dataset_annotators}

We employ crowdsourcing for data annotation. All annotators have TEM-4 or TEM-8 English proficiency, and have experience in video captioning annotation (e.g., several annotators have previously annotated video-caption pairs for Kling\footnote{https://kling.kuaishou.com} project). Prior to formal annotation, they undergo our specialized training to guarantee the quality of their annotation.

\subsubsection{Annotator Training}
\label{appendix:dataset_annotator_training}

We prepare a detailed note document for instructing human annotators on annotation. The documentation encompasses 5 key components: (1) Visual Characteristic Classification, (2) Video Element Guidelines, (3) Video Captioning Protocol, (4) Event Splitting and Element Extraction Criterion, and (5) Annotation Examples.

\noindent\textbf{Visual Characteristic Classification.} Detailed criteria for categorizing videos based on their visual characteristics. 
\begin{itemize}[leftmargin=*, nolistsep, noitemsep]
\item \textbf{Low/High-Dynamic:} Based on the number and frequency of dynamic elements in the video.
\item \textbf{Multi-Scene:} Presence of at least one camera transition or scene transition. Excludes those that just have camera zooming, panning, or rotating.
\item \textbf{Multi-Subject:} Presence of at least two subjects. Non-major objects are not counted.
\end{itemize}

\noindent\textbf{Video Element Guidelines.} Comprehensive definitions and key considerations for essential video elements:
\begin{itemize}[leftmargin=*, nolistsep, noitemsep]
\item \textbf{Camera:} Camera states, including panning, rotating, zooming, following, shaking, transition, etc. It is necessary to indicate a specific direction.
\item \textbf{Scene:} Describe the background scene, including environment, weather, time, etc.
\item \textbf{Action:} Recognize actions and their temporal evolving sequences.
\item \textbf{Attribute:} Identify objects and describe their appearance (e.g., characters' gender, age, and dress, objects' color, shape, and number) and spatial orientation (location and relative positional relationships).
\end{itemize}

\noindent\textbf{Video Captioning Protocol.} We emphasizing:
\begin{itemize}[leftmargin=*, nolistsep, noitemsep]
\item Strict chronological ordering of events.
\item Objective descriptions without summarization and subjective feelings.
\item If multiple similar characters/objects appear, distinguish them in expression by unique attributes (e.g., age, dress, etc.).
\end{itemize}

\noindent\textbf{Event Splitting and Element Extraction Criterion.} To ensure systematic and standardized annotation, we establish the following comprehensive guidelines:
\begin{itemize}[leftmargin=*, nolistsep, noitemsep]
\item Divide captions into chronologically ordered events, where each event represents distinct temporal activities. Further decompose each event into its constituent visual elements.
\item Ensure explicit subject identification in all visual elements. Replace missing subjects and pronouns with their corresponding specific noun references to maintain clarity and precision.
\item Element weighting criteria for scoring: (1) Weight 3: primary and conspicuous contents in the video. (2) Weight 2: primary and inconspicuous contents, or secondary but conspicuous contents. (3) Weight 1: secondary and inconspicuous contents.
\end{itemize}

\noindent\textbf{Annotation Examples.} Some complete annotation examples provide human annotators with a further guidance for annotation.

To ensure annotation quality and consistency, we implemented a rigorous annotator selection and training process. Initially, all potential annotators underwent a trial annotation phase using a shared subset of videos. This phase served both as a training exercise and a qualification assessment. Through careful evaluation of their trial annotations, we selected only those annotators who demonstrated high consistency, accuracy, and thorough understanding of the annotation guidelines. These qualified annotators then proceeded to participate in the main annotation task. This systematic approach helped maintain annotation quality while minimizing potential inconsistencies across different annotators.

\subsubsection{Annotation}
\label{appendix:dataset_annotation}

\noindent\textbf{Video Filter.} We first filter out undesired videos based on specific rules, e.g., videos with low resolution (not satisfying 480p) and long duration (>40s). Then, human annotators filter out near-static or NSFW videos to ensure the high quality and temporal dynamics of the selected videos.

\begin{figure*}[!htbp]
    \centering
    \includegraphics[width=\linewidth]{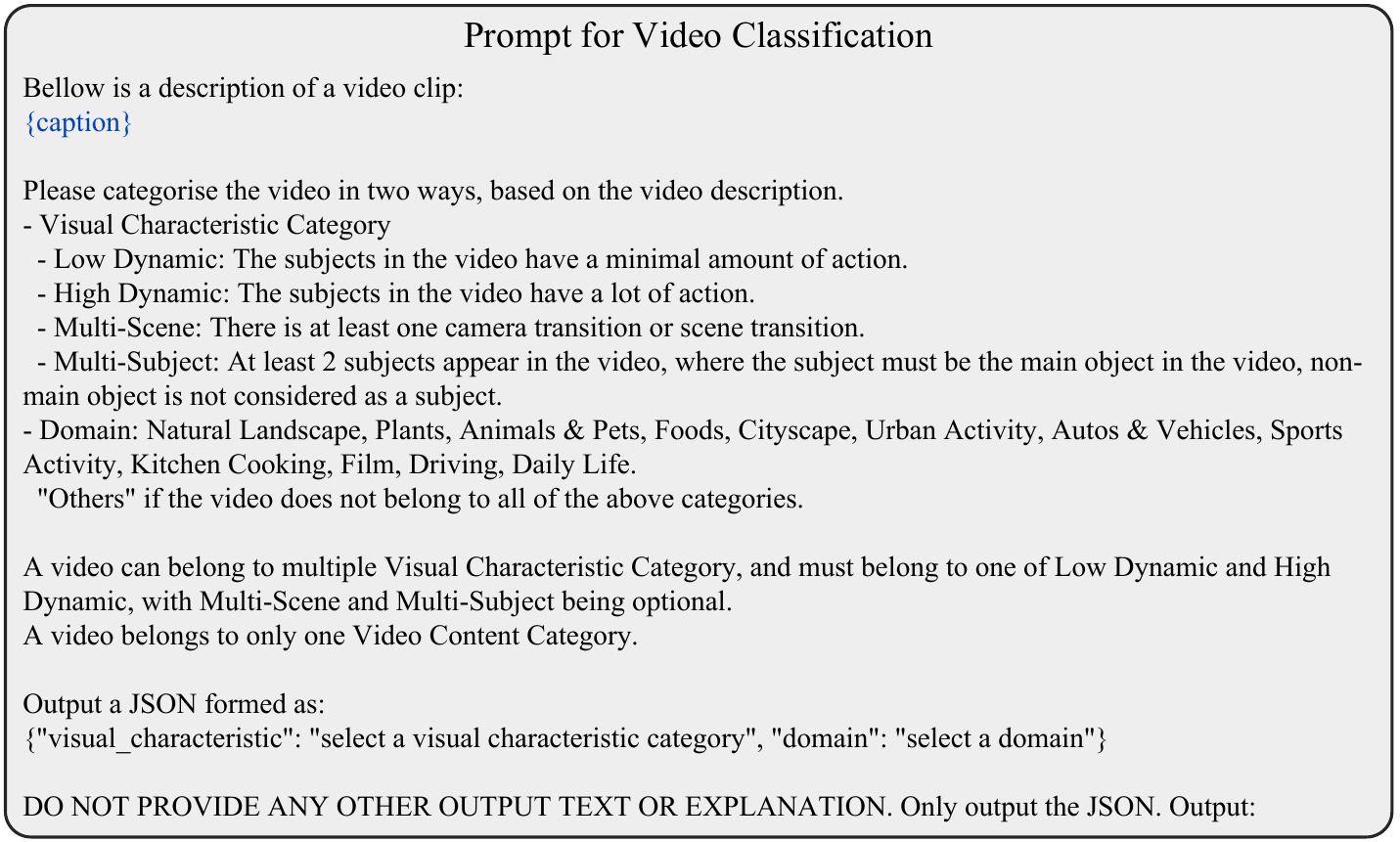}
    \caption{The prompt used to classify videos for the \benchcap construction in Section \ref{sec:method_captioning}.}
    \label{fig:prompt_video_classification}
\end{figure*}

\noindent\textbf{Video Cluster.} Firstly, we assign a caption for each video. If the original source provides a caption, it is utilized; otherwise, a caption is generated using \texttt{gpt-4o-2024-05-13}. Then, we utilize GPT-4o to classify visual characteristic category and domain for each video. Thus far, we have obtained raw videos with initial model-generated visual characteristics and domains. Initially, we obtain raw videos with model-generated visual characteristics and domains. Annotators then observe the videos, correcting and supplementing the visual characteristic categories and domains as needed. The prompt instruction used in this step is shown in Figure \ref{fig:prompt_video_classification}.

\begin{figure*}[t]
    \centering
    \includegraphics[width=\linewidth]{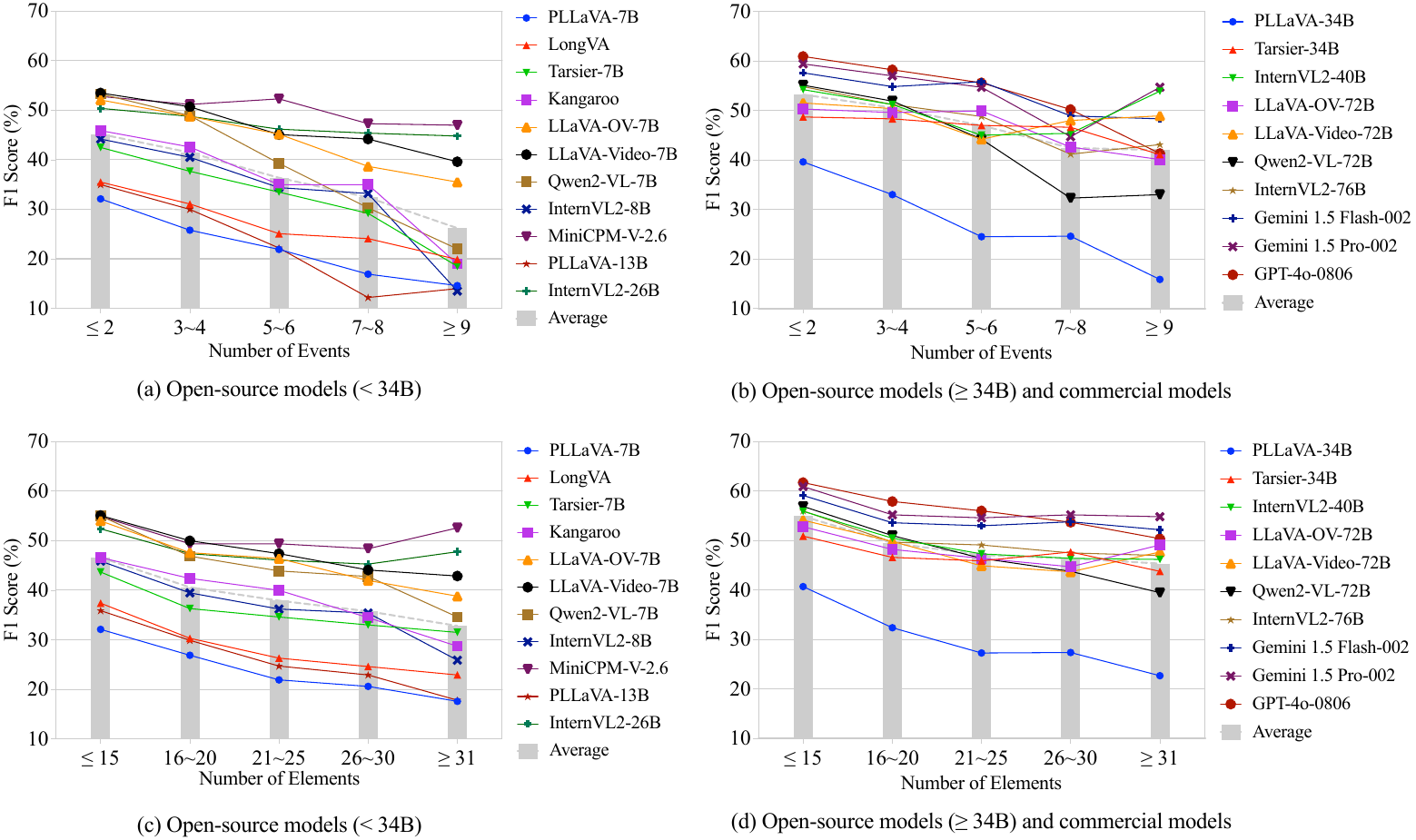}
    \setlength{\abovecaptionskip}{-0.5em}
    \caption{The whole performance comparison on \benchcap across different video complexities.}
    \label{fig:comparison_complexity_full}
    \vspace{-0.5em}
\end{figure*}

\noindent\textbf{Temporally Dense Caption Annotation.} Annotators are tasked with providing a detailed chronological description of each video. They must divide the caption into multiple events based on criteria such as camera transitions, scene transitions, or story advancements. Each event is further split into multiple atomic visual elements, categorized by type and weighted by importance on a scale of 1-3. The types include \textit{camera}, \textit{scene}, \textit{action}, and \textit{attribute}. 

\noindent\textbf{Quality Review.} For quality assurance, cross-inspections are performed between annotators. Furthermore, trained video experts (non-authors) continuously review the annotations, offering feedback and prompting annotators to refine their work to ensure the high-quality annotations. During cross-inspections and expert reviews, the checking covers all annotation results including video caption, event splitting and visual element extraction as well as the type and weight of the elements.

\subsubsection{Visualized Examples}
\label{appendix:dataset_example}

\begin{figure*}[!htbp]
    \centering
    \includegraphics[width=\linewidth]{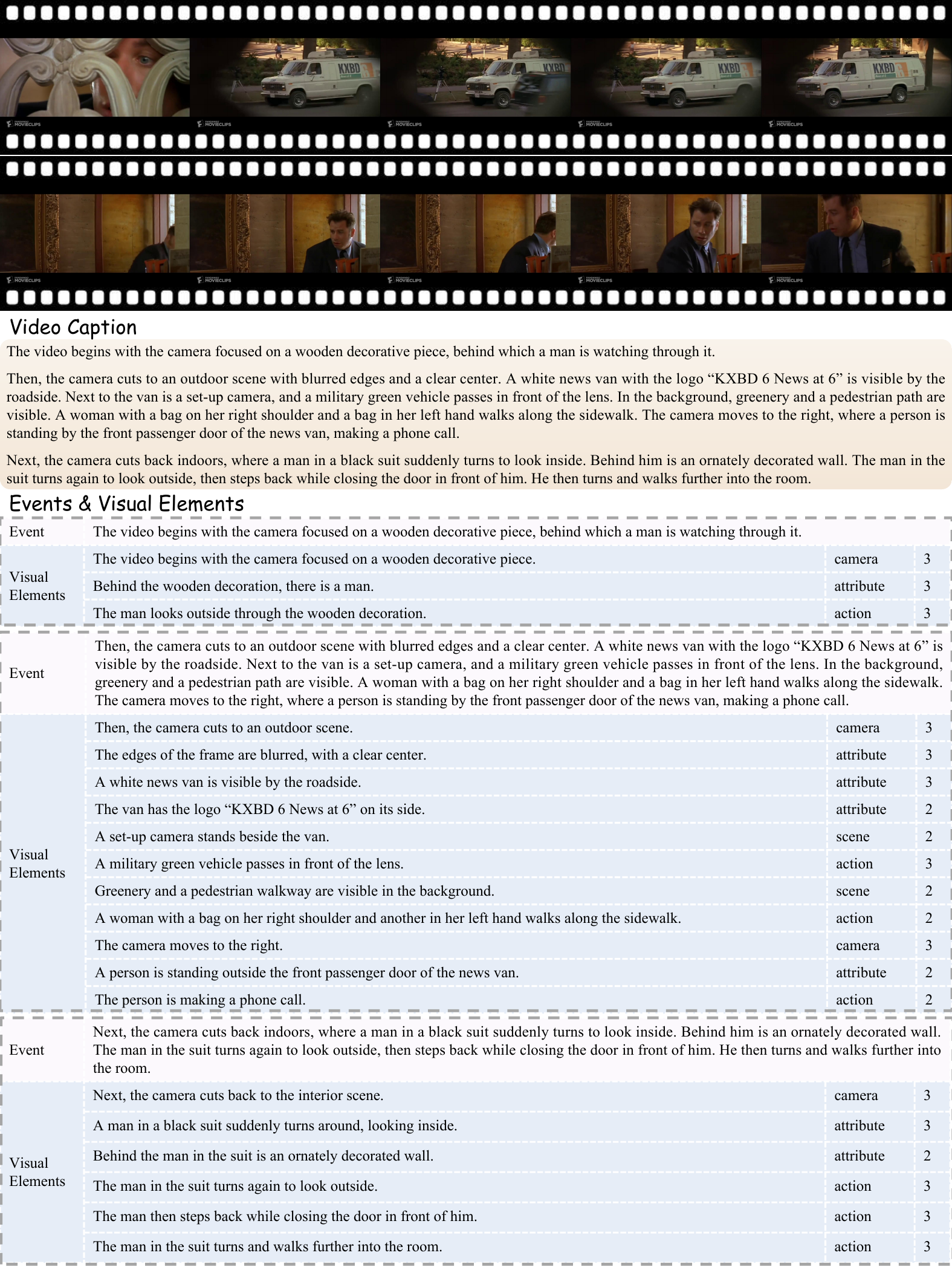}
    \caption{A detailed example in \dataset.}
    \label{fig:dataset_example_detail}
\end{figure*}

A detailed example in \dataset is shown in Figure \ref{fig:dataset_example_detail}.

\section{\benchcap}
\label{appendix:bench_captioning}

\subsection{Experimental Settings}
\label{appendix:captioning_settings}

The configuration and experimental settings for all test models are shown in Table \ref{tab:inference_settings}. 

The specific version of the closed-source models we tested are \texttt{gemini-1.5-flash-002}, \texttt{gemini-1.5-pro-002}, and \texttt{gpt-4o-2024-08-06}.
Incidentally, a few samples (less than 5) in our \benchcap and \benchmcq do not receive any responses from the Gemini \citep{reid2024gemini1.5} series, possibly due to security mechanisms. Therefore, we calculated the scores using only the samples with responses, rather than assigning a score of 0 to those without responses.

\noindent\textbf{Input Frames.} By default, we uniformly sample 32 frames from each video, which is sufficient to capture the entire content of the video in our \bench. For Qwen2-VL \citep{wang2024qwen2vl} and PLLaVA \citep{xu2024pllava}, the official strategy is followed to sample frames at 2 FPS and uniformly sample 16 frames, respectively. For closed-source models, we sampled frames dynamically with 1/2 FPS, meaning that when the video event is less than 16s, it is sampled at 2 FPS, otherwise at 1 FPS.

\begin{figure*}[!htbp]
    \centering
    \includegraphics[width=\linewidth]{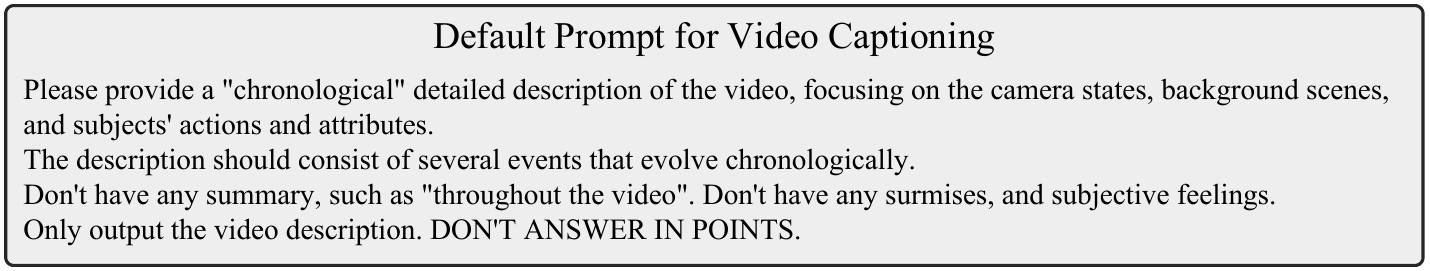}
    \caption{The default prompt used for the \benchcap experiments in Section \ref{sec:experiments_captioning}.}
    \label{fig:prompt_captioning}
\end{figure*}

\begin{figure*}[!htbp]
    \centering
    \includegraphics[width=\linewidth]{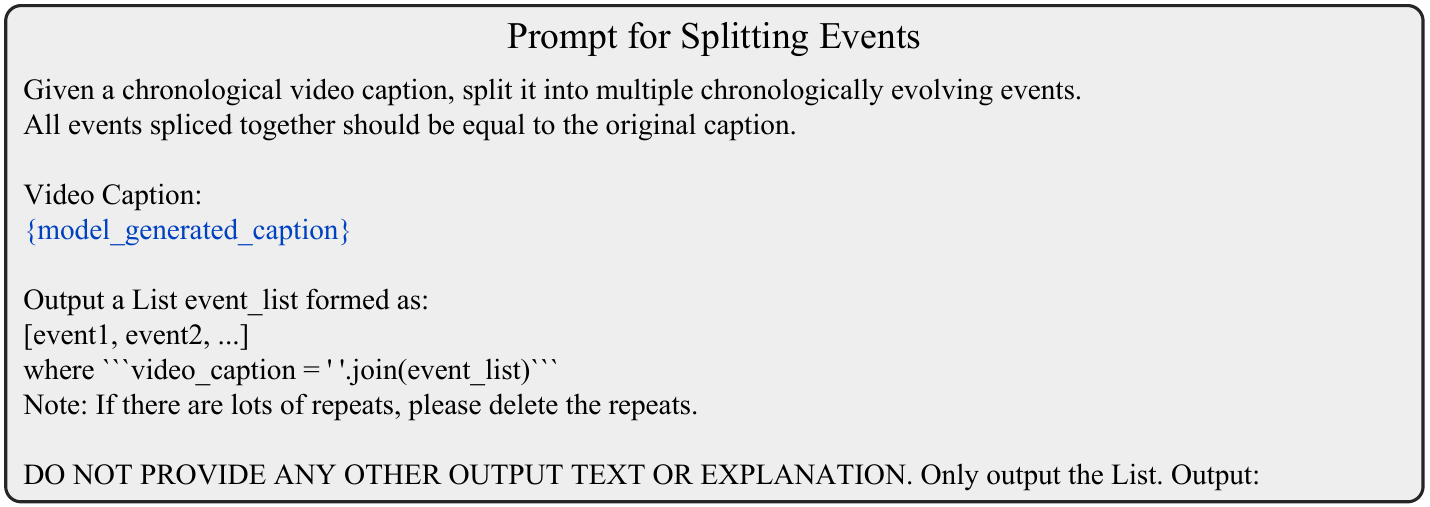}
    \caption{The prompt used to split events for the \benchcap experiments in Section \ref{sec:method_captioning}.}
    \label{fig:prompt_split_events}
\end{figure*}

\begin{figure*}[!htbp]
    \centering
    \includegraphics[width=\linewidth]{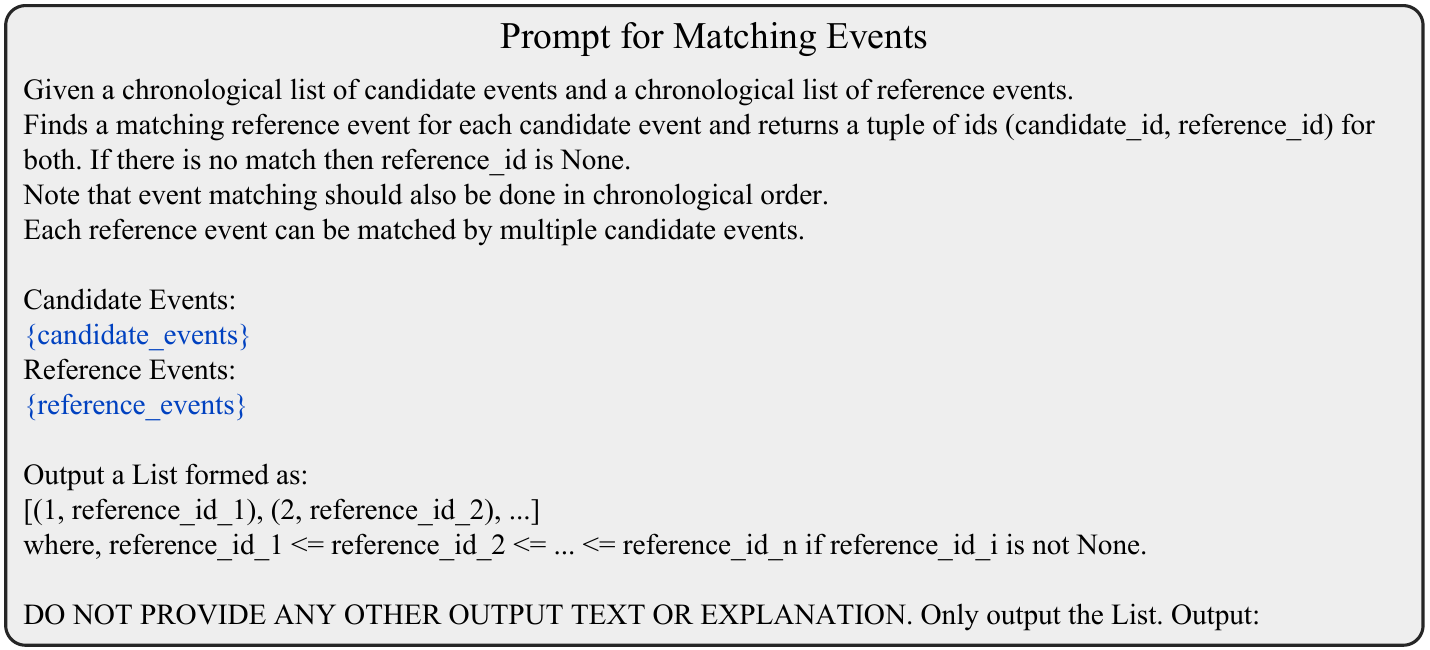}
    \caption{The prompt used to match events for the \benchcap experiments in Section \ref{sec:method_captioning}.}
    \label{fig:prompt_match_events}
\end{figure*}

\begin{figure*}[!htbp]
    \centering
    \includegraphics[width=\linewidth]{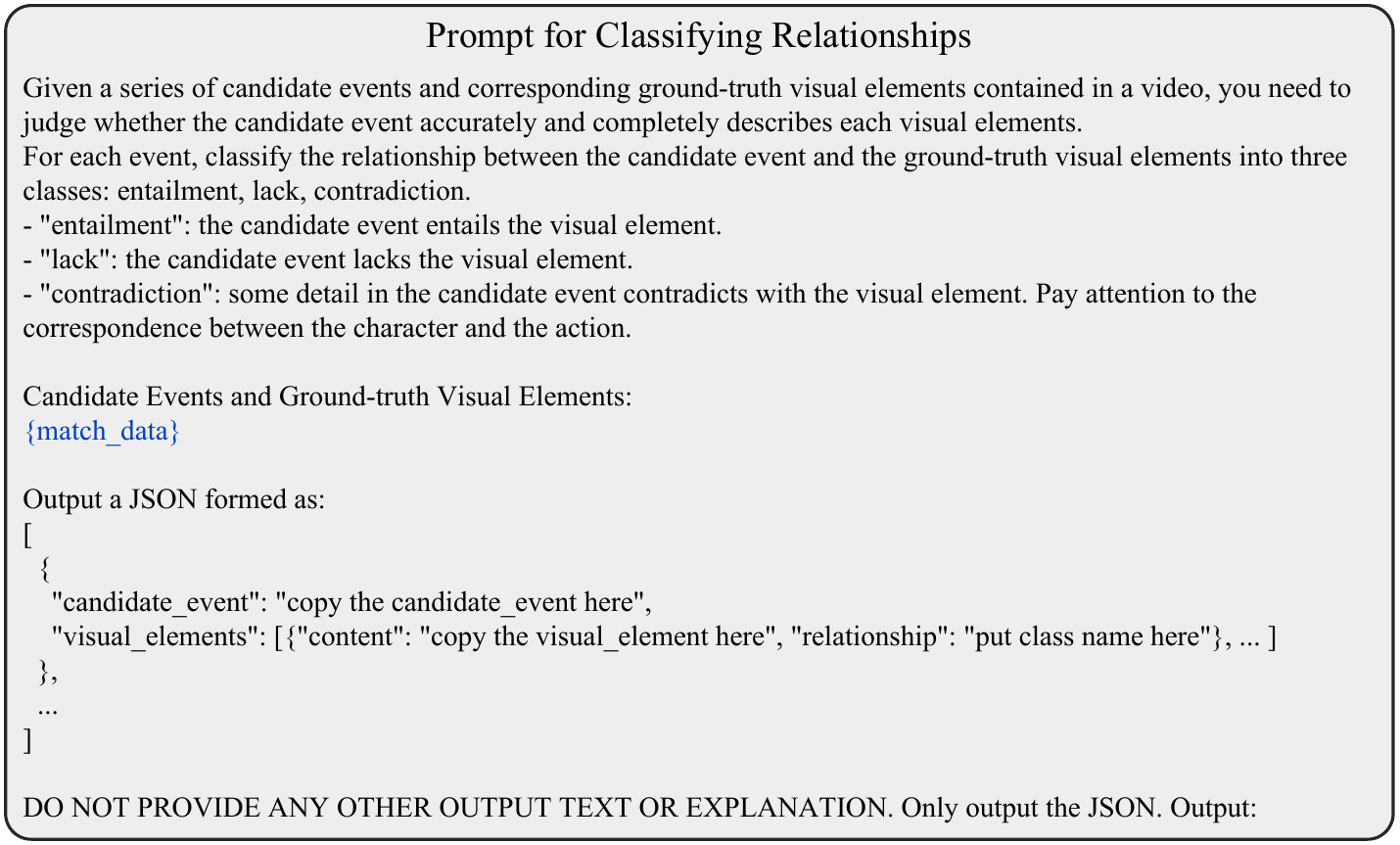}
    \caption{The prompt used to classify relationships for the \benchcap experiments in Section \ref{sec:method_captioning}.}
    \label{fig:prompt_classify_relationships}
\end{figure*}

\noindent\textbf{Detailed Prompts.} The default prompt template for captioning is shown in Figure \ref{fig:prompt_captioning}. Figures \ref{fig:prompt_split_events}, \ref{fig:prompt_match_events}, and \ref{fig:prompt_classify_relationships} illustrate the prompt templates used to evaluate \benchcap.

\subsection{More Experimental Analysis}
\label{appendix:benchcap_more_exp}

\benchcap results of all tested models are shown in Table \ref{tab:full_results_cap_dynamic} and Table \ref{tab:full_results_cap_characteristic}, as a complement result to Table \ref{tab:results_cap_full}.

\subsubsection{Video Complexity}
\label{appendix:captioning_exp_complexity}

We partition the video complexity according to the number of events and the number of visual elements in the video, to observe the impact of the model on increasing video complexity. The visualization results of selected models are shown in Figure \ref{fig:comparison_complexity}. The detailed results of all tested models are in Table \ref{tab:detailed_video_complexity}, and its visualization results are shown in Figure \ref{fig:comparison_complexity_full}.

As demonstrated in Table \ref{tab:detailed_video_complexity} and Figure \ref{fig:comparison_complexity_full}, model performance consistently declines with increasing video complexity. Larger models ($\geq$34B parameters) exhibit better robustness to complex videos, showing smaller performance drops (2.8\% for event count, 2.5\% for element count) compared to their smaller counterparts ($<$34B parameters), which experience steeper declines (4.7\% and 3.5\% respectively). Moreover, the performance gap between large and small models becomes more pronounced in highly complex videos. When event count exceeds 9 (from 7\textasciitilde8 events), small models suffer a substantial 6.2\% performance drop, while large models remain stable with only a 0.7\% variation. Similarly, for videos with more than 31 elements (increased from 26\textasciitilde30), small models show a 3.0\% fluctuation compared to just 0.7\% for large models. This evidence strongly suggests that larger models possess superior adaptability to complex video content.

\subsubsection{Enrichment of Visual Inputs}
\label{appendix:captioning_nframe}

The number of input frames is crucial for video understanding, as it directly impacts whether the model receives sufficient visual content. This is particularly important in long-video scenarios, where the model's ability to answer a question depends on whether the sampled frames contain the necessary visual information. Limited by the number of input frames in existing LMMs, our \dataset ensures that $32$ frames are sufficient to cover the content of each video, considering that \dataset has an average duration of $15s$ and a maximum duration of $38s$. To explore the effect of frame number on performance, we compare the \benchcap performance with different input frame numbers across several classical models.

\begin{figure}[!htbp]
    \centering
    \includegraphics[width=\linewidth]{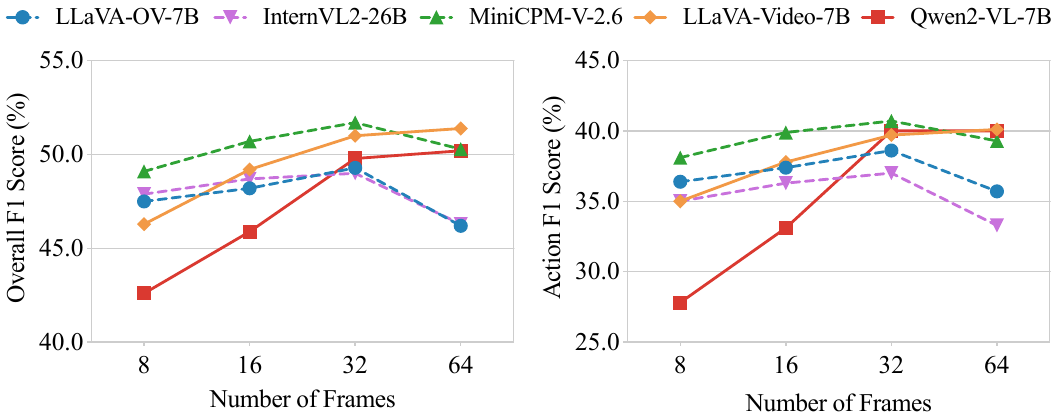}
    \caption{Performance comparison across different number of input frames.}
    \label{fig:comparison_nframes}
    \vspace{-0.5em}
\end{figure}

As shown in Figure \ref{fig:comparison_nframes}, increasing the number of frames generally improves the F1 score, with an average increase of $1.86\%$ from 8 to 16 frames and $1.62\%$ from 16 to 32 frames. This underscores the importance of providing sufficient visual information, especially when the frame count is low. Similar pattern is shown in action perception, with average improvements of $3.48\%$ from 8 to 16 frames and $2.48\%$ from 16 to 32 frames, indicating that dynamic actions are more sensitive to frame numbers. However, we observe a performance drop in some earlier models, including LLaVA-OV-7B \citep{li2024llavaov}, InternVL2-26B \citep{chen2024internvl1.5}, MiniCPM-V-2.6 \citep{yao2024minicpmv}, when the frame number is increased to 64. We attribute this decline to the fact that these earlier models rarely involve 64 frames of input (context length over 8K) during training, leading to poorer performance at 64 frames. In contrast, LLaVA-Video-7B \citep{zhang2024llavavideo} and Qwen2-VL-7B \citep{wang2024qwen2vl}, which are trained on longer contexts, achieve better results when the number of frames reaches 64. This indicates that providing more frames can indeed enhance performance when the context length is not constrained. More frames can improve the ability to capture intricate temporal dynamics and rich contextual information in videos. Consequently, exploring how to efficiently utilize more frames for training will emerge as a pivotal topic in the field of multimodal video understanding.

To further explore the effect of frame numbers on video understanding, we select LLaVA-Video and Qwen2-VL, which are trained with longer contexts, to illustrate the performance disparity across different video complexities with varying input frame numbers. Figure \ref{fig:comparison_nframes_complexity} presents the visualized results, while Table \ref{tab:detailed_nframes_complexity} provides the corresponding specific scores. These results demonstrate that increasing the number of frames is more beneficial for understanding more complex videos. However, excessive complexity can lead to performance anomalies, indicating that understanding highly complex videos remains a prominent challenge.

\subsubsection{Scaling Law}
\label{appendix:captioning_scaling_law}

As shown in Table \ref{tab:results_cap_full}, there is a general law that the performance of models increases as the model scale increases. Therefore, the scaling law is equally valid for video captioning task. Larger models typically have more parameters, enabling them to capture more complex patterns and nuances in the data, leading to improved performance. However, we notice that the LLaVA-Video series shows inconsistent performance scaling with model size. This anomaly may be attributed to the Slow-Fast approach used in LLaVA-Video-72B, which results in $2/3$ of the visual tokens being compressed to $1/4$ of the others. This compression leads to a extensive loss of fine-grained information, which is crucial for detailed video understanding and accurate captioning. This observation suggests that the efficient usage of visual information is essential and may even outweigh the impact yielded by the language model scale. The quality and richness of the visual tokens play a critical role in the overall performance of video captioning models.

\noindent\textbf{Discussion.} This observation has sparked an intriguing discussion in the field: video LMMs demonstrate superior performance when processing a higher number of input frames. While increased frame coverage provides a more comprehensive representation of video content, capturing nuanced details and temporal dynamics, this advantage is constrained by context length limitations. Specifically, accommodating more frames typically involves the compression of visual tokens, a process that remains a key technical challenge. Future research should focus on the development of more efficient visual token compression techniques and the innovation in architectural designs that can handle extended context lengths, to unlock the full potential of large-scale models in video understanding tasks.

\subsubsection{Correlation with Human Judgments}
\label{appendix:captioning_correlation}

Given a video-caption pair, this task is to check whether the metric is consistent with human scoring. Specifically, we randomly sample 40 videos containing 687 visual elements. We provided human scorers with reference meta-information and model-generated captions. The scorers were asked to sequentially determine whether each reference visual element appeared accurately and completely in the candidate captions in the correct temporal order, ultimately resulting in human-assigned scores. Finally, we calculate Kendall's $\tau$, Spearman's $\rho$, and Pearson $r$ to test the consistency of \benchcap's automatic evaluation method with human scoring. The calculated Kendall's $\tau$, Spearman's $\rho$, and Pearson $r$ are 57.2\%, 76.7\%, and 69.9\%, respectively, with all p-values $<0.05$, demonstrating the validity of our automatic evaluation method. CLAIR \citep{chan-etal-2023-clair}, an evaluation method for image captioning, is an LLM-based strategy for scoring based on reference captions. We migrate this approach seamlessly to assess video captioning as a comparative object. DREAM-1K \citep{wang2024tarsier} is a recently proposed method for video captioning evaluation with interpretability. However, it only focuses on subject actions, leading to its weak performance on our comprehensive video captioning data that focuses on camera, scene, action, and attributes.

\section{\benchmcq}
\label{appendix:bench_mcq}

\subsection{Statistics}
\label{appendix:statistics_mcq}

\begin{figure}[!htbp]
    \centering
    \includegraphics[width=\linewidth]{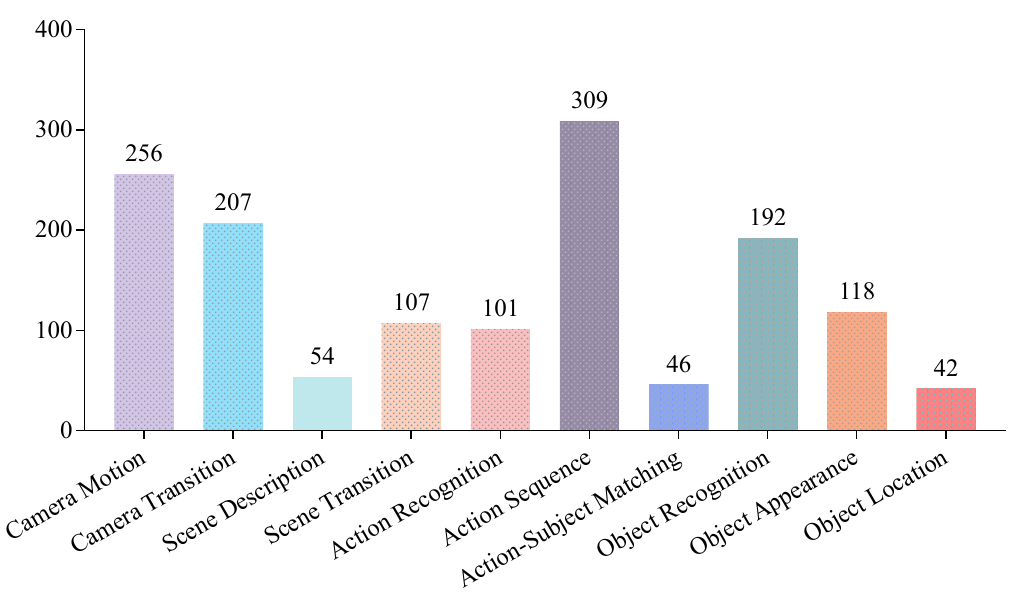}
    \caption{Sample distribution of task types in the \benchmcq, covering 10 task types.}
    \label{fig:statistics_mcq_task_type}
    \vspace{-0.5em}
\end{figure}

Figure \ref{fig:statistics_mcq_task_type} illustrates the sample distribution of the \benchmcq, across 10 tasks: (1) camera motion, e.g, zooming, panning, and rotating. (2) camera transition. (3) scene description. (4) scene transition. (5) action recognition. (6) action sequence. (7) action-subject matching. (8) object recognition. (9) object appearance, e.g., gender, age, dress, color, shape, number. and (10) object location.

\begin{figure}[!htbp]
    \centering
    \includegraphics[width=0.7\linewidth]{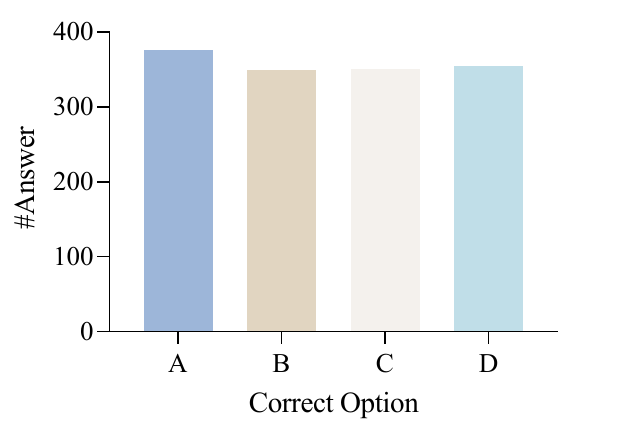}
    \setlength{\abovecaptionskip}{-0.2em}
    \caption{Sample distribution of correct option in the \benchmcq.}
    \label{fig:statistics_mcq_option}
    \vspace{-0.5em}
\end{figure}

To eliminate the bias and varied sensitivity of the models towards order and token, we ensure that the distribution of correct options is uniform, as shown in Figure \ref{fig:statistics_mcq_option}.

\subsection{More Details of \benchmcq Construction}
\label{appendix:mcq_construction}

\begin{figure*}[!htbp]
    \centering
    \includegraphics[width=\linewidth]{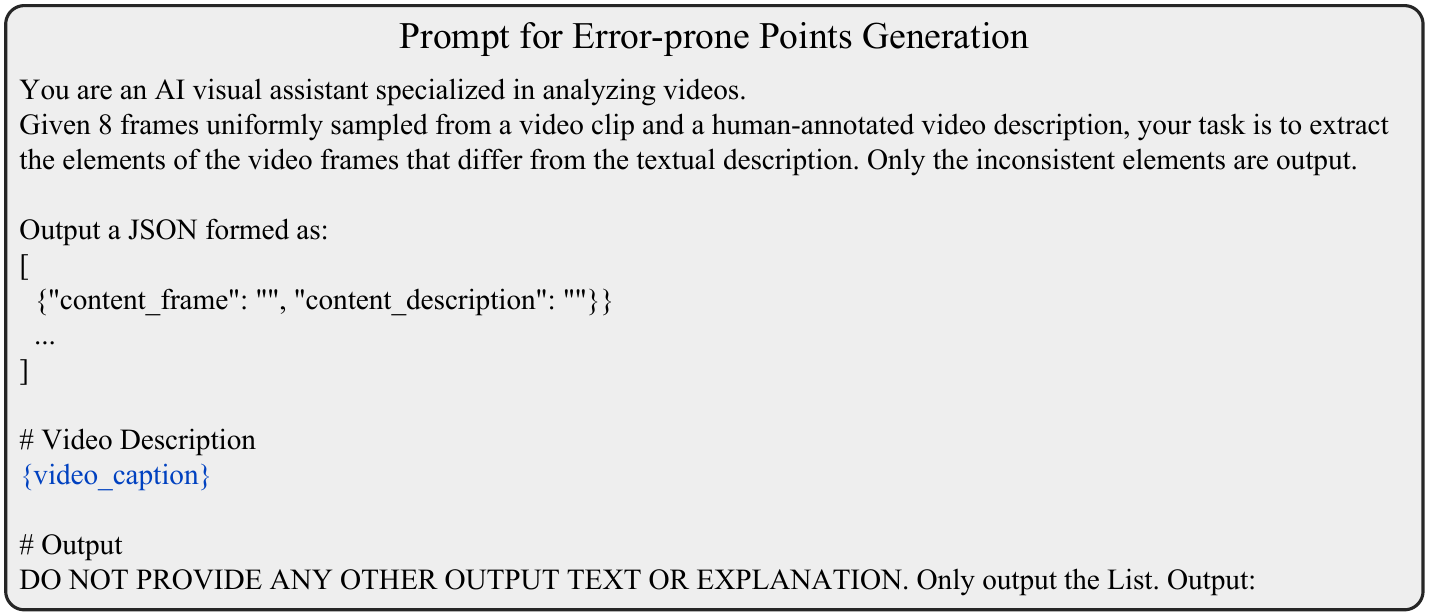}
    \caption{The prompt used to generate error-prone points for the \benchmcq construction in Section \ref{sec:method_mcq}.}
    \label{fig:prompt_error_prone_points_generation}
\end{figure*}

\noindent\textbf{Error-prone Points Extraction.} To obtain challenging questions, we obtain some error-prone points through an automated approach. Specifically, we provide the video LMM with 8 frames from the video and its ground-truth textual description, and ask it to generate what it thinks it sees that the video is inconsistent with the textual description. The prompt instruction used in this step is shown in Figure \ref{fig:prompt_error_prone_points_generation}.

\begin{figure*}[!htbp]
    \centering
    \includegraphics[width=\linewidth]{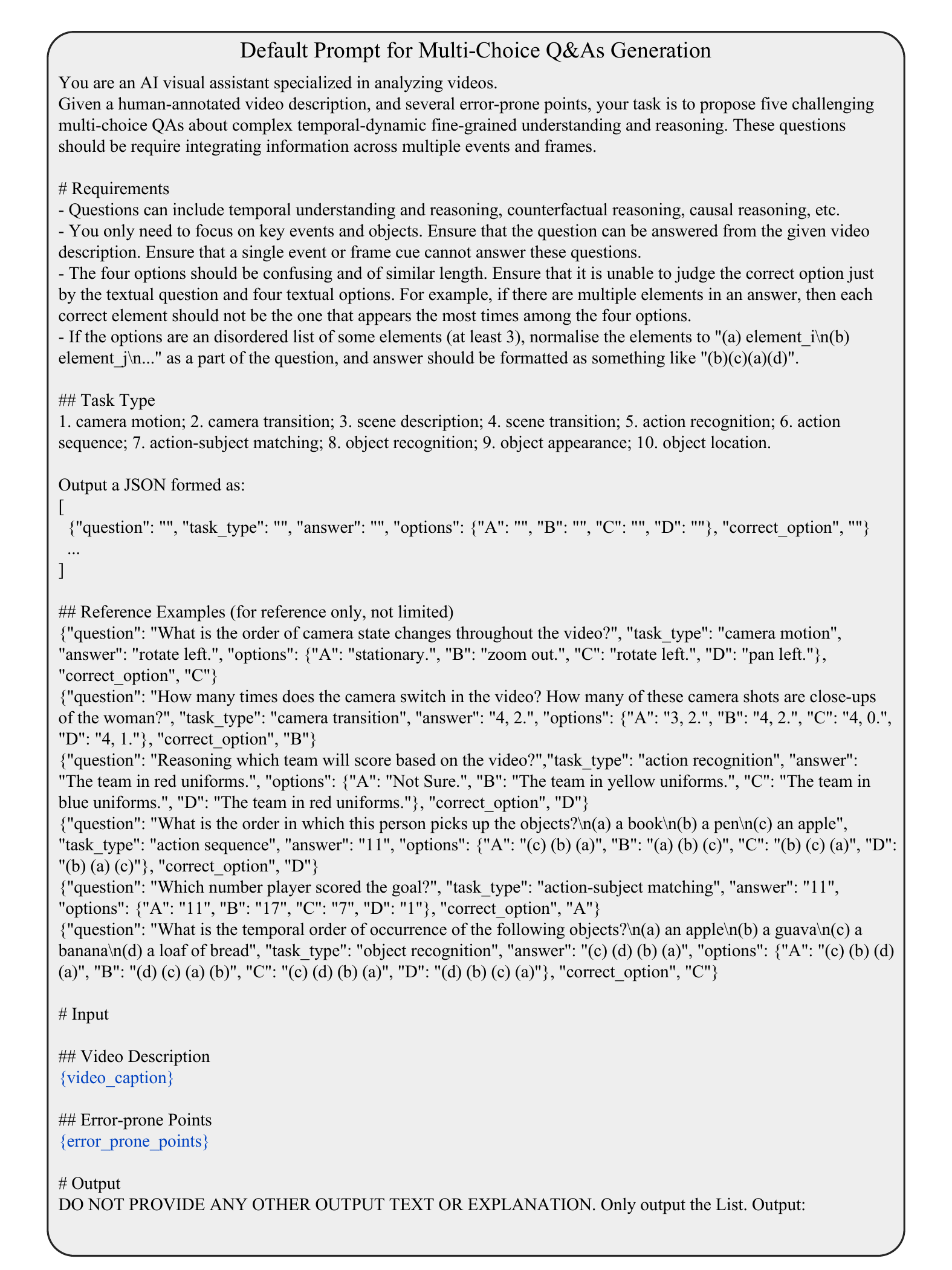}
    \caption{The prompt used to generate multi-choice QAs for the \benchmcq construction in Section \ref{sec:method_mcq}.}
    \label{fig:prompt_mcq_generation}
\end{figure*}

\noindent\textbf{Multi-Choice QA Generation.} Based on a predefined set of task types, error-prone points and textual descriptions, LLM generates several multi-choice QAs for each video. The prompt instruction used in this step is shown in Figure \ref{fig:prompt_mcq_generation}.

\noindent\textbf{Quality Review.} To ensure that data is high-quality and time-sensitive, we employ crowdsourcing to optimize the automatically generated data. In addition, human annotators perform cross-inspections to ensure quality. To guarantee that the questions are relevant to capture temporal dynamics, we employ LLaVA-Video-7B to filter them. A question is deemed temporal-indispensable if it can be accurately answered using both a single frame and multiple frames. Specifically, we deem the question to be temporal-indispensable if it can be answered correctly by both 1-frame and 16-frame inputs.

\subsubsection{Visualized Examples}
\label{appendix:mcq_example}

\begin{figure*}[!htbp]
    \centering
    \includegraphics[width=\linewidth]{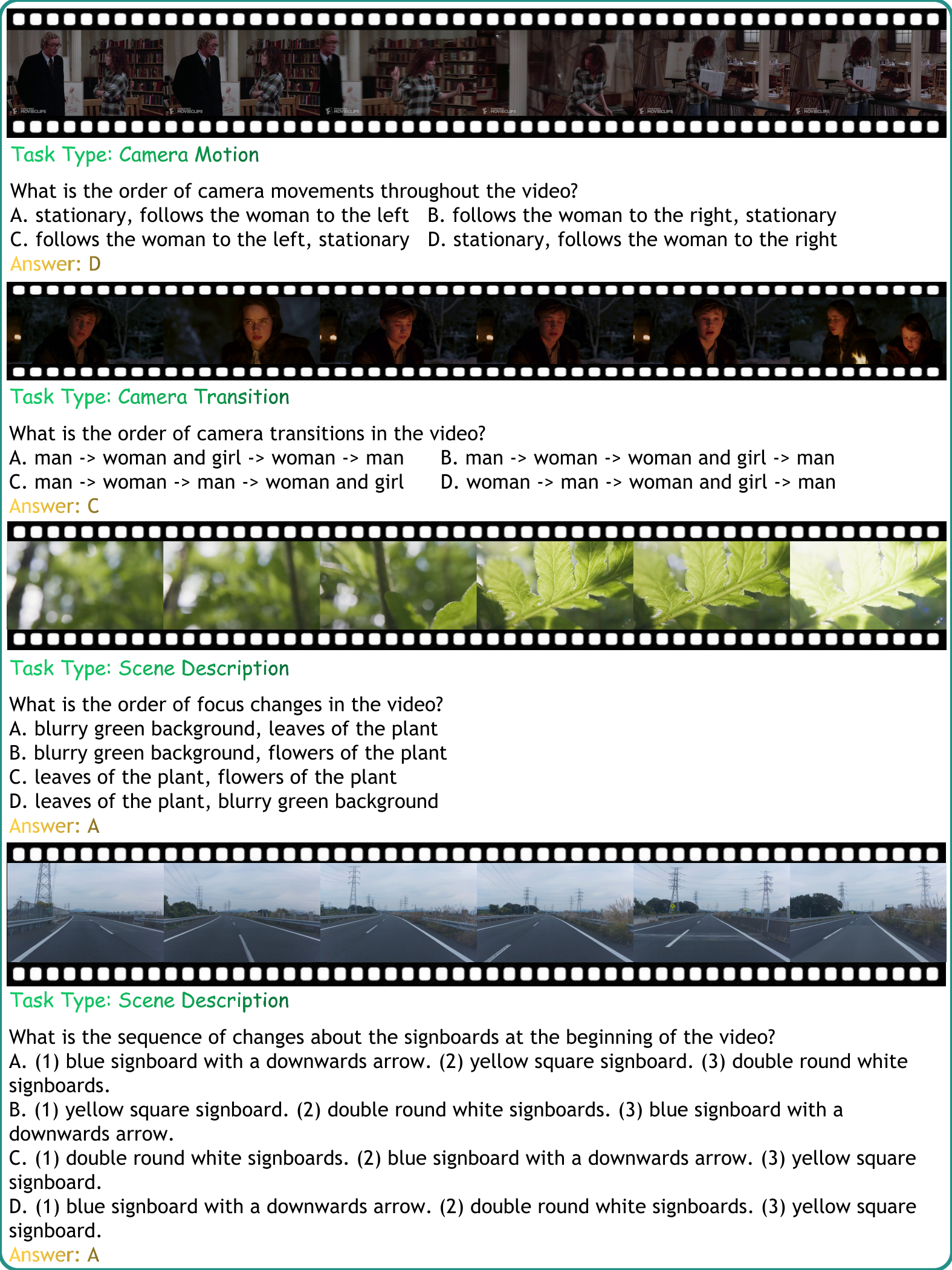}
    \caption{Several examples in \benchmcq, involving \textit{Camera Motion}, \textit{Camera Transition}, \textit{Scene Description} and \textit{Scene Transition} tasks.}
    \label{fig:mcq_examples_camera_scene}
\end{figure*}

\begin{figure*}[!htbp]
    \centering
    \includegraphics[width=\linewidth]{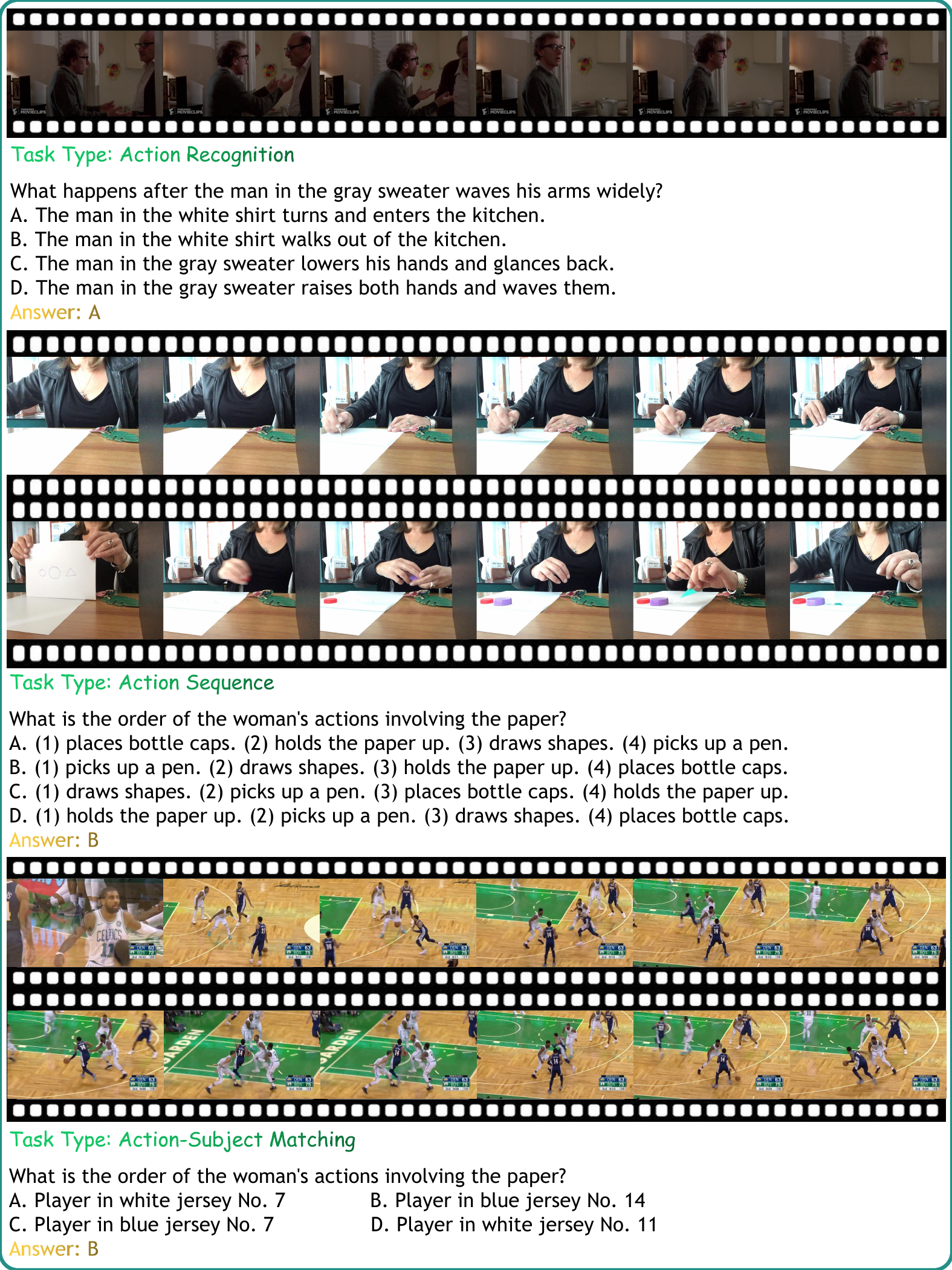}
    \caption{Several examples in \benchmcq, involving \textit{Action Recognition}, \textit{Action Sequence}, and \textit{Action-Subject Matching} tasks.}
    \label{fig:mcq_examples_action}
\end{figure*}

\begin{figure*}[!htbp]
    \centering
    \includegraphics[width=\linewidth]{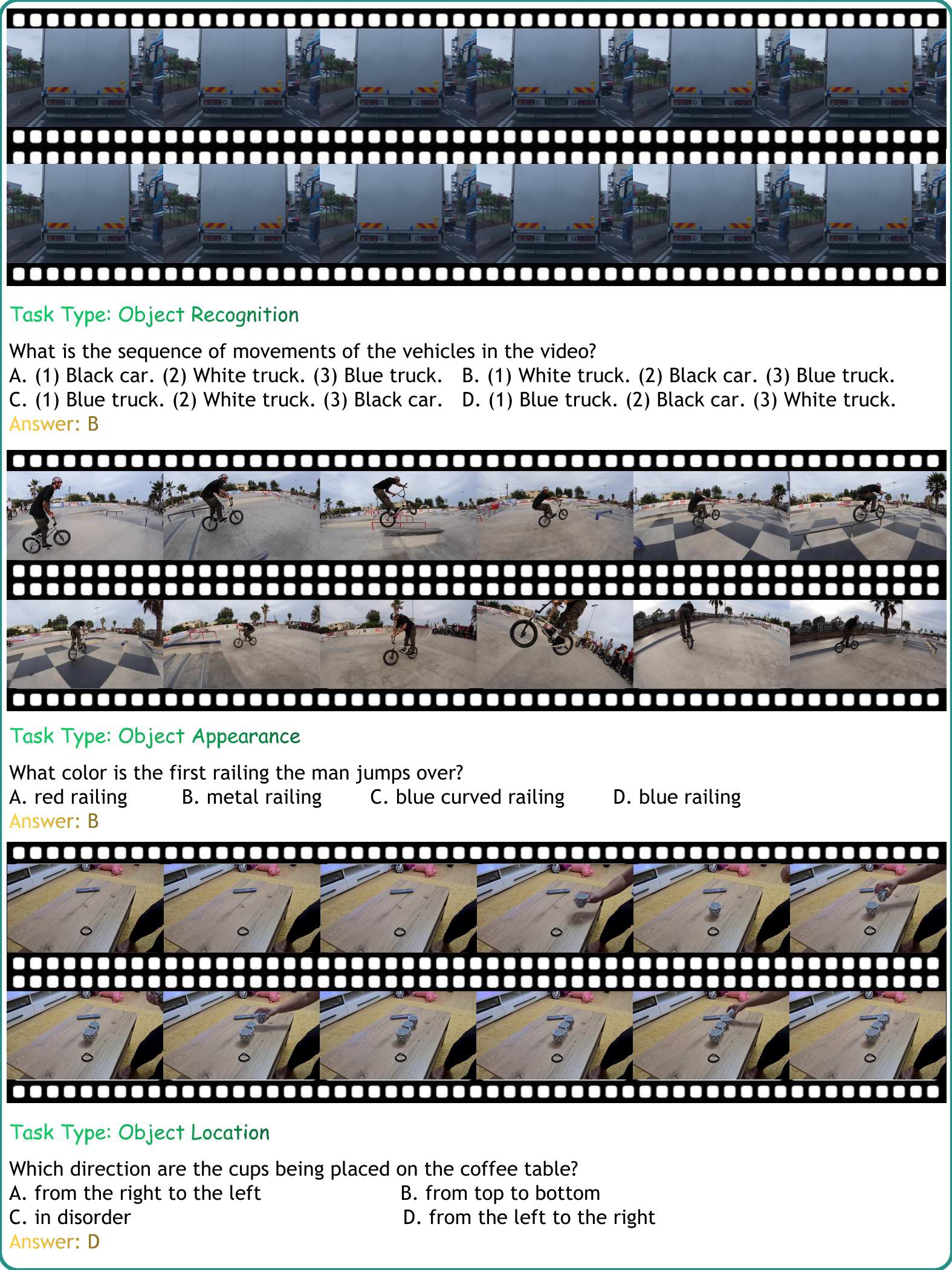}
    \caption{Several examples in \benchmcq, involving \textit{Object Recognition}, \textit{Object Appearance}, and \textit{Object Location} tasks.}
    \label{fig:mcq_examples_attribute}
\end{figure*}

Several examples in \benchmcq are shown in Figure \ref{fig:mcq_examples_camera_scene}, \ref{fig:mcq_examples_action}, and \ref{fig:mcq_examples_attribute}.

\subsection{Experimental Settings}
\label{appendix:mcq_settings}

\begin{figure*}[!htbp]
    \centering
    \includegraphics[width=\linewidth]{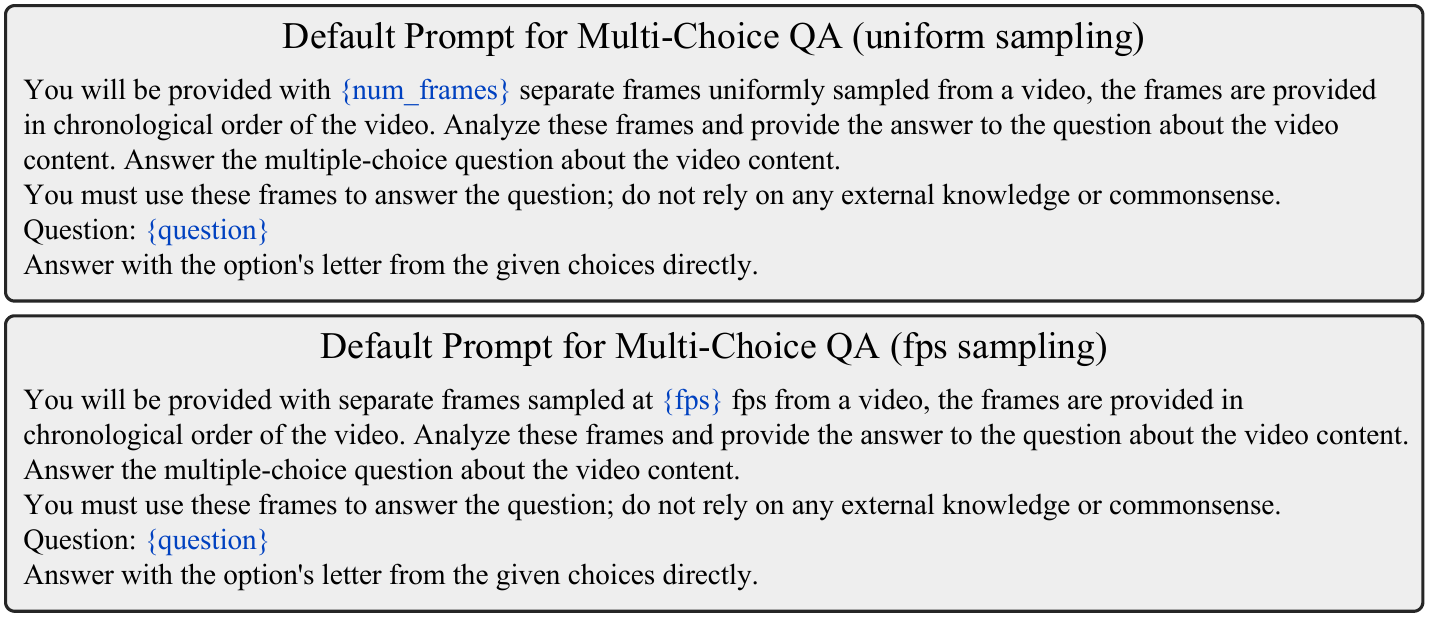}
    \caption{The default prompt used for the \benchmcq experiments in Section \ref{sec:experiments_mcq}.}
    \label{fig:prompt_mcq}
\end{figure*}

The number of input frames in \benchmcq is consistent with \benchcap, which is shown in Table \ref{tab:inference_settings}. The default prompt template for multi-choice QA is shown in Figure \ref{fig:prompt_mcq}.

Incidentally, a few samples (less than 10) in our \benchmcq do not receive any responses from the Gemini series, possibly due to security mechanisms. Therefore, we calculated the scores using only the samples with responses, rather than assigning a score of 0 to those without responses.

\subsection{More Experimental Analysis}
\label{appendix:benchmcq_more_exp}

\benchmcq results of all tested models are shown in Table \ref{tab:full_results_mcq}, as a complement result to Table \ref{tab:main_results_mcq}.

\subsubsection{Scaling Law}
\label{appendix:mcq_scaling_law}

On \benchmcq, while most models demonstrate predictable scaling patterns, InternVL2 \citep{chen2024internvl1.5} exhibits an unexpected trend where its 76B variant underperforms the 40B version and its 26B variant underperforms the 8B version. This anomaly is consistently observed across multiple video comprehension benchmarks: Video-MME (76B: 64.7\% vs. 40B: 66.1\%), MVBench (76B: 69.6\% vs. 40B: 72.0\%), MMBench-Video (76B: 1.71\% vs. 40B: 1.78\%), MLVU (76B: 69.9\% vs. 40B: 71.0\%). Notably, this counter-intuitive scaling behavior can be attributed to architectural differences: each InternVL2 variant employs distinct LLM backbone families and vision encoders, making direct performance comparisons less meaningful for establishing scaling laws.

\section{Future Work}
\label{appendix:future_work}

Considering that different models have diverse capabilities in following complex instructions, we deliberately adopted simple prompting templates to ensure fair comparison and clear assessment. While this approach helps isolate models' inherent temporal understanding abilities, advanced prompting strategies like Multimodal-CoT~\cite{zhang2023multimodal} reasoning show promising potential for performance enhancement. Although such sophisticated prompting techniques may improve performance on \benchmcq, their applicability to captioning tasks like \benchcap remains challenging. We encourage future research to explore advanced prompting strategies that can effectively enhance temporal understanding across different tasks while maintaining a balance between performance optimization and the assessment of fundamental temporal comprehension abilities.

\section{More Related Work}
\label{appendix:related_work}

\noindent\textbf{Video LMMs.} Large Mulitmodal Models (LMMs) have mushroomed, showcasing impressive visual understanding capabilities \citep{li2024multimodal, zhang2024mm, caffagni-etal-2024-revolution, amirloo2024understanding, zhang2025data}. These advances have catalyzed the development of diverse and innovative applications across multiple domains. \citep{pan2023kosmosg, zhang-etal-2024-stickerconv, liu2025llavaplus, kong2025unite}. Existing works bridge visual encoders and Large Language Models (LLMs) using a small intermediate architecture, as seen in models like LLaVA \citep{liu2024llava, liu2024llava1.5}, BLIP-2 \citep{li2023blip2}, and MiniGPT-4 \citep{zhu2023minigpt4}, which facilitate the evolution of visual-language LMMs. On this basis, recent researches \citep{li2024llavanextinterleave,zhang2024internlmxcomposer2.5,lin2024vila,cheng2024videollama2,lin2023videollava,maaz2023videochatgpt} have extended these techniques from static images to dynamic videos, demonstrating promising results in video understanding by processing videos as multiple image frames.

\begin{table*}[!htbp]
  \centering
  \resizebox{\textwidth}{!}{
    \begin{tabular}{p{0.8cm}<{\centering}|p{3cm}<{\raggedright\arraybackslash}|p{3cm}<{\centering}|p{3.5cm}<{\centering}|p{16cm}<{\raggedright\arraybackslash}}
    \toprule
    \textbf{Type} & \multicolumn{1}{c|}{\textbf{Source}} & \textbf{Domain} & \textbf{Visual Characteristic} & \multicolumn{1}{c}{\textbf{Description}} \\
    \midrule
    \multirow{14}{*}{\rotatebox{90}{Web Data}} &
    \multirow{6}{3cm}{Pexels \citep{pexels}} & Animals \& Pets   & \multirow{6}{*}{\major{Low-Dynamic}} 
    & \multirow{6}{16cm}{A website offer stock videos and motion graphics free from copyright issues, which are usually exceptionally high-quality videos uploaded by skilled photographers. We sample 46 videos, as a source of Low-Dynamic scenarios, covering diverse domains.} \\
    & & Autos \& Vehicles & &  \\
    & & Cityscape         & &  \\
    & & Foods             & &  \\
    & & Natural Landscape & &  \\
    & & Urban Activity    & &  \\
    \cmidrule(lr){2-5}
    & \multirow{5}{3cm}{Pixabay \citep{pixabay}} & Animals \& Pets & \multirow{5}{*}{\major{Low-Dynamic}} & 
    \multirow{5}{16cm}{A website offer stock videos and motion graphics free from copyright issues, which are usually exceptionally high-quality videos uploaded by skilled photographers. We sample 13 videos, as a source of Low-Dynamic scenarios, covering diverse domains.} \\
    & & Cityscape         & &  \\
    & & Foods             & &  \\
    & & Natural Landscape & &  \\
    & & Urban Activity    & &  \\
    \cmidrule(lr){2-5}
    & \multirow{3}{3cm}{MixKit \citep{mixkit}} & \multirow{3}{*}{Natural Landscape} & \multirow{3}{*}{\major{Low-Dynamic}} 
    & \multirow{3}{16cm}{A website offer stock videos and motion graphics free from copyright issues, which are usually exceptionally high-quality videos uploaded by skilled photographers. We sample 7 videos, as a source of Low-Dynamic scenarios.} \\
    & & & & \\
    & & & & \\
    \midrule
    \multirow{14}{*}{\rotatebox{90}{Academic Video Understanding Data}} 
    & \multirow{4}{3cm}{DREAM-1K \citep{wang2024tarsier}} & \multirow{4}{*}{Film} & \minor{Low-Dynamic} & \multirow{4}{16cm}{DREAM-1K consists of 1,000 video clips from five categories: live-action movies, animated movies, stock videos, YouTube videos, and TikTok-style short videos. These videos typically feature multiple events and subjects across various shots. We sample 148 videos from live-action movies that meet our selection principles, mostly as a source of High-Dynamic, Multi-Scene, and Multi-Subject scenarios.} \\
    & & & \major{High-Dynamic}  &  \\
    & & & \major{Multi-Scene}   &  \\
    & & & \major{Multi-Subject} &  \\
    \cmidrule(lr){2-5}
    & \multirow{4}{3cm}{VELOCITI \citep{saravanan2024velociti}} & \multirow{4}{*}{Film} & \minor{Low-Dynamic} 
    & \multirow{4}{16cm}{A benchmark using complex movie clips and dense semantic role label annotations to test perception and binding in video LMMs. The videos feature challenging scenarios with frequent shot changes, fast action sequences, multi-event situations, role switching, and entity co-referencing over time. We sample 266 videos, mostly as a source of High-Dynamic, Multi-Scene, and Multi-Subject scenarios.} \\
    & & & \major{High-Dynamic}  &  \\
    & & & \major{Multi-Scene}   &  \\
    & & & \major{Multi-Subject} &  \\
    \cmidrule(lr){2-5}
    & \multirow{3}{3cm}{PerceptionTest \citep{patraucean2024perceptiontest}} & \multirow{3}{*}{\makecell{Daily Life \\ (Indoor)}} & \minor{Low-Dynamic} 
    & \multirow{3}{16cm}{A dataset evaluates performance across skill areas (memory, abstraction, physics, semantics) and reasoning types (descriptive, explanatory, predictive, counterfactual). We sample 114 videos, mostly as a source of High-Dynamic scenarios.} \\
    & & & \major{High-Dynamic} &  \\
    & & & \minor{Multi-Scene}  &  \\
    \cmidrule(lr){2-5}
    & \multirow{3}{3cm}{YouCook2 \citep{zhou2018youcook2}} & \multirow{3}{*}{Cooking} & \major{High-Dynamic} 
    & \multirow{3}{16cm}{A dataset of YouTube videos covering 89 recipes from four major cuisines (Africa, Americas, Asia, Europe), featuring diverse cooking styles and challenges like fast camera motion, camera zooms, video defocus, and scene-type changes. We sample 100 videos, mostly as a source of High-Dynamic scenarios.} \\
    & & & \minor{Multi-Scene}   &  \\
    & & & \minor{Multi-Subject} &  \\
    \midrule
    \multirow{15}{*}{\rotatebox{90}{Academic Video Generation Data}} & 
    \multirow{8}{3cm}{VIDGEN-1M \citep{tan2024vidgen1m}} & Animals \& Pets & \multirow{8}{*}{\makecell{\major{Low-Dynamic} \\ \major{High-Dynamic} \\ \major{Multi-Scene}}}
    & \multirow{8}{16cm}{Open-domain Text-to-Video dataset with high video quality, high temporal consistency, and balanced categories. We sample 154 videos, as a source of High-Dynamic, Multi-Scene (Sports Activity) scenarios, and Low-Dynamic (other domains) scenarios.} \\
    & & Autos \& Vehicles & &  \\
    & & Cityscape         & &  \\
    & & Foods             & &  \\
    & & Natural Landscape & &  \\
    & & Plants            & &  \\
    & & Urban Activity    & &  \\
    & & Sports Activity   & &  \\
    \cmidrule(lr){2-5}
    & \multirow{7}{3cm}{MiraData \citep{ju2024miradata}} & Animals \& Pets & \multirow{7}{*}{\makecell{\major{Low-Dynamic} \\ \minor{Multi-Scene}}} 
    & \multirow{7}{16cm}{A large-scale, high-quality video dataset designed to meet the key expectations of video generation tasks: diverse content, high visual quality, long duration, and significant motion strength. Unlike existing text-to-video datasets that primarily source videos from YouTube, MiraData includes videos from YouTube, Videvo, Pixabay, and Pexels, ensuring a more comprehensive and suitable data source. We sample 102 videos, mostly as a source of Low-Dynamic scenarios, covering diverse domains.} \\
    & & Autos \& Vehicles & &  \\
    & & Cityscape         & &  \\
    & & Foods             & &  \\
    & & Natural Landscape & &  \\
    & & Plants            & &  \\
    & & Urban Activity    & &  \\
    \midrule
    \multirow{5}{*}{\rotatebox{90}{\makecell{Others}}}
    & \multirow{5}{3cm}{CoVLA \citep{arai2024covla}} & \multirow{5}{*}{Driving} & \multirow{5}{*}{\makecell{\minor{Low-Dynamic} \\ \major{Multi-Scene} \\ \minor{Multi-Subject}}} 
    & \multirow{5}{16cm}{The CoVLA (Comprehensive Vision-Language-Action) dataset is a novel large-scale resource designed to advance autonomous driving research. The dataset includes synchronized multi-modal data streams from front-facing cameras, in-vehicle signals, and other sensors, providing a comprehensive view of diverse driving scenarios. We choose it due to its complex scene variations. We sample 50 videos as a source of Multi-Scene scenarios.} \\
    & & & &  \\
    & & & &  \\
    & & & & \\
    & & & & \\
    \bottomrule
    \end{tabular}%
}
  \caption{Rich video sources within \dataset. \textbf{Domain}  denote the domains represented in the sampled data. \textbf{Visual Characteristic} indicates the visual characteristics present in the sampled data, with \major{bold} representing major features and \minor{grey} representing minor features.. We also provide a brief description of each dataset, along with our our selection criteria and counts.}
  \label{tab:detailed_video_source}%
\end{table*}

\begin{table*}[!htbp]
  \centering
  \resizebox{0.65\textwidth}{!}{
    \begin{tabular}{lllc}
    \toprule
    Model & LLM & Vision Model & \#Frames \\
    \midrule
    \rowcolor{gray!10}\multicolumn{4}{l}{\textit{\textbf{Open-Source LMMs}}} \\ 
    Qwen2-VL-72B & Qwen2-72B & ViT-600M & 2FPS \\
    Qwen2-VL-7B & Qwen2-7B & ViT-600M & 2FPS \\
    LLaVA-Video-72B & Qwen2-72B & SigLIP-400M & 32 \\
    LLaVA-Video-7B & Qwen2-7B & SigLIP-400M & 32 \\
    LLaVA-OneVision-72B & Qwen2-72B & SigLIP-400M & 32 \\
    LLaVA-OneVision-7B & Qwen2-7B & SigLIP-400M & 32 \\
    InternVL2-76B & Llama-3-70B-Instruct & InternViT-6B & 32 \\
    InternVL2-40B & Nous-Hermes-2-Yi-34B & InternViT-6B & 32 \\
    InternVL2-26B & InternLM2-20B & InternViT-6B & 32 \\
    InternVL2-8B & InternLM2.5-7B & InternViT-300M & 32 \\
    Tarsier-34B & Nous-Hermes-2-Yi-34B & CLIP ViT-L/14 & 32 \\
    Tarsier-7B & Vicuna-v1.5-7B & CLIP ViT-L/14 & 32 \\
    PLLaVA-34B & Nous-Hermes-2-Yi-34B & CLIP ViT-L/14 & 16 \\
    PLLaVA-13B & Vicuna-v1.5-13B & CLIP ViT-L/14 & 16 \\
    PLLaVA-7B & Vicuna-v1.5-7B & CLIP ViT-L/14 & 16 \\
    MiniCPM-V-2.6 & Qwen2-7B & SigLIP-400M & 32 \\
    Kangaroo & Llama3-8B-Instruct & EVA-CLIP-L & 32 \\
    LongVA-7B & Qwen2-7B-Instruct-224K & CLIP ViT-L/14 & 32 \\
    \midrule
    \rowcolor{gray!10}\multicolumn{4}{l}{\textit{\textbf{Closed-Source LMMs}}}\\
    GPT-4o & Unknown & Unknown & 1/2 FPS$^\ast$ \\
    Gemini 1.5 Pro & Unknown & Unknown & 1/2 FPS$^\ast$ \\
    Gemini 1.5 Flash & Unknown & Unknown & 1/2 FPS$^\ast$ \\
    \bottomrule
    \end{tabular}%
}
  \caption{The number of frames used in the \bench evaluation in Section \ref{sec:experiments_captioning}, \ref{sec:experiments_mcq}. By default, 32 frames are sampled uniformly, which is enough to cover the content of each video in \benchcap. Some models take a different number of frames because they are limited by the input length or according to their sampling recommendations. $^\ast$ indicates that 2 FPS is employed when the video duration $<16s$, otherwise 1 FPS is employed. The versions of the closed-source models are \texttt{gpt-4o-2024-08-06}, \texttt{gemini-1.5-pro-002}, \texttt{gemini-1.5-flash-002}.}
  \label{tab:inference_settings}%
\end{table*}

\begin{table*}[!htbp]
  \centering
  \vspace{1mm}
  \resizebox{\textwidth}{!}{
    \begin{tabular}{l|ccc|ccc|ccc|ccc|ccc}
    \toprule
    \multirow{2}[2]{*}{Model} & \multicolumn{3}{c|}{Camera} & \multicolumn{3}{c|}{Scene} & \multicolumn{3}{c|}{Action} & \multicolumn{3}{c|}{Attribute} & \multicolumn{3}{c}{Overall} \\
    \cmidrule(lr){2-4}\cmidrule(lr){5-7}\cmidrule(lr){8-10}\cmidrule(lr){11-13}\cmidrule(lr){14-16}
    & P & R & F1 & P & R & F1 & P & R & F1 & P & R & F1 & P & R & F1  \\
    \midrule
    \rowcolor{gray!10}\multicolumn{16}{l}{\textit{\textbf{Open-Source LMMs}}}\\
    PLLaVA-7B & 49.4 & 22.6 & 28.9 & 52.2 & 30.9 & 36.6 & 30.5 & 12.6 & 16.5 & 44.5 & 19.5 & 25.3 & 60.0 & 19.1 & 27.4 \\
    LongVA-7B & 52.3 & 26.0 & 32.5 & 56.5 & 34.4 & 40.6 & 38.9 & 17.2 & 22.0 & 50.6 & 22.0 & 28.4 & 71.6 & 22.3 & 31.8 \\
    Tarsier-7B & 56.9 & 27.3 & 34.8 & 45.3 & 28.2 & 33.1 & 56.7 & 28.9 & 36.2 & 56.4 & 26.0 & 33.3 & 73.0 & 27.9 & 38.6 \\
    Kangaroo & 65.2 & 36.5 & 44.1 & 67.8 & 45.4 & 51.9 & 49.3 & 26.0 & 31.9 & 59.8 & 32.2 & 39.5 & 69.5 & 32.5 & 42.7 \\
    LLaVA-OV-7B & 75.2 & 42.0 & 51.0 & 71.8 & 51.2 & 57.6 & 54.1 & 30.4 & 36.8 & 66.2 & 42.0 & 49.3 & 73.6 & 38.6 & 49.3 \\
    LLaVA-Video-7B & 74.0 & 41.5 & 50.4 & \secondcell{73.6} & 52.3 & 58.9 & 57.0 & 30.8 & 37.8 & \bestcell{72.1} & \secondcell{44.8} & \bestcell{53.1} & 77.0 & 39.7 & 51.0 \\
    Qwen2-VL-7B & 72.3 & 40.7 & 49.0 & 71.9 & 50.0 & 56.7 & 55.9 & 30.1 & 37.0 & 68.2 & 38.4 & 46.7 & \bestcell{77.8} & 37.6 & 48.9 \\
    InternVL2-8B & 64.8 & 33.7 & 41.7 & 59.4 & 38.7 & 44.7 & 45.2 & 24.7 & 30.0 & 59.8 & 35.5 & 42.3 & 67.2 & 31.1 & 40.8 \\
    MiniCPM-V-2.6 & \secondcell{76.5} & \bestcell{47.8} & \bestcell{56.0} & \bestcell{75.0} & \secondcell{54.1} & \secondcell{60.6} & 57.2 & 31.8 & 38.8 & \secondcell{68.7} & 42.3 & 50.2 & 76.0 & 40.7 & \secondcell{51.7} \\
    \midrule
    PLLaVA-13B & 57.0 & 25.8 & 33.0 & 57.3 & 34.0 & 40.3 & 36.2 & 13.8 & 18.5 & 50.0 & 23.3 & 29.8 & 65.0 & 21.4 & 30.6 \\
    InternVL2-26B & 73.2 & 43.2 & 51.6 & 72.5 & 52.6 & 58.7 & 51.7 & 30.9 & 37.0 & 63.9 & 42.3 & 49.1 & 70.0 & 39.2 & 49.0 \\
    PLLaVA-34B & 60.8 & 29.6 & 37.4 & 56.2 & 33.7 & 39.9 & 38.7 & 17.3 & 22.3 & 55.1 & 26.1 & 33.2 & 67.8 & 24.5 & 34.2 \\
    Tarsier-34B & 63.6 & 34.3 & 42.3 & 59.0 & 38.4 & 44.4 & \bestcell{65.6} & \bestcell{39.9} & \bestcell{47.6} & 63.6 & 34.3 & 42.2 & \secondcell{77.1} & 36.7 & 48.2 \\
    InternVL2-40B & \bestcell{77.8} & \secondcell{46.3} & \secondcell{55.1} & 71.9 & 53.1 & 59.0 & 53.4 & 33.1 & 39.3 & 65.9 & \bestcell{45.7} & \secondcell{52.3} & 71.3 & \secondcell{42.1} & \secondcell{51.7} \\
    \midrule
    LLaVA-OV-72B & 73.5 & 43.7 & 51.9 & 71.5 & 51.1 & 57.5 & 51.2 & 30.2 & 36.0 & 65.7 & 41.4 & 48.8 & 72.7 & 39.2 & 49.6 \\
    LLaVA-Video-72B & 72.7 & 41.7 & 50.3 & 71.1 & 49.9 & 56.4 & 55.7 & 32.7 & 39.3 & 68.1 & 43.2 & 50.8 & 73.7 & 39.6 & 50.2 \\
    Qwen2-VL-72B & 73.6 & 45.9 & 54.0 & 67.6 & 46.3 & 52.8 & \secondcell{59.1} & \secondcell{35.7} & \secondcell{42.6} & 66.6 & 40.7 & 48.5 & 74.7 & 41.1 & \secondcell{51.7} \\
    InternVL2-76B & 75.1 & 45.4 & 53.9 & 73.3 & \bestcell{55.8} & \bestcell{61.4} & 55.7 & 34.9 & 41.2 & 64.3 & 44.5 & 50.9 & 70.7 & \bestcell{42.3} & \bestcell{51.9} \\
    \midrule
    \rowcolor{gray!10}\multicolumn{16}{l}{\textit{\textbf{Closed-Source LMMs}}}\\
    Gemini 1.5 Flash & 74.6 & 52.8 & 59.6 & \secondcell{77.2} & \secondcell{59.3} & \secondcell{65.1} & 58.7 & 36.4 & 42.9 & \secondcell{69.0} & 48.4 & 55.2 & 72.7 & 46.4 & 55.7 \\
    Gemini 1.5 Pro & \secondcell{78.7} & \secondcell{53.0} & \secondcell{60.7} & 75.7 & 57.4 & 63.3 & \secondcell{59.0} & \secondcell{40.3} & \secondcell{46.3} & \secondcell{69.0} & \secondcell{49.4} & \secondcell{56.0} & \secondcell{73.7} & \secondcell{48.1} & \secondcell{57.4} \\
    GPT-4o & \bestcell{80.1} & \bestcell{53.3} & \bestcell{61.3} & \bestcell{79.5} & \bestcell{60.2} & \bestcell{66.4} & \bestcell{64.0} & \bestcell{41.1} & \bestcell{48.0} & \bestcell{73.8} & \bestcell{50.1} & \bestcell{57.8} & \bestcell{77.7} & \bestcell{48.2} & \bestcell{58.5} \\
    \bottomrule
    \end{tabular}%
}
  \caption{Evaluation results in terms of dynamic element categoryies on \benchcap. The best and second-best results are marked with \colorbox{best}{\makebox(26,6){\vspace{-1mm}orange}} and \colorbox{second}{\makebox(16,6){blue}}, respectively.}
  \label{tab:full_results_cap_dynamic}%
\end{table*}

\begin{table*}[!htbp]
  \centering
  \vspace{1mm}
  \resizebox{\textwidth}{!}{
    \begin{tabular}{l|ccc|ccc|ccc|ccc|ccc}
    \toprule
    \multirow{2}[2]{*}{Model} & \multicolumn{3}{c|}{Low-Dynamic} & \multicolumn{3}{c|}{High-Dynamic} & \multicolumn{3}{c|}{Multi-Scene} & \multicolumn{3}{c|}{Multi-Subject} & \multicolumn{3}{c}{Overall} \\
    \cmidrule(lr){2-4}\cmidrule(lr){5-7}\cmidrule(lr){8-10}\cmidrule(lr){11-13}\cmidrule(lr){14-16}
    & P & R & F1 & P & R & F1 & P & R & F1 & P & R & F1 & P & R & F1  \\
    \midrule
    \rowcolor{gray!10}\multicolumn{16}{l}{\textit{\textbf{Open-Source LMMs}}}\\
    PLLaVA-7B & 66.5 & 23.0 & 32.7 & 56.6 & 17.1 & 24.7 & 55.7 & 15.5 & 22.8 & 56.2 & 15.3 & 22.5 & 60.0 & 19.1 & 27.4 \\
    LongVA-7B & 75.9 & 26.5 & 37.3 & 69.4 & 20.1 & 29.0 & 68.3 & 19.0 & 27.6 & 67.3 & 15.7 & 23.7 & 71.6 & 22.3 & 31.8 \\
    Tarsier-7B & \bestcell{81.2} & 34.3 & 46.5 & 68.7 & 24.5 & 34.5 & 71.7 & 25.3 & 35.8 & 67.8 & 23.2 & 33.2 & 73.0 & 27.9 & 38.6 \\
    Kangaroo & 73.2 & 34.7 & 45.6 & 67.6 & 31.3 & 41.1 & 66.2 & 29.7 & 39.3 & 63.5 & 26.3 & 35.7 & 69.5 & 32.5 & 42.7 \\
    LLaVA-OV-7B & 78.6 & 38.4 & 50.0 & 71.0 & 38.8 & 48.9 & 71.7 & 38.3 & 48.4 & 67.1 & 33.8 & 43.8 & 73.6 & 38.6 & 49.3 \\
    LLaVA-Video-7B & \secondcell{80.7} & 40.0 & 52.2 & 75.1 & 39.5 & 50.3 & \secondcell{77.1} & 38.6 & 50.0 & 73.5 & 34.6 & 45.8 & 77.0 & 39.7 & 51.0 \\
    Qwen2-VL-7B & \bestcell{81.2} & 42.0 & 53.8 & \bestcell{76.0} & 35.3 & 46.4 & 76.8 & 33.2 & 44.4 & \secondcell{73.6} & 28.9 & 39.9 & \bestcell{77.8} & 37.6 & 48.9 \\
    InternVL2-8B & 71.6 & 34.0 & 44.5 & 64.9 & 29.7 & 38.9 & 65.6 & 29.1 & 38.4 & 61.5 & 26.6 & 35.2 & 67.2 & 31.1 & 40.8 \\
    MiniCPM-V-2.6 & 79.3 & 41.4 & 53.0 & 74.3 & 40.4 & \secondcell{51.0} & 76.5 & \secondcell{40.8} & \bestcell{51.7} & 73.5 & 38.3 & 49.0 & 76.0 & 40.7 & \secondcell{51.7} \\
    \midrule
    PLLaVA-13B & 69.8 & 25.7 & 36.0 & 62.5 & 19.1 & 27.8 & 62.3 & 17.6 & 26.0 & 60.3 & 16.3 & 24.3 & 65.0 & 21.4 & 30.6 \\
    InternVL2-26B & 71.9 & 39.1 & 49.4 & 69.0 & 39.2 & 48.9 & 70.3 & 38.6 & 48.4 & 67.2 & 36.3 & 45.8 & 70.0 & 39.2 & 49.0 \\
    PLLaVA-34B & 74.5 & 28.1 & 38.9 & 64.3 & 22.6 & 31.8 & 63.9 & 21.3 & 30.2 & 60.7 & 19.2 & 27.6 & 67.8 & 24.5 & 34.2 \\
    Tarsier-34B & 79.6 & 37.2 & 49.1 & \secondcell{75.8} & 36.5 & 47.8 & \bestcell{77.6} & 38.1 & 49.6 & \bestcell{74.4} & 36.0 & 47.3 & \secondcell{77.1} & 36.7 & 48.2 \\
    InternVL2-40B & 75.0 & \secondcell{43.8} & \secondcell{53.9} & 69.5 & \secondcell{41.2} & 50.5 & 70.7 & 40.8 & 50.5 & 67.9 & 38.7 & 48.0 & 71.3 & \secondcell{42.1} & \secondcell{51.7} \\
    \midrule
    LLaVA-OV-72B & 75.4 & 37.3 & 48.6 & 71.3 & 36.7 & 45.9 & 71.4 & 40.1 & 50.1 & 72.3 & \secondcell{39.1} & \bestcell{49.4} & 72.7 & 39.2 & 49.6 \\
    LLaVA-Video-72B & 77.3 & 39.2 & 50.6 & 71.9 & 39.8 & 50.0 & 73.9 & 38.6 & 49.3 & 70.5 & 35.1 & 45.7 & 73.7 & 39.6 & 50.2 \\
    Qwen2-VL-72B & 79.2 & \bestcell{44.6} & \bestcell{55.7} & 72.4 & 39.3 & 49.7 & 73.6 & 37.2 & 48.0 & 69.1 & 32.8 & 43.3 & 74.7 & 41.1 & \secondcell{51.7} \\
    InternVL2-76B & 72.0 & 43.1 & 52.8 & 70.1 & \bestcell{41.9} & \bestcell{51.5} & 71.4 & \bestcell{41.1} & \secondcell{51.1} & 68.6 & \bestcell{39.7} & \secondcell{49.3} & 70.7 & \bestcell{42.3} & \bestcell{51.9} \\
    \midrule
    \rowcolor{gray!10}\multicolumn{16}{l}{\textit{\textbf{Closed-Source LMMs}}}\\
    Gemini 1.5 Flash & 74.0 & 46.5 & 56.0 & 72.0 & 46.4 & 55.5 & \secondcell{73.4} & 46.2 & 55.9 & \secondcell{73.4} & \bestcell{46.2} & \bestcell{55.9} & 72.7 & 46.4 & 55.7 \\
    Gemini 1.5 Pro & \secondcell{76.7} & \bestcell{48.7} & \bestcell{58.7} & \secondcell{72.1} & \secondcell{47.8} & \secondcell{56.7} & \secondcell{73.4} & \bestcell{47.7} & \secondcell{57.0} & 69.9 & 44.1 & 53.3 & \secondcell{73.7} & \secondcell{48.1} & \secondcell{57.4} \\
    GPT-4o & \bestcell{79.1} & \secondcell{47.3} & \secondcell{58.2} & \bestcell{77.0} & \bestcell{48.6} & \bestcell{58.7} & \bestcell{78.7} & \secondcell{47.2} & \bestcell{58.1} & \bestcell{76.8} & \secondcell{44.4} & \secondcell{55.5} & \bestcell{77.7} & \bestcell{48.2} & \bestcell{58.5} \\
    \bottomrule
    \end{tabular}%
}
  \caption{Evaluation results in terms of visual characteristic categoryies on \benchcap. The best and second-best results are marked with \colorbox{best}{\makebox(26,6){\vspace{-1mm}orange}} and \colorbox{second}{\makebox(16,6){blue}}, respectively.}
  \label{tab:full_results_cap_characteristic}%
\end{table*}

\begin{table*}[!htbp]
  \centering
  \vspace{1mm}
  \resizebox{\textwidth}{!}{
    \begin{tabular}{l|lllll|lllll|c}
    \toprule
    \multirow{2}[2]{*}{Model} & \multicolumn{5}{c|}{\#Events} & \multicolumn{5}{c|}{\#Elements} & \multirow{2}[2]{*}{Overall} \\
    \cmidrule(lr){2-6}\cmidrule(lr){7-11}
    & $\leq$ 2  & 3\textasciitilde 4  & 5\textasciitilde6  & 7\textasciitilde8  & $\geq$ 9  & $\leq$ 15 & 16\textasciitilde20 & 21\textasciitilde25 & 26\textasciitilde30 & $\geq$ 31 & \\
    \midrule
    \rowcolor{gray!10}\multicolumn{12}{l}{\textit{\textbf{Open-Source LMMs}}}\\
    PLLaVA-7B        & 32.1 & 25.8 & 21.9 & 16.9 & 14.6 & 32.1 & 26.9  & 21.9  & 20.6  & 17.6 & 27.4 \\
    LongVA-7B        & 35.5 & 31.1 & 25.1 & 24.1 & 19.9 & 37.4 & 30.3  & 26.3  & 24.6  & 22.9 & 31.8 \\
    Tarsier-7B       & 42.5 & 37.7 & 33.5 & 29.2 & 18.5 & 43.7 & 36.3  & 34.6  & 33.0  & 31.5 & 38.6 \\
    Kangaroo         & 45.9 & 42.6 & 35.0 & 35.0 & 19.0 & 46.6 & 42.4  & 40.0  & 34.5  & 28.7 & 42.7 \\
    LLaVA-OV-7B      & 52.1 & 48.8 & 45.2 & 38.7 & 35.5 & 54.0 & 47.6  & 46.4  & 42.0  & 38.8 & 49.3 \\
    LLaVA-Video-7B   & 53.5 & 50.7 & 45.1 & 44.2 & 39.6 & 55.1 & 50.0  & 47.4  & 44.1  & 42.9 & 51.0 \\
    Qwen2-VL-7B      & 53.3 & 48.9 & 39.3 & 30.3 & 22.0 & 55.0 & 46.9  & 43.9  & 42.8  & 34.6 & 48.9 \\
    InternVL2-8B     & 44.2 & 40.5 & 34.4 & 33.2 & 13.5 & 45.9 & 39.5  & 36.2  & 35.4  & 25.9 & 40.8 \\
    MiniCPM-V-2.6    & 52.8 & 51.2 & 52.3 & 47.3 & 47.0 & 54.9 & 49.4  & 49.4  & 48.4  & 52.6 & 51.7 \\
    PLLaVA-13B       & 35.0 & 30.0 & 22.2 & 12.2 & 14.0 & 35.9 & 29.9  & 24.7  & 22.9  & 17.8 & 30.6 \\
    InternVL2-26B    & 50.4 & 48.8 & 46.2 & 45.4 & 44.8 & 52.4 & 47.4  & 46.1  & 45.3  & 47.8 & 49.0 \\
    \midrule
    Avg ($<$34B) & 45.2 & 41.5 \down{3.7} & 36.4 \down{5.1} & 32.4 \down{4.0} & 26.2 \down{6.2} & 46.6 & 40.6 \down{6.0} & 37.9 \down{2.7} & 35.8 \down{2.1} & 32.8 \down{3.0} & 42.0 \\
    \midrule
    PLLaVA-34B       & 39.6 & 33.0 & 24.5 & 24.6 & 15.9 & 40.7 & 32.4  & 27.3  & 27.4  & 22.7 & 34.2 \\
    Tarsier-34B      & 48.7 & 48.3 & 47.0 & 46.6 & 41.1 & 50.9 & 46.6  & 45.9  & 47.7  & 43.8 & 48.2 \\
    InternVL2-40B    & 54.2 & 51.2 & 45.0 & 45.4 & 53.9 & 55.9 & 50.5  & 47.3  & 46.4  & 46.2 & 51.7 \\
    LLaVA-OV-72B     & 50.3 & 49.6 & 49.9 & 42.6 & 40.1 & 52.8 & 48.2  & 46.4  & 44.7  & 49.1 & 49.6 \\
    LLaVA-Video-72B  & 51.5 & 50.4 & 44.2 & 48.0 & 48.9 & 54.1 & 49.8  & 44.9  & 43.7  & 47.7 & 50.2 \\
    Qwen2-VL-72B     & 55.1 & 51.9 & 44.2 & 32.3 & 33.0 & 56.9 & 51.0  & 46.4  & 43.8  & 39.5 & 51.7 \\
    InternVL2-76B    & 54.8 & 51.2 & 48.8 & 41.2 & 43.1 & 56.0 & 49.7  & 49.1  & 47.5  & 47.0 & 51.9 \\
    \midrule
    Avg ($\geq$34B) & 50.6 & 47.9 \down{2.7} & 43.4 \down{4.6} & 40.1 \down{3.3} & 39.4 \down{0.7} & 52.5 & 46.9 \down{5.6} & 43.9 \down{3.0} & 43.0 \down{0.9} & 42.3 \down{0.7} & 48.2 \\
    \midrule
    \midrule
    \rowcolor{gray!10}\multicolumn{12}{l}{\textit{\textbf{Closed-Source LMMs}}}\\
    Gemini 1.5 Flash & 57.6 & 54.8 & 55.8 & 48.9 & 48.3 & 59.1 & 53.6 & 53.0 & 53.8 & 52.2 & 55.7 \\
    Gemini 1.5 Pro & 59.4 & 57.0 & 54.7 & 44.7 & 54.7 & 60.9 & 55.2 & 54.6 & 55.2 & 54.8 & 57.4\\
    GPT-4o & 60.9 & 58.2 & 55.6 & 50.2 & 41.3 & 61.7 & 57.9 & 56.0 & 53.7 & 50.4 & 58.5 \\
    \midrule
    Avg (close-source) & 59.3 & 56.7 \down{2.6} & 55.4 \down{1.3} & 47.9 \down{7.4} & 48.1 \up{0.2} & 60.6 & 55.6 \down{5.0} & 54.5 \down{1.0} & 54.2 \down{0.3} & 52.5 \down{1.8} & 57.2 \\
    \midrule
    \midrule
    Avg (Total) & 49.0\enspace\enspace\enspace & 45.8 \down{3.2} & 41.4 \down{4.4} & 37.2 \down{4.2} & 33.7 \down{3.4} & 50.6\enspace\enspace\enspace & 44.8 \down{5.7} & 42.3 \down{2.6} & 40.8 \down{1.4} & 38.8 \down{2.0} & 46.2 \\
    \bottomrule
    \end{tabular}%
}
  \caption{Detailed performance comparison with varying video complexities. Video complexity is measured by the number of events and the number of visual elements in the video. The inference setup is consistent with Table \ref{tab:inference_settings}.}
  \label{tab:detailed_video_complexity}%
\end{table*}

\begin{table*}[!htbp]
  \centering
  \vspace{1mm}
  \resizebox{\textwidth}{!}{
    \begin{tabular}{l|c|llll|llll|l}
    \toprule
    Model & Frames & Camera & Scene & Action & Attribute & Low-Dynamic & High-Dynamic & Multi-Scene & Multi-Subject & Overall \\
    \midrule
    \multirow{4}{*}{LLaVA-OV-7B} 
    & 8  & 50.2 & 56.6 & 33.0 & 47.8 & 50.4 & 46.1 & 46.2 & 42.6 & 47.5  \\
    & 16 & 49.3 \down{0.9} & 57.2 \up{0.6} & 35.6 \up{2.6} & 48.9 \up{1.1} & 50.2 \down{0.2} & 47.1 \up{1.0} & 47.0 \up{0.8} & 42.5 \up{0.1} & 48.2 \up{0.7} \\
    & 32 & 51.0 \up{1.7} & 57.6 \up{0.4} & 36.8 \up{1.2} & 49.3 \up{0.4} & 50.0 \down{0.2} & 48.9 \up{1.8} & 48.4 \up{1.4} & 43.8 \up{1.3} & 49.3 \up{1.1} \\
    & 64 & 47.4 \down{3.6} & 54.6 \down{3} & 33.5 \down{3.3} & 45.9 \down{3.4} & 48.8 \down{1.2} & 44.8 \down{4.1} & 44.6 \down{3.8} & 39.9 \down{3.9} & 46.2 \down{3.1}  \\
    \midrule
    \multirow{4}{*}{MiniCPM-V-2.6} 
    & 8  & 56.3 & 59.8 & 33.0 & 47.3 & 52.9 & 47.1 & 48.3 & 44.8 & 49.1  \\
    & 16 & 55.5 \down{0.8} & 60.5 \up{0.7} & 36.7 \up{3.7} & 47.9 \up{0.6} & 52.6 \down{0.3} & 49.7 \up{2.6} & 50.8 \up{2.5} & 48.1 \up{3.3} & 50.7 \up{1.6}  \\
    & 32 & 56.0 \up{0.5} & 60.6 \up{0.1} & 38.8 \up{2.1} & 50.2 \up{2.3} & 53.0 \up{0.4} & 51.0 \up{1.3} & 51.7 \up{0.9} & 49.0 \up{0.9} & 51.7 \up{1.0} \\
    & 64 & 52.6 \down{3.4} & 58.2 \down{2.4} & 39.1 \up{0.3} & 48.6 \down{1.6} & 50.5 \down{2.5} & 50.3 \down{0.7} & 50.0 \down{1.7} & 46.9 \down{2.1} & 50.3 \down{1.4} \\
    \midrule
    \multirow{4}{*}{InternVL2-26B} 
    & 8  & 50.1 & 58.3 & 35.0 & 48.8 & 49.6 & 47.1 & 47.0 & 43.9 & 47.9  \\
    & 16 & 50.0 \down{0.1} & 59.1 \up{0.8} & 36.3 \up{1.3} & 49.9 \up{1.1} & 49.4 \down{0.2} & 48.4 \up{1.3} & 48.3 \up{1.3} & 45.4 \up{1.5} & 48.7 \up{0.8} \\
    & 32 & 51.6 \up{1.6} & 58.7 \down{0.4} & 37.0 \up{0.7} & 49.1 \down{0.8} & 49.4 \tie & 48.9 \up{0.5} & 48.4 \up{0.1} & 45.8 \up{0.4} & 49.0 \up{0.3} \\
    & 64 & 49.6 \down{2} & 55.1 \down{3.6} & 33.3 \down{3.7} & 46.4 \down{2.7} & 47.5 \down{1.9} & 45.7 \down{3.2} & 44.3 \down{4.1} & 42.2 \down{3.6} & 46.3 \down{2.7}  \\
    \midrule
    \multirow{4}{*}{LLaVA-Video-7B}
    & 8  & 49.3 & 55.1 & 31.8 & 46.8 & 49.8 & 44.6 & 44.3 & 41.0 & 46.3  \\
    & 16 & 50.7 \up{1.4} & 57.0 \up{1.9} & 36.3 \up{4.5} & 49.0 \up{2.2} & 51.7 \up{1.9} & 47.9 \up{3.3} & 47.0 \up{2.7} & 43.0 \up{2.0} & 49.2 \up{2.9} \\
    & 32 & 50.4 \up{0.3} & 58.9 \up{1.9} & 37.8 \up{1.5} & 53.1 \up{4.1} & 52.2 \up{0.5} & 50.3 \up{2.4} & 50.0 \up{3.0} & 45.8 \up{2.8} & 51.0 \up{1.8} \\
    & 64 & 51.0 \up{0.6} & 58.7 \down{0.2} & 39.0 \up{1.2} & 52.4 \down{0.7} & 51.3 \down{0.9} & 51.4 \up{1.1} & 50.1 \up{0.1} & 46.9 \up{1.1} & 51.4 \up{0.4} \\
    \midrule
    \multirow{4}{*}{Qwen2-VL-7B} 
    & 8  & 44.2 & 55.5 & 27.8 & 41.7 & 49.6 & 39.0 & 37.1 & 33.2 & 42.6 \\
    & 16 & 47.7 \up{3.5} & 55.9 \up{0.4} & 33.1 \up{5.3} & 43.6 \up{1.9} & 51.5 \up{1.9} & 43.0 \up{4.0} & 42.0 \up{4.9} & 36.7 \up{3.5} & 45.9 \up{3.3} \\
    & 32 & 48.8 \up{1.1} & 57.0 \up{1.1} & 40.0 \up{6.9} & 47.1 \up{3.5} & 52.6 \up{1.1} & 48.4 \up{5.4} & 46.5 \up{4.5} & 43.0 \up{6.3} & 49.8 \up{3.9} \\
    & 64 & 50.1 \up{1.3} & 53.1 \down{3.9} & 40.0 \tie & 49.4 \up{2.3} & 53.2 \up{0.6} & 48.7 \up{0.3} & 47.1 \up{0.6} & 43.4 \up{0.4} & 50.2 \up{0.4}\\
    \bottomrule
    \end{tabular}%
}
  \caption{Detailed performance comparison with different number of input frames. Consistent visual results in Figure \ref{fig:comparison_nframes}.}
  \label{tab:detailed_input_frames}%
\end{table*}

\begin{table*}[!htbp]
  \centering
  \vspace{1mm}
  \resizebox{\textwidth}{!}{
    \begin{tabular}{l|c|lllll|lllll|l}
    \toprule
    \multirow{2}[2]{*}{Model} & \multirow{2}[2]{*}{Frames} & \multicolumn{5}{c|}{\#Events} & \multicolumn{5}{c|}{\#Elements} & \multirow{2}[2]{*}{Overall} \\
    \cmidrule(lr){3-7}\cmidrule(lr){8-12}
    & & $\leq$ 2  & 3\textasciitilde 4  & 5\textasciitilde6  & 7\textasciitilde8  & $\geq$ 9  & $\leq$ 15 & 16\textasciitilde20 & 21\textasciitilde25 & 26\textasciitilde30 & $\geq$ 31 & \\
    \midrule
    \multirow{4}{*}{LLaVA-Video-7B} 
    & 8  & 49.4 & 46.3 & 39.6 & 35.2 & 24.9 & 51.7 & 45.3  & 40.9  & 38.6  & 35.0 & 46.3 \\
    & 16 & 52.9 \up{3.5} & 48.8 \up{2.5} & 41.2 \up{1.6} & 36.7 \up{1.5} & 35.7 \up{10.8} & 54.6 \up{2.9} & 47.7  \up{2.4} & 44.0 \up{3.1} & 42.4 \up{3.8} & 38.2 \up{3.2} & 49.2 \up{2.9} \\
    & 32 & 53.5 \up{0.6} & 50.7 \up{1.9} & 45.1 \up{3.9} & 44.2 \up{7.5} & 39.6 \up{3.9} & 55.1 \up{0.5} & 50.0 \up{2.3} & 47.4 \up{3.4} & 44.1 \up{1.7} & 42.9 \up{4.7} & 51.0 \up{1.8} \\
    & 64 & 53.7 \up{0.2} & 51.1 \up{0.4} & 46.6 \up{1.5} & 44.2 \tie & 39.4 \down{0.2} & 55.2 \up{0.1} & 50.9 \up{0.9} & 47.9 \up{0.5} & 44.7 \up{0.6} & 41.4 \down{1.5} & 51.4 \up{0.4} \\
    \midrule
    \multirow{4}{*}{Qwen2-VL-7B} 
    & 8  & 46.8 & 43.2 & 29.7 & 25.5 & 16.0 & 48.6 & 42.1  & 36.9  & 30.6  & 29.1 & 42.6 \\
    & 16 & 50.4 \up{3.6} & 46.1 \up{2.9} & 32.9 \up{3.2} & 32.7 \up{7.2} & 14.6 \down{1.4} & 52.3 \up{3.7} & 44.9 \up{2.8} & 39.1 \up{2.2} & 36.3 \up{5.7} & 32.6 \up{3.5} & 45.9 \up{3.3} \\
    & 32 & 53.5 \up{3.1} & 49.8 \up{3.7} & 40.0 \up{7.1} & 37.7 \up{5} & 30.6 \up{16} & 55.0 \up{2.7} & 48.5 \up{3.6} & 45.3 \up{6.2} & 42.3 \up{6} & 38.3 \up{5.7} & 49.8 \up{3.9} \\
    & 64 & 52.7 \down{0.8} & 50.5 \up{0.7} & 42.7 \up{2.7} & 45.1 \up{7.4} & 27.6 \down{3} & 55.1 \up{0.1} & 49.8 \up{1.3} & 45.2 \down{0.1} & 42.9 \up{0.6} & 36.8 \down{1.5} & 50.2 \up{0.4} \\
    \bottomrule
    \end{tabular}%
}
  \caption{Performance comparison across different video complexities with varying input frame numbers. Consistent visualization results in Figure \ref{fig:comparison_nframes_complexity}.}
  \label{tab:detailed_nframes_complexity}%
\end{table*}

\begin{table*}[!htbp]
  \centering
  \resizebox{\textwidth}{!}{
    \begin{tabular}{l|cc|cc|ccc|ccc|c}
    \toprule
    \multirow{2}[2]{*}{Model} & \multicolumn{2}{c|}{Camera State} & \multicolumn{2}{c|}{Background Scene} & \multicolumn{3}{c|}{Subject Action} & \multicolumn{3}{c|}{Object Attribute} & \multirow{2}[2]{*}{Overall} \\
    \cmidrule(lr){2-3} \cmidrule(lr){4-5} \cmidrule(lr){6-8} \cmidrule(lr){9-11}
    & Motion & Transition & Description & Transition & Recognition & Sequence & Matching & Recognition & Appearance & Location &  \\
    \midrule
    \rowcolor{gray!10}\multicolumn{12}{c}{\textit{\textbf{Open-Source LMMs}}}\\
    PLLaVA-7B        & 29.7 & 31.9 & 48.1 & 22.4 & 43.6 & 34.6 & 30.4 & 32.3 & 38.1 & 45.2 & 33.7 \\
    LongVA-7B        & 37.5 & 41.5 & 63.0 & 30.8 & 44.6 & 44.7 & 43.5 & 41.7 & 47.6 & 40.5 & 42.4 \\
    Tarsier-7B       & 23.0 & 24.6 & 40.7 & 20.6 & 38.6 & 26.9 & 45.7 & 20.9 & 25.9 & 23.8 & 26.5 \\
    Kangaroo         & 33.2 & 47.3 & 53.7 & 38.3 & 49.5 & 38.8 & 54.3 & 47.2 & 43.5 & 59.5 & 42.9 \\
    LLaVA-OV-7B      & 42.2 & 54.6 & 57.4 & 48.6 & 42.6 & 41.4 & 60.9 & 47.9 & 50.0 & 59.5 & 47.4 \\
    LLaVA-Video-7B   & 39.1 & 50.7 & 59.3 & 46.7 & 52.5 & 52.4 & 56.5 & 53.6 & 61.9 & 47.6 & 50.6 \\
    Qwen2-VL-7B      & 41.0 & 51.7 & 66.7 & 45.8 & 54.5 & 52.8 & 65.2 & 49.0 & 60.2 & 57.1 & 51.3 \\
    InternVL2-8B     & 41.0 & 53.1 & 66.7 & 40.2 & 45.5 & 50.5 & 50.0 & 45.8 & 56.8 & 45.2 & 48.4 \\
    MiniCPM-V-2.6    & 39.8 & 45.9 & 59.3 & 34.6 & 49.5 & 51.1 & 52.2 & 42.2 & 46.6 & 50.0 & 45.7 \\
    \midrule
    PLLaVA-13B       & 31.2 & 31.9 & 46.3 & 23.4 & 48.5 & 41.1 & 45.7 & 37.0 & 41.5 & 45.2 & 37.2 \\
    InternVL2-26B    & 38.7 & 45.4 & 63.0 & 42.1 & 48.5 & 46.0 & 58.7 & 42.7 & 55.1 & 50.0 & 45.9 \\
    PLLaVA-34B       & 42.6 & 41.5 & 63.0 & 43.9 & 45.5 & 48.5 & 56.5 & 43.2 & 56.8 & 57.1 & 46.9 \\
    Tarsier-34B      & 43.0 & 48.3 & 72.2 & 45.8 & 51.5 & 50.2 & 56.5 & 49.7 & 53.7 & \secondcell{61.9} & 50.1 \\
    InternVL2-40B    & 40.2 & 58.0 & 74.1 & 51.4 & 56.4 & 53.4 & 63.0 & 57.3 & 66.9 & \secondcell{61.9} & 54.7 \\
    \midrule
    LLaVA-OV-72B     & 46.5 & \bestcell{67.6} & \secondcell{75.9} & \secondcell{57.0} & 59.4 & \secondcell{56.6} & \bestcell{73.9} & \bestcell{63.5} & 69.5 & 59.5 & \secondcell{60.0} \\
    LLaVA-Video-72B  & \secondcell{47.7} & \bestcell{67.6} & \bestcell{77.8} & \bestcell{61.7} & \secondcell{61.4} & \bestcell{57.0} & 65.2 & 62.5 & \secondcell{73.7} & 57.1 & \bestcell{60.7} \\
    Qwen2-VL-72B     & \bestcell{52.7} & \secondcell{64.7} & 74.1 & 55.1 & \bestcell{62.4} & 54.4 & \secondcell{67.4} & \secondcell{63.0} & \bestcell{76.3} & \bestcell{66.7} & \bestcell{60.7} \\
    InternVL2-76B    & 43.8 & 61.8 & 74.1 & 43.0 & 50.5 & 50.5 & 54.3 & 52.1 & 66.1 & 57.1 & 53.1 \\
    \midrule
    \rowcolor{gray!10}\multicolumn{12}{c}{\textit{\textbf{Closed-Source LMMs}}}\\
    Gemini 1.5 Flash & 40.8 & \secondcell{58.3} & \secondcell{70.4} & 52.3 & 48.0 & 54.2 & \secondcell{63.0} & 49.0 & 66.7 & \secondcell{64.3} & 53.3 \\
    Gemini 1.5 Pro   & \secondcell{49.4} & \bestcell{68.4} & 64.8 & \bestcell{59.8} & \secondcell{55.0} & \secondcell{60.4} & \bestcell{69.6} & \bestcell{64.6} & \secondcell{65.0} & \bestcell{66.7} & \bestcell{60.8} \\
    GPT-4o           & \bestcell{53.9} & 56.0 & \bestcell{81.5} & \secondcell{56.1} & \bestcell{59.4} & \bestcell{67.6} & 58.7 & \secondcell{56.8} & 63.6 & 59.5 & \secondcell{60.3} \\
    \bottomrule
    \end{tabular}%
}
  \caption{\benchmcq performance of all tested video LMMs. We provide detailed scores on 10 temporal-dynamic tasks. The best and second-best results are marked with \colorbox{best}{\makebox(26,6){\vspace{-1mm}orange}} and \colorbox{second}{\makebox(16,6){blue}}, respectively.}
  \label{tab:full_results_mcq}%
  \vspace{-1.2em}
\end{table*}

\begin{figure*}[!htbp]
    \centering
    \includegraphics[width=\linewidth]{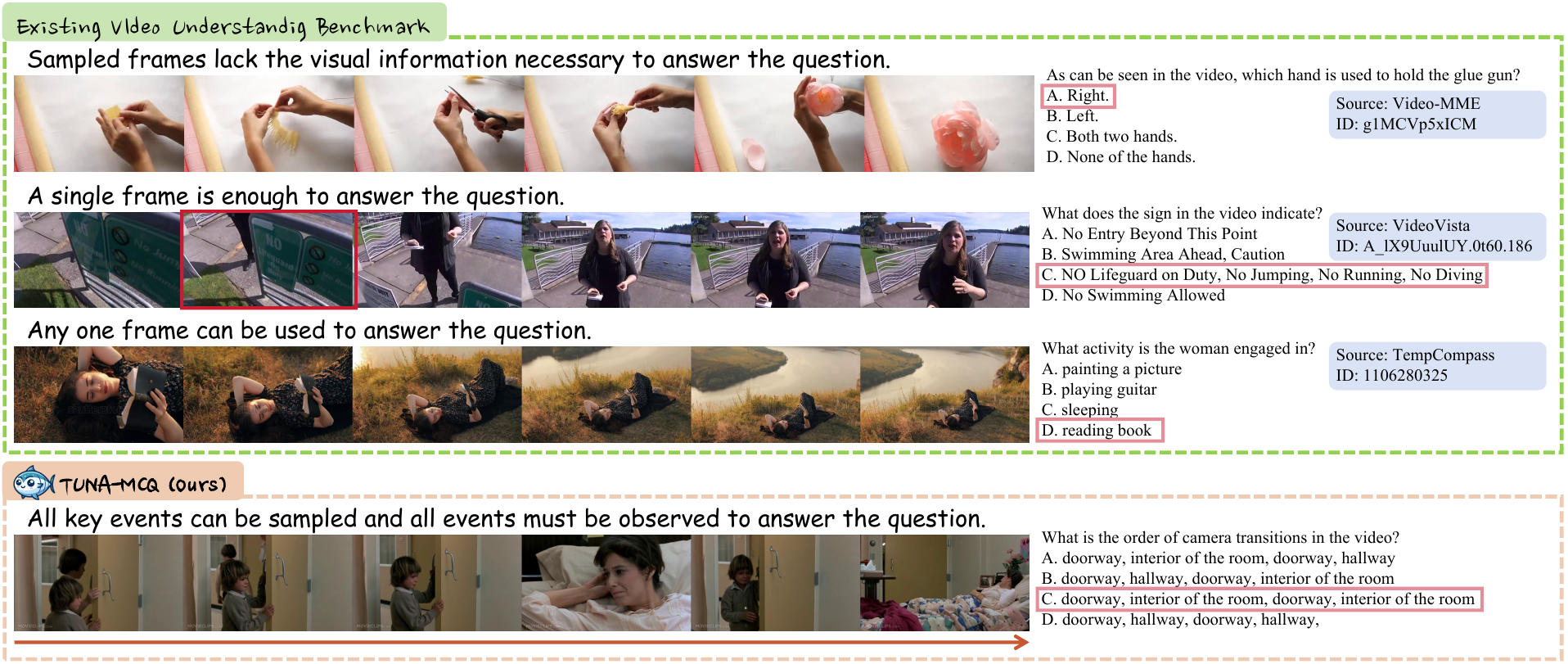}
    \caption{Several video understanding benchmark examples and analysis.}
    \label{fig:mcq_comparison}
\end{figure*}

\end{document}